\documentclass{article}

\usepackage{microtype}
\usepackage{graphicx}
\usepackage{rotating}%
\usepackage{refcount}%
\usepackage{subfigure}
\usepackage{booktabs}
\usepackage{array}%
\usepackage{xcolor}%
\usepackage{multirow}%
\usepackage{amsmath,amssymb}
\usepackage{amsthm}%
\theoremstyle{plain}
\newtheorem{theorem}{Theorem}
\newtheorem{lemma}{Lemma}
\newtheorem{corollary}{Corollary}

\theoremstyle{definition}
\newtheorem{definition}{Definition}
\newtheorem{assumption}{Assumption}

\newcommand{\restates}[2]{\expandafter\def\csname the#1\endcsname{\ref*{#2}}}
\usepackage{enumitem}%
\usepackage{wrapfig}%
\usepackage{needspace}%
\usepackage[normalem]{ulem}%
\usepackage{listings}%
\lstdefinestyle{prompt}{basicstyle=\ttfamily\scriptsize, columns=fixed, keepspaces=true, breaklines=true,
  breakatwhitespace=false, postbreak=\mbox{\textcolor{gray}{$\hookrightarrow$}\space}, frame=single,
  rulecolor=\color{gray!60}, framesep=3pt, xleftmargin=3pt, xrightmargin=3pt, showstringspaces=false,
  upquote=false, extendedchars=true,
  literate={`}{{\char18}}1 {'}{{\char13}}1 {·}{{$\cdot$}}1 {μ}{{$\mu$}}1 {…}{{\ldots}}1 {→}{{$\to$}}1 {∈}{{$\in$}}1 {−}{{-}}1 {∞}{{$\infty$}}1
    {≤}{{$\le$}}1 {≥}{{$\ge$}}1 {⊆}{{$\subseteq$}}1 {⌊}{{$\lfloor$}}1 {⌋}{{$\rfloor$}}1 {─}{{-}}1 {│}{{|}}1
    {└}{{+}}1 {├}{{+}}1 {§}{{\S}}1 {×}{{$\times$}}1 {Δ}{{$\Delta$}}1 {ε}{{$\varepsilon$}}1 {η}{{$\eta$}}1
    {θ}{{$\theta$}}1 {λ}{{$\lambda$}}1 {ν}{{$\nu$}}1 {‖}{{$\|$}}1 {₀}{{$_0$}}1 {ℓ}{{$\ell$}}1
    {◆}{{$\diamond$}}1 {⟨}{{$\langle$}}1 {⟩}{{$\rangle$}}1}
\lstdefinestyle{promptmd}{style=prompt, moredelim=[is][\itshape]{*}{*},
  moredelim=[il][]{\#\#\#\ }, moredelim=[il][]{\#\#\ }, moredelim=[il][]{\#\ }}
\usepackage[most]{tcolorbox}%
\definecolor{promptaccent}{RGB}{31,58,96}
\definecolor{promptback}{RGB}{246,248,251}
\lstdefinestyle{promptbox}{style=prompt, frame=none, xleftmargin=0pt, xrightmargin=0pt,
  basicstyle=\sffamily\footnotesize\linespread{1.05}\selectfont, columns=fullflexible, aboveskip=0pt, belowskip=0pt,
  postbreak={}, breakindent=0pt, moredelim=[is][\itshape]{*}{*},
  literate={{`}{{`}}1 {'}{{'}}1 {·}{{$\cdot$}}1 {μ}{{$\mu$}}1 {…}{{\ldots}}1 {→}{{$\to$}}1 {∈}{{$\in$}}1 {−}{{-}}1 {∞}{{$\infty$}}1
    {≤}{{$\le$}}1 {≥}{{$\ge$}}1 {⊆}{{$\subseteq$}}1 {⌊}{{$\lfloor$}}1 {⌋}{{$\rfloor$}}1 {─}{{-}}1 {│}{{|}}1
    {└}{{+}}1 {├}{{+}}1 {§}{{\S}}1 {×}{{$\times$}}1 {Δ}{{$\Delta$}}1 {ε}{{$\varepsilon$}}1 {η}{{$\eta$}}1
    {θ}{{$\theta$}}1 {λ}{{$\lambda$}}1 {ν}{{$\nu$}}1 {‖}{{$\|$}}1 {₀}{{$_0$}}1 {ℓ}{{$\ell$}}1
    {◆}{{$\diamond$}}1 {⟨}{{$\langle$}}1 {⟩}{{$\rangle$}}1},
  moredelim=[il][\normalsize\bfseries\color{promptaccent}]{\#\ },
  moredelim=[il][\small\bfseries\color{promptaccent}]{\#\#\ },
  moredelim=[il][\bfseries\color{promptaccent}]{\#\#\#\ }}
\newtcbinputlisting{\promptinput}[2]{enhanced, breakable, listing only, listing file={#2},
  listing options={style=promptbox}, title={#1}, title after break={},%
  fonttitle=\sffamily\bfseries\small, coltitle=white, colbacktitle=promptaccent, colframe=promptaccent,
  colback=promptback, boxrule=0.6pt, arc=2.5pt, left=7pt, right=7pt, top=5pt, bottom=5pt,
  toptitle=2pt, bottomtitle=2pt, before skip=6pt, after skip=6pt}
\newtcblisting{promptlisting}[1]{enhanced, breakable, listing only, 
  listing options={style=promptbox}, title={#1}, title after break={},%
  fonttitle=\sffamily\bfseries\small, coltitle=white, colbacktitle=promptaccent, colframe=promptaccent,
  colback=promptback, boxrule=0.6pt, arc=2.5pt, left=7pt, right=7pt, top=5pt, bottom=5pt,
  toptitle=2pt, bottomtitle=2pt, before skip=6pt, after skip=6pt}
\usepackage{hyperref}
\usepackage{xurl}%
\usepackage{etoc}%

\let\tocaddcontentsline\addcontentsline%
\usepackage[accepted]{mlsys2025}
\makeatletter\renewcommand{\Notice@String}{Preprint.}\makeatother%
\hypersetup{pdfsubject={Preprint}}
\makeatletter
\long\def\@makecaption#1#2{%
 \vskip 4pt
        \baselineskip 11pt
        \setbox\@tempboxa\hbox{#1. #2}%
        \ifdim \wd\@tempboxa >\hsize
        \sbox{\newcaptionbox}{\small\sl #1.~}%
        \newcaptionboxwid=\wd\newcaptionbox
        \usebox\newcaptionbox {\footnotesize #2}%
        \else
          \centerline{{\small\sl #1.} {\small #2}}%
        \fi}
\makeatother
\usepackage{etoolbox}
\makeatletter
\long\def\lofclaim#1{\lofclaim@a#1\lofclaim@end}
\long\def\lofclaim@a#1#2\lofclaim@end{\ifx\textbf#1\expandafter\@firstoftwo\else\expandafter\@secondoftwo\fi
  {\lofclaim@b#2\lofclaim@end}{#1#2}}
\long\def\lofclaim@b#1#2\lofclaim@end{#1}
\patchcmd{\@caption}{{\ignorespaces #2}}{{\ignorespaces \lofclaim{#2}}}{}{\typeout{WARNING: list-of-figures claim patch failed}}
\makeatother
\mlsystitlerunning{AutoAssumption: Searching over Assumptions Discovers ML-Systems Algorithms}

\newcommand{\oursmark}{~(ours)}
\definecolor{fieldkv}{HTML}{2A78D6}\definecolor{fieldprune}{HTML}{008300}\definecolor{fieldspec}{HTML}{6250D6}
\newcommand{\fkv}[1]{\textcolor{fieldkv}{#1}}\newcommand{\fprune}[1]{\textcolor{fieldprune}{#1}}\newcommand{\fspec}[1]{\textcolor{fieldspec}{#1}}

\newcommand{\MCMPos}{56}

\newcommand{\MCMBenchUp}{26}

\newcommand{\BTPkvFam}{11}

\newcommand{\BTPspecFam}{8}

\newcommand{\BTPpruneFam}{7}

\newcommand{\FBpruneOursRatio}{1.195}
\newcommand{\FBpruneBestPriorRatio}{1.213}

\makeatletter
\AtBeginDocument{\let\origcaption\caption\let\origlabel\label}
\newcommand{\linkedtable}[3]{\begingroup
  \renewcommand{\caption}[2][]{\origcaption{#2}\origlabel{#1}\let\label\@gobble}%
  \renewcommand{\label}[1]{}%
  #3\endgroup}
\newcommand{\linkedtablehere}[3]{\begingroup
  \let\floattable\table\let\endfloattable\endtable
  \renewenvironment{table}[1][]{\floattable[H]}{\endfloattable}%
  \renewenvironment{table*}[1][]{\begin{table}}{\end{table}}%
  \linkedtable{#1}{#2}{#3}\endgroup}

\newcommand{\plaincaption}[2]{\def\@captype{#1}\caption{#2}}
\makeatother

\begin{document}
\let\bookmarkaddcontentsline\addcontentsline
\def\addcontentsline#1#2#3{\bookmarkaddcontentsline{#1}{#2}{#3}\tocaddcontentsline{#1}{#2}{#3}}

\twocolumn[
\mlsystitle{AutoAssumption: Searching over Assumptions Discovers ML-Systems Algorithms}

\begin{mlsysauthorlist}
\mlsysauthor{Igor Sadalski}{csail}
\mlsysauthor{Minghao Guo}{csail}
\mlsysauthor{Benjamin Jones}{csail}
\mlsysauthor{Wojciech Matusik}{csail}
\end{mlsysauthorlist}
\mlsysaffiliation{csail}{MIT CSAIL, Cambridge, MA, USA}
\mlsyscorrespondingauthor{Igor Sadalski}{igors14@csail.mit.edu}
\mlsyskeywords{Machine Learning, MLSys, KV cache compression, speculative decoding, formal verification}

\vskip 0.15in%
\begin{abstract}
Many advances in ML systems came from breaking an assumption that earlier methods took for granted: speculative decoding dropped the assumption that a model needs one forward pass per token. Yet algorithms rarely state their assumptions, so experts and discovery harnesses search over programs, where one edit changes several design choices and gains are difficult to attribute. Searching over the assumptions themselves remains largely unexplored. Writing algorithms as proofs makes their assumptions explicit and turns algorithm design into a search over a small, structured space that an LLM can explore and a proof checker can audit. We formalize 89 algorithms from KV-cache eviction, weight pruning and speculative decoding as proof graphs with explicit assumptions, checked in Lean. Our harness, AutoAssumption, relaxes one assumption of a published method, re-derives and re-proves it, and generates its code. In KV-cache eviction and weight pruning, this search improved on every published algorithm we ran and produced state-of-the-art algorithms. The eviction algorithm has the highest RULER-4K score of the published methods we ran, and the pruning algorithm the lowest perplexity of thirteen on four LLaMA models. At equal budget, the harness scores within a point of OpenEvolve, an open-source implementation of AlphaEvolve, with a lower standard deviation across repeats. Future work will formalize more fields and combine them through the principal assumptions they share.
\end{abstract}

\vskip 0.1in
\begin{minipage}{\textwidth}
  \centering
  {\includegraphics[width=0.85\textwidth]{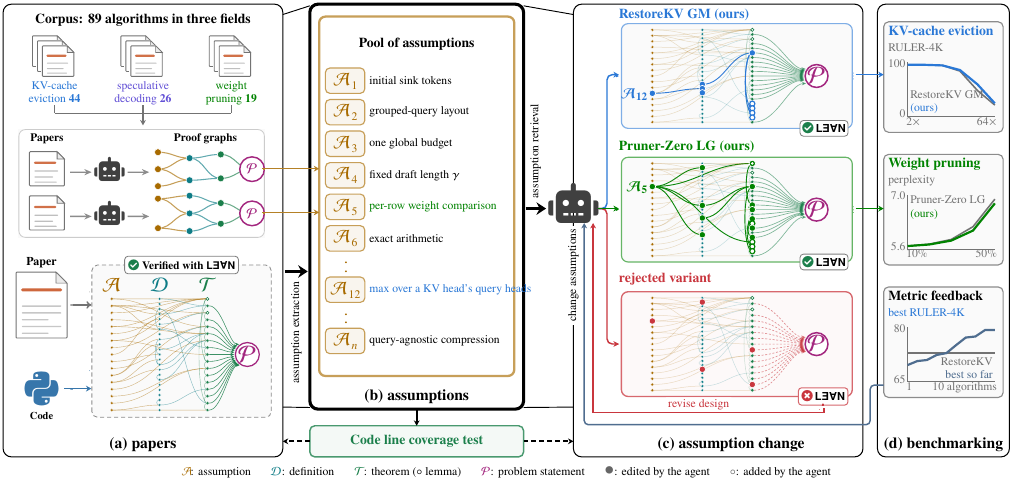}}%
  \vspace{-0.5ex}%
  \plaincaption{figure}{\textbf{AutoAssumption turns published algorithms into Lean-checked proofs and discovers new ones by changing one assumption at a time.}
  (a)~An agent writes each published algorithm as a proof graph of assumptions, definitions and theorems, checked in
  Lean. (b)~The assumptions of all proofs form one pool. (c)~A variant changes one assumption of its parent and re-proves
  only the statements that depend on it; Lean rejects a proof that fails (dashed red). (d)~Variants that pass are
  benchmarked, here on RULER-4K and WikiText-2 perplexity, and the scores guide the next change.}
  \label{fig:lead}
\end{minipage}
\vskip 0.1in
]

\printAffiliationsAndNotice{}

\etocdepthtag.toc{body}%
\section{Introduction}
\label{sec:intro}

\begin{figure*}[t]
  \centering
  \includegraphics[width=\linewidth]{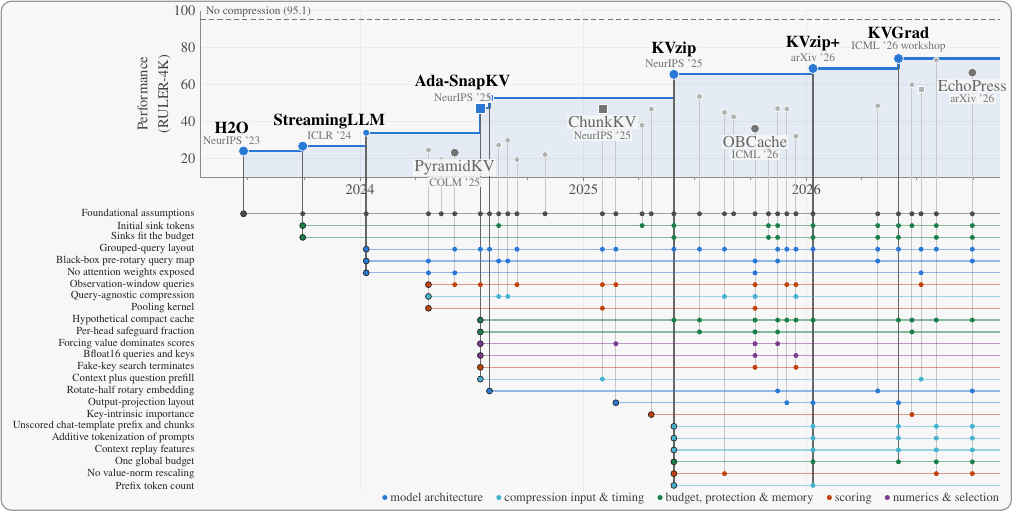}%
  \vspace{-1ex}
  \caption{\textbf{Records in KV-cache eviction coincide with new assumptions.} Top: mean RULER-4K score of
  published algorithms by release date (KVPress, Llama-3.1-8B-Instruct and Qwen3-8B, $2\times$--$64\times$
  compression, subsampled questions). The step line is the best score so far, bold labels name records, and squares mark
  query-aware algorithms. Bottom: one row per assumption shared by at least two algorithms; a dot marks an algorithm
  that makes it, and a ring the first algorithm to do so. The Foundational row holds the assumptions of H2O.}
  \label{fig:sota-assumptions}
\end{figure*}

Many advances in ML systems came from overturning an assumption that earlier methods took for granted. Speculative
decoding \citep{leviathan2023fast,chen2023accelerating} dropped the assumption that every token needs its own pass of
the large model, and PagedAttention \citep{kwon2023vllm} the assumption that a request's KV cache is contiguous. In
KV-cache eviction, five of the seven records after H2O \citep{zhang2023h2o} introduced a new assumption
(Figure~\ref{fig:sota-assumptions}). Each such advance is easy to explain afterwards but was difficult to anticipate.
We therefore hypothesize that assumptions, not code, are the productive unit of algorithm design.

Yet both experts and LLM harnesses search over programs, where it is difficult to tell which assumption produced a gain.
Discovery harnesses such as AlphaEvolve, OpenEvolve and AIDE
\citep{novikov2025alphaevolve,sharma2025openevolve,jiang2025aide} and AI-scientist systems \citep{lu2024aiscientist}
edit code, and a single edit typically touches several design decisions together. They work on programs because
published algorithms seldom spell out their assumptions. A paper often states an assumption in a single sentence or leaves it in its code,
e.g., whether one KV-cache budget spans all layers, or when a speculative decoder stops drafting. Experts carry
such assumptions as intuition, which makes it difficult to change them deliberately.

We propose to make the assumptions explicit by writing each algorithm as a logical proof. Starting from 89 algorithms with their papers
and code, 44 in KV-cache eviction, 26 in speculative decoding and 19 in LLM weight pruning, an LLM proves that each
solves its field's problem. We record the proof as a \emph{proof graph}: its nodes are assumptions, definitions and
theorems, and its edges are citations (Figures~\ref{fig:proofgraph} and~\ref{fig:fourgraphs}). A proof counts only if Lean \citep{moura2021lean4}
checks it completely, with no unproved statements.

Once we have a proof, we can change one assumption and derive a new proof and a new algorithm. Changing one assumption of a proof graph points to the
theorems that cite it, and thus to the parts of the algorithm that must change; re-deriving them yields a new
algorithm with its own proof. For example, SVIP \citep{zhang2025svip} breaks the fixed draft length of speculative
sampling, and its graph shows that only the cost claim depends on it. Doing this by hand is laborious, and these
fields publish new algorithms faster than anyone can re-derive them (Appendix~\ref{app:wall}).

We automate this process with our harness, AutoAssumption, which formalizes published algorithms and changes their
assumptions (Figures~\ref{fig:lead} and~\ref{fig:chain}). It identifies the problem a field's algorithms share, starting from the
field's benchmark. It writes each algorithm's proof graph from its paper and code, checks it in Lean together with a
coverage test that maps every line of code to a statement of the proof, and pools the assumptions. It then relaxes one
assumption of a parent, re-derives and re-proves the dependent theorems, and generates the code before any benchmark
run. We ask two questions about the harness: can it discover new performant algorithms, and how does it compare
with other LLM harnesses?

In two fields, AutoAssumption improves on every published algorithm we ran, and it performs similarly to other
harnesses. It found a higher-scoring
variant for each of the 26 published algorithms it started from. RestoreKV group mean\oursmark{} (RestoreKV GM)
outperforms every published KV-cache eviction algorithm we ran on RULER-4K, averaged over $2\times$--$64\times$
compression and two models (Figure~\ref{fig:grid}A). It outperforms its parent by up to 12.0 points at high
compression (Figure~\ref{fig:grid}B). Pruner-Zero layer group\oursmark{} (Pruner-Zero LG) adds less perplexity over the dense model than
any of the eleven published pruners we tested, at every sparsity on LLaMA-7B (tied with its parent at 10\,\%) and on every model at 50\,\% (Figure~\ref{fig:grid}E and Table~\ref{tab:settings}, bottom). SVIP lookup fallback\oursmark{} (SVIP LK) is 4\,\% faster than SVIP at batch size 1
(Figure~\ref{fig:grid}G) while keeping its exactness proof. With the same model and budget, the harness nearly matches
OpenEvolve and scores 0.8 points above AIDE, and its results vary less across repeats (Section~\ref{sec:res-search}).

Searching over assumptions thus yields strong algorithms whose every change is interpretable and checked by proof. As
next steps, we plan to formalize more fields, look for principal assumptions shared across them and unify the fields
with this method.

Our contributions are:
\begin{enumerate}[label=\arabic*., leftmargin=1.4em, itemsep=0pt, parsep=0pt, topsep=2pt, partopsep=0pt]
\item \textbf{Two new state-of-the-art algorithms for KV-cache eviction and LLM weight pruning, discovered by our
  method.} The eviction algorithm scores 75.7 on RULER-4K, averaged over two models and $2\times$--$64\times$
  compression, against 74.1 for the best published algorithm we ran. Under the SparseGPT and Wanda protocol, the pruning
  algorithm reaches a WikiText-2 perplexity of 6.78 on LLaMA-7B at 50\,\% sparsity, against 6.89, and stays best on
  four LLaMA models (Section~\ref{sec:exp-results}). A third variant is a competitive speculative decoder.
\item \textbf{A blueprint for assumption-based algorithm search, with performance competitive with code-level
  harnesses.} AutoAssumption relaxes one assumption of a parent, re-derives only
  what depends on it, and checks the new proof in Lean before any benchmark run. With the same model and budget, it
  scores 0.4 points below OpenEvolve and 0.8 points above AIDE, with a lower standard deviation (SD) across repeats (0.75, against 1.39 and 4.09) (Section~\ref{sec:res-search},
  Figure~\ref{fig:harness}).
\item \textbf{Reformulation of 89 ML-systems algorithms across three fields.} We formalize 89 algorithms as proof graphs and merge their assumptions into one
  pool that points back to every paper.
\end{enumerate}

\section{Method}
\label{sec:method}

\label{sec:method-overview}

We write each algorithm as a proof, change one assumption at a time and automate both with an LLM harness.
Section~\ref{sec:method-sps} writes a published algorithm as a \emph{proof graph}, a
proof that it solves a formally stated problem with its assumptions explicit. Section~\ref{sec:method-harness}
describes the LLM harness that automates the formalization and the derivation of variants over 89 algorithms in three fields (Appendix~\ref{app:wall}).

\subsection{Algorithms from three ML-systems fields can be formalized as proof graphs}
\label{sec:method-sps}
\label{sec:method-variants}

\begin{figure}[t]%
  \centering
  \includegraphics{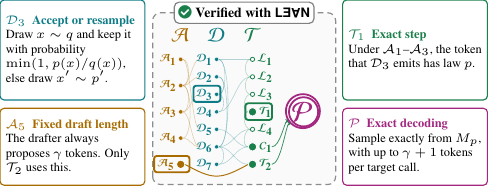}
  \vspace{-3ex}
  \caption{\textbf{The exactness proof of speculative sampling \citep{leviathan2023fast} can be represented as a proof graph.}
  The graph contains every node of Appendix~\ref{app:sps}. Its columns hold the assumptions $\mathcal A$, the
  definitions $\mathcal D$ and the theorems $\mathcal T$; open discs are lemmas $\mathcal L$. Each call-out paraphrases the highlighted node of the same colour, and the badge marks that the proof
  compiles in Lean.}
  \label{fig:proofgraph}
\end{figure}

We cast every algorithm of a field as a \emph{proof graph}. A proof graph is a directed acyclic
graph whose nodes are assumptions $\mathcal A$, definitions $\mathcal D$ and theorems $\mathcal T$, whose edges are
citations, and which ends in the problem statement $\mathcal P$. The assumptions are what the argument takes as given,
the definitions are what the program computes, and the theorems derive from both that the program solves $\mathcal P$.
Figure~\ref{fig:proofgraph} shows the proof graph of speculative sampling (SpS)
\citep{leviathan2023fast,chen2023accelerating}. In SpS, a small drafter $M_q$ proposes $\gamma$ tokens and the large
target $M_p$ scores them in one forward pass; an accept-or-resample rule keeps the output distribution exactly that
of $M_p$. Among its assumptions, the target and the
drafter are distributions over the same finite vocabulary (Assumptions~\ref{as:target-full} and~\ref{as:draft}).
Among its definitions is one accept-or-resample step (Definition~\ref{def:step-full}). From these, it proves that the
emitted token has the target's distribution (Theorem~\ref{thm:exact-full}) and derives how many tokens a step emits in
expectation (Theorem~\ref{thm:tokens-full}). Appendix~\ref{app:sps} states and proves every node.

The problem statement is the algorithm's \emph{contract}, and the theorems prove it. For speculative sampling, the
contract states the output distribution and the number of tokens one target pass emits. A benchmark
measures performance metrics, such as the speedup on Spec-Bench.

\begin{figure}[t]%
  \centering
  \includegraphics{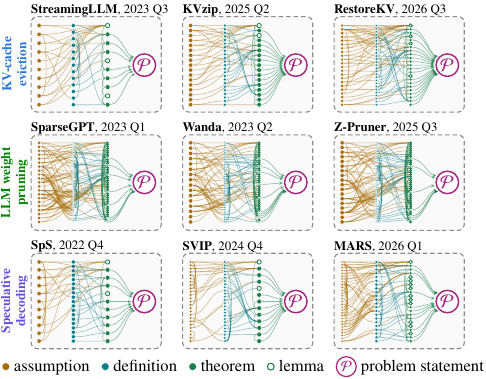}
  \vspace{-3ex}
  \caption{\textbf{Algorithms from three fields of ML systems can be written as proof graphs that answer the problem
  $\mathcal P$ shared within each field.} Each card is one algorithm's graph, drawn as in
  Figure~\ref{fig:proofgraph}; a line takes the colour of the node at its left end, and $\mathcal P$ is the problem
  of the card's field. The quarter after each name is the algorithm's first public release. Rows, left to right:
  StreamingLLM \citep{xiao2024streamingllm}, KVzip \citep{kim2025kvzip}, RestoreKV \citep{baek2026restorekv}; SparseGPT
  \citep{frantar2023sparsegpt}, Wanda \citep{sun2024wanda}, Z-Pruner \citep{bhuiyan2025zpruner}; SpS
  \citep{leviathan2023fast}, SVIP \citep{zhang2025svip}, MARS \citep{song2026mars}. Appendix~\ref{app:wall} lists
  every algorithm we analyze.}
  \label{fig:fourgraphs}
\end{figure}

Proof graphs apply across fields. Figure~\ref{fig:fourgraphs} shows three speculative decoders, SpS, SVIP and MARS,
next to three KV-cache eviction algorithms and three LLM weight-pruning algorithms.

\subsection{Variant proofs can be automated with an LLM harness}
\label{sec:method-dataset}
\label{sec:method-harness}
\label{sec:method-designproof}
\label{sec:method-lean}
\label{sec:method-ablationproof}
\begin{figure}[t]
  \centering
  \includegraphics{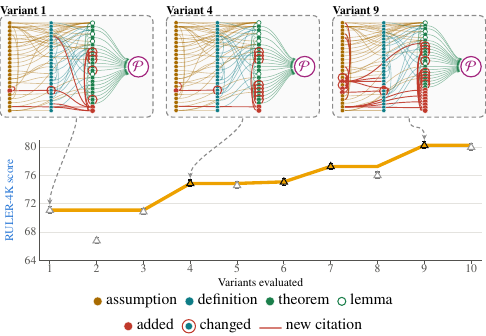}
  \vspace{-2ex}
  \caption{\textbf{A search chain in KV-cache eviction: each accepted variant changes one assumption of the current best and becomes the next parent.}
  The best of the three repeats of Figure~\ref{fig:harness}, starting from RestoreKV, on a fixed draw of 200 RULER-4K questions.
  Above: the proof graphs of variants 1, 4 and 9. The line is the best variant score so far (RestoreKV itself, 72.9, is not
  plotted); filled triangles mark variants that set a new best and became the next parent, hollow ones variants that
  did not. Error bars ($\pm1$ SD over question bootstraps, about $0.5$; Appendix~\ref{app:unc-questions}) barely extend
  beyond the markers.}
  \label{fig:chain}
\end{figure}

Our harness automates the method in three stages (Appendix~\ref{app:repro-ours}). It identifies and formalizes the
central problem, formalizes the existing solutions, and changes the assumptions of those solutions to generate variants.

\textbf{Stage 1: identifying the problem.} Stage 1 collects a field's reproducible algorithms from its benchmark
and states one problem they all solve. We trace citation
networks from the core evaluation of a field, i.e., its benchmark (Appendix~\ref{app:repro-selection},
Figure~\ref{fig:selection}). We take the papers of the algorithms that the benchmark already implements, then the papers
that cite them, and a person keeps the algorithms that are relevant and can run on the benchmark. We rewrite each kept
algorithm in the benchmark as one self-contained Python file. The file must reproduce the benchmark's reported scores for an
algorithm that the benchmark implements, and some of the results of the algorithm's own paper for an algorithm we ported. An LLM
then reads these algorithms and proposes one problem that all of them can be proved to solve. We keep the problem once a
random sample of algorithms can be formalized on it, and ask for a new formulation otherwise. A shared problem makes the
assumptions of different algorithms comparable. For speculative decoding, the benchmark is Spec-Bench
\citep{xia2024specbench}, and five of our algorithms are decoders it already implements. We ported 18 more and added 3
baselines, for 26 algorithms in all (Appendix~\ref{app:wall}).

\textbf{Stage 2: formalizing existing solutions.} Two agents turn each algorithm into a Lean-checked proof, each following a fixed prompt
(prompts in Appendix~\ref{app:prompts}, workflow in Appendix~\ref{app:repro-stages} and Figure~\ref{fig:formalization}).
The first agent receives the algorithm's folder, with its code, scores and paper (Appendix~\ref{app:prompts-design}). It
confirms by hash that the code is the version that produced the scores, and writes the proof document from the code.
We treat the code as the source of truth for what is defined, and use the paper as a guide to which assumptions to
state. The document has one definition per logic line and ends in a main theorem that the code meets the problem's
contract. A coverage test checks that the document covers every logic line of the code. The second agent
transcribes the document into Lean (Appendix~\ref{app:prompts-lean}). Its file states the assumptions, the definitions
and the main theorem, and proves the theorem. A proof counts only if Lean checks it completely,
with no step left unproved. For speculative decoding, the harness formalized all 26 algorithms. Their proofs
establish exactness for eight published decoders, five of which had claimed it without proof
(Appendix~\ref{sec:res-proofs}). We then merge the assumptions of all proofs into one pool.

\textbf{Stage 3: changing assumptions.} We derive variants one assumption at a time (workflow in
Appendix~\ref{app:repro-ablate} and Figure~\ref{fig:design-search}). An agent receives a parent's code, proof and
scores, the pool, and a prompt that describes how to derive a variant and its proof from the parent's proof
(Appendix~\ref{app:prompts-ablation}). It makes one move on one assumption, which either relaxes or strengthens it. A
relaxation drops the assumption, weakens it or replaces it with another algorithm's answer from the pool. Ideally, the
move comes with a \emph{dial}, a parameter of the new assumption whose value recovers the parent. The agent writes the new assumption and its hypothesis before
any code, and only then derives the definitions and the code from them. Before any GPU time, the variant must differ
from every known algorithm and from the parent on some input, and, if it has a dial, reproduce the parent at the dial's
recovering value. Its proof is
the parent's, edited only within the \emph{impact set} of the changed assumption, the statements that cite it
directly or through other statements; the rest of the proof stays as it is. Changes are marked in red.
We then benchmark the variant in the parent's setting, and
its scores guide the next choice of parent and assumption. Iterating from the best algorithm so far gives a search chain.
Figure~\ref{fig:chain} shows one such chain in KV-cache eviction.

\begin{table*}[t]
  \caption{\textbf{We evaluate our designs on established benchmarks in each field.} The benchmark, model, metric and error bars behind
  Figure~\ref{fig:grid}; Section~\ref{sec:experiments} gives the other settings,
  Appendices~\ref{app:repro} and~\ref{app:repro-uncertainty} the versions, hardware and error bars in full.}
  \label{tab:setup}
  \centering\footnotesize
  \setlength{\tabcolsep}{4pt}
  \begin{tabular}{@{}>{\raggedright\arraybackslash}p{2.0cm}>{\raggedright\arraybackslash}p{4.7cm}>{\raggedright\arraybackslash}p{4.7cm}>{\raggedright\arraybackslash}p{4.7cm}@{}}
    \toprule
     & \textcolor{fieldkv}{\textbf{KV-cache eviction}} & \textcolor{fieldprune}{\textbf{LLM weight pruning}} & \textcolor{fieldspec}{\textbf{Speculative decoding}} \\
    \midrule
    Benchmark & KVPress RULER-4K (13 tasks, 6,500 questions) & WikiText-2 test perplexity, SparseGPT and Wanda protocol
      (128 C4 calibration windows) & Spec-Bench (480 questions, 1,024 new tokens) \\
    \addlinespace[1pt]\cmidrule(lr){1-4}\addlinespace[1pt]
    Model & Llama-3.1-8B-Instruct \citep{grattafiori2024llama3} and Qwen3-8B \citep{yang2025qwen3} & LLaMA-7B \citep{touvron2023llama} & Vicuna-7B-v1.3 \citep{chiang2023vicuna} target with the vicuna-68m drafter \\
    \addlinespace[1pt]\cmidrule(lr){1-4}\addlinespace[1pt]
    Main metric & RULER-4K score (mean task accuracy), averaged over compression ratios $2\times$--$64\times$ and over the
      two models & WikiText-2 perplexity divided by the dense model's, at 50\,\% unstructured sparsity & speedup over autoregressive decoding at batch
      size 1, at sampling temperatures $T=0$ to $1$ \\
    \addlinespace[1pt]\cmidrule(lr){1-4}\addlinespace[1pt]
    Error bars & $\pm1$ SD over 10,000 paired bootstrap resamples of the questions & $\pm1$ paired bootstrap SE over the 166
      WikiText-2 windows & $\pm1$ SD over three paired timing runs \\
    \bottomrule
  \end{tabular}
\end{table*}

\section{Experimental Setup}
\label{sec:experiments}
\label{sec:exp-setup}

Our experiments ask two questions:
\begin{itemize}[leftmargin=2.4em, itemsep=0pt, parsep=0pt, topsep=2pt, partopsep=0pt]
\item[\textbf{Q1}] Can the harness discover new performant algorithms?
\item[\textbf{Q2}] How does it compare with other LLM harnesses?
\end{itemize}
For Q1, we run the harness in three fields and score what it finds against each field's published methods. The
fields are KV-cache eviction, LLM weight pruning and speculative decoding, each scored on its own benchmark.
Table~\ref{tab:setup} gives the main benchmark, model, metric and error bar of each field; this section adds the
other settings, so that others can reproduce each result. For Q2, we start OpenEvolve, AIDE and our harness from the
same parent algorithm, with the same model, evaluator and budget, and measure the best score each finds. Appendix~\ref{app:repro} lists model checkpoints, software versions and hardware
(Table~\ref{tab:repro-hardware}), and Appendix~\ref{app:uncertainty} explains every error bar.

\textbf{\fkv{KV-cache eviction}.} KV-cache eviction is scored on RULER and LongBench in KVPress, up to $64\times$ compression,
with timing on a dedicated H100. KVPress \citep{nvidia2026kvpress} is NVIDIA's eviction library and
public leaderboard; it implements many published algorithms, and we ported the others to it. RULER \citep{hsieh2024ruler}
(Table~\ref{tab:setup}) is a synthetic long-context benchmark of retrieval, variable-tracking, aggregation and
question-answering tasks. To test generalization, we also use RULER at 8K and 16K tokens and LongBench
\citep{bai2024longbench}, 3,750 questions from 16 English tasks with contexts of 2.6K to 30K tokens. Algorithms evict
after prefill, before the question. Most prior work stops at $16\times$ compression; we extend to the high-compression
regime of $16\times$ to $64\times$. System metrics are time to first token (TTFT), including prefill and compression,
and the peak GPU memory that PyTorch allocates. To profile them reliably, we give each job a whole H100 GPU with no
other load (verified in every run), run it at batch size 1 after an untimed warm-up, and time it over seven passes.

\textbf{\fprune{LLM weight pruning}.} Weight pruning follows the SparseGPT and Wanda protocol on four LLaMA models, scored
by perplexity relative to the dense model. The protocol is that of SparseGPT \citep{frantar2023sparsegpt} and
Wanda \citep{sun2024wanda} (Table~\ref{tab:setup}); beyond LLaMA-7B, we prune LLaMA-13B and 30B \citep{touvron2023llama}
and LLaMA-2-7B \citep{touvron2023llama2}. Every linear layer of the decoder is pruned once, without retraining, to
10--50\,\% unstructured sparsity, using calibration windows of 2,048 tokens from C4 \citep{raffel2020c4}, in fp16 on one
H100 GPU. As in Pruner-Zero's official code, Pruner-Zero and Pruner-Zero LG\oursmark{} also take gradient statistics of
the dense model on WikiText-2. WikiText-2 \citep{merity2017wikitext} is a language-modeling corpus of Good and Featured Wikipedia articles;
we concatenate its test split (60 articles, about 245K words), cut it into non-overlapping 2,048-token windows and
take the exponential of the mean per-token negative log-likelihood. We report this perplexity divided by the dense
model's, so that 1 means no loss from pruning. Baselines are eleven published algorithms,
re-implemented from their official code (Appendix~\ref{app:repro-prune}), and magnitude pruning
\citep{han2015magnitude}. Each configuration is one run with a fixed calibration seed.

\textbf{\fspec{Speculative decoding}.} Speculative decoding is scored on Spec-Bench with two target--drafter pairs, with paired
timing on a dedicated H200. Spec-Bench \citep{xia2024specbench} (Table~\ref{tab:setup})
comes with reference implementations of some published decoders; we ported the others to it. It spans six generation tasks, from
multi-turn conversation and translation to mathematical reasoning and retrieval-augmented generation. Besides the Vicuna pair, we run Llama-3.1-8B-Instruct as target with Llama-3.2-1B-Instruct \citep{grattafiori2024llama3} as
drafter. At $T=0$, decoding is greedy and a correct decoder reproduces the target's output token for token; at $T>0$,
decoding samples with top-$k=50$ and a correct decoder reproduces the target's output distribution. Besides the
speedup, we report the mean number of accepted tokens per target pass (\#MAT), throughput at $T=0$ and $T=1$, and
greedy time per output token (TPOT) at batch sizes 1 to 32. For reliable timing, each run has a whole H200 GPU with no
other load, paired runs start minutes apart on one machine, and each speedup divides by an autoregressive run of the
same target and temperature (Appendix~\ref{app:spec-timing}).

\textbf{Harness comparison.} The harness comparison gives OpenEvolve, AIDE, Fable from scratch and our harness the
same model, evaluator and budget. OpenEvolve \citep{sharma2025openevolve} and AIDE \citep{jiang2025aide} are
code-level harnesses; \emph{Fable from scratch}, a lower reference, writes algorithms from the task description alone.
The model is Fable 5.1 at high reasoning effort, the budget ten benchmarked algorithms, and all except Fable from
scratch start from RestoreKV. The benchmark is RULER-4K with
Llama-3.1-8B-Instruct and Qwen3-8B, as above. Because our GPU runtime was limited, we score each algorithm on a fixed
subset of 200 of its 6,500 questions. The metric is the best score found so far, where an algorithm's score is its mean accuracy
over the two models and six compression ratios. We repeat each search three times and report the mean across
repeats with the minimum and maximum as error bars. Appendix~\ref{app:harnesses} describes how we ran OpenEvolve
and AIDE.

\section{Results}
\label{sec:exp-results}
\label{sec:exp-kv}
\label{sec:exp-sota}

\begin{figure}[t]
  \centering
  \includegraphics{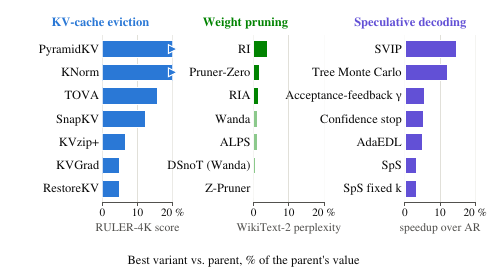}
  \caption{\textbf{For every parent algorithm, AutoAssumption generates a variant that scores higher than the parent on
  its field's search benchmark.} Across
  three fields, we selected \the\numexpr\BTPkvFam+\BTPspecFam+\BTPpruneFam\relax{} published algorithms as parents,
  and for each of them our harness found a better variant. Shown: the seven parents with the largest gain per field;
  Appendix Table~\ref{tab:best-variant-all} gives the results for the others. Triangles: gains beyond the axis; light
  bars: gains under 1\,\%.}
  \label{fig:best-variant-gain}
\end{figure}

\begin{figure*}[t]
  \centering
  \includegraphics[width=\textwidth]{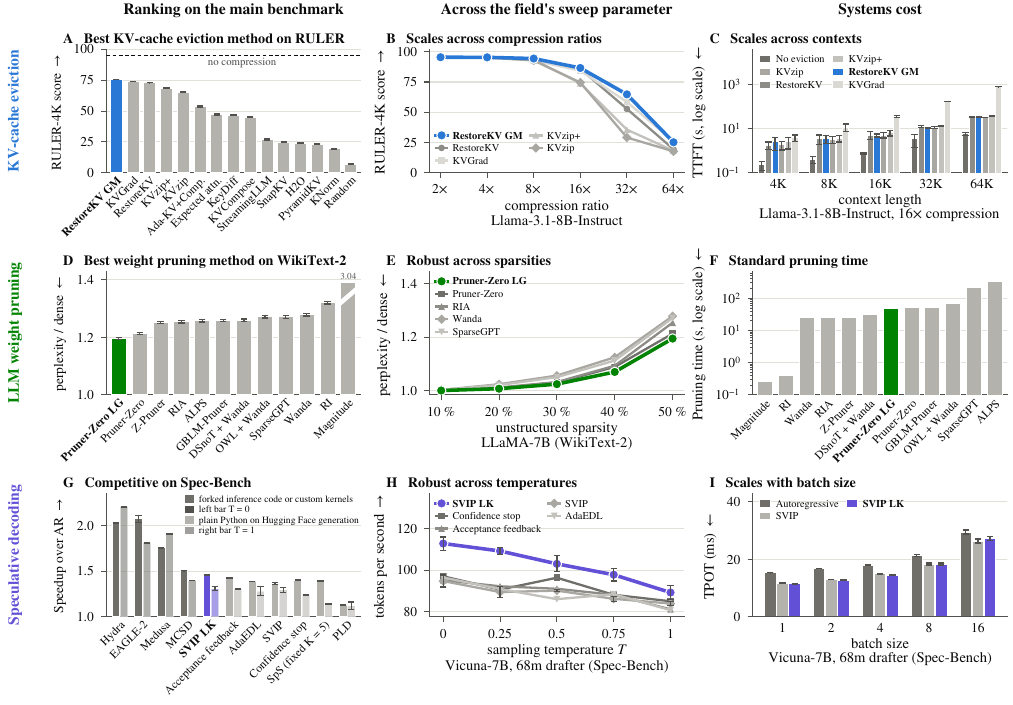}
  \vspace{-2ex}
  \caption{\textbf{Our three variants rank at or near the top of their fields, hold up across each field's sweep and
  remain practical to run.} AutoAssumption pushes the state of the art in KV-cache eviction and LLM
  weight pruning and creates a competitive speculative decoding algorithm (A, D, G). The three variants match or
  outperform the published algorithms shown at every point of the sweep each field reports, except at $64\times$ KV-cache
  compression on Llama, where KVGrad leads by 0.8 points (B, E, H). Our variants
  take about as much time as their parents in all three fields (C, F, I). Error bars as in Table~\ref{tab:setup};
  in D, a paired bootstrap SE over the WikiText-2 windows.}
  \label{fig:grid}
\end{figure*}

Our harness discovers state-of-the-art algorithms and matches other LLM harnesses. To answer Q1, whether the harness discovers new performant algorithms, we ran it from
\the\numexpr\BTPkvFam+\BTPspecFam+\BTPpruneFam\relax{} published algorithms in three fields (\BTPkvFam{} in KV-cache
eviction, \BTPpruneFam{} in weight pruning and \BTPspecFam{} in speculative decoding), scored every variant on its
field's search benchmark and kept the best one per parent. Every parent has a variant that scores higher
(Figure~\ref{fig:best-variant-gain}), although the smallest gains lie within the benchmark noise. Three of these
variants are new algorithms, each one assumption away from its parent (Table~\ref{tab:new-algorithms}): RestoreKV GM
and Pruner-Zero LG outperform every published algorithm we ran in KV-cache eviction and weight pruning
(Figure~\ref{fig:grid}A and D), and SVIP LK is faster than its parent SVIP at batch size 1
(Figure~\ref{fig:grid}G). Section~\ref{sec:res-q1} tests them further against each field's published algorithms, on
settings the search never saw and in cost. Section~\ref{sec:res-q2} answers Q2, whether the harness performs as well
as other LLM harnesses, by comparing it with OpenEvolve and AIDE at the same budget;
Appendix~\ref{app:method-ablations} adds what the proofs catch that benchmarks miss.

\subsection{Q1: Can the harness discover new performant algorithms?}
\label{sec:res-q1}
\label{sec:res-transfer}
\label{sec:exp-spec}
\label{sec:designs}
\begin{table}[t]
  \caption{\textbf{We propose three new algorithms, each one assumption away from a published parent.} Names in the
  colour of their field; pseudocode in Appendix~\ref{app:algorithms}.}
  \label{tab:new-algorithms}
  \centering\footnotesize
  \setlength{\tabcolsep}{4pt}
  \begin{tabular}{@{}>{\raggedright\arraybackslash}p{2.0cm}>{\raggedright\arraybackslash}p{4.7cm}l@{}}
    \toprule
    Algorithm & Changed assumption & Code \\
    \midrule
    \fkv{\textbf{RestoreKV GM}} & a cached pair is kept when the query heads that share its KV head attend to it on
      average, not when any single head does & Alg.~\ref{alg:restorekv} \\
    \addlinespace[2pt]
    \fprune{\textbf{Pruner-Zero LG}} & weight scores are compared within a whole matrix instead of within each output
      row & Alg.~\ref{alg:prunerzero} \\
    \addlinespace[2pt]
    \fspec{\textbf{SVIP LK}} & when the entropy test ends a draft, tokens copied from an earlier occurrence in the
      prompt are appended instead & Alg.~\ref{alg:svip} \\
    \bottomrule
  \end{tabular}
\end{table}

AutoAssumption sets the state of the art in two of the three tested ML-systems fields.\label{sec:res-sota}
In KV-cache eviction, RestoreKV GM\oursmark{} scores 75.7 on RULER-4K, averaged over both models and six compression
ratios, against 74.1 for KVGrad, the best published algorithm we ran (Figure~\ref{fig:grid}A). It leads on each model
separately (77.0 against 75.4 on Llama, 74.5 against 72.9 on Qwen; Appendix Table~\ref{tab:space-scores}). The strong
algorithms all stay near 95 at $2\times$--$8\times$; the lead comes at $16\times$ and $32\times$, up to 12.0 points
over its parent (Figure~\ref{fig:grid}B). In weight pruning, Pruner-Zero LG\oursmark{} has the lowest perplexity of
thirteen pruning algorithms we ran: on LLaMA-7B at 50\,\% sparsity it reaches \FBpruneOursRatio{} times the dense model's,
against \FBpruneBestPriorRatio{} for its parent Pruner-Zero \citep{dong2024prunerzero} and 1.25 or more for every
other published pruner (Figure~\ref{fig:grid}D). In speculative decoding, SVIP LK\oursmark{} is faster than SVIP
\citep{zhang2025svip} at batch size 1, with a speedup of 1.46 against 1.37 at $T=0$ and 1.31 against 1.30 at $T=1$
(Figure~\ref{fig:grid}G), and keeps its exactness proof. It is the fastest plain Python implementation we ran; only decoders with custom kernels or forked inference code
are faster (Figure~\ref{fig:grid}G). But does this performance generalize beyond the search setting?

\begin{table}[t]
  \caption{\textbf{The two best discovered variants hold beyond the search setting.} Top: RestoreKV GM\oursmark{}
  against its parent, score averaged over $2\times$--$64\times$; the search used 200 RULER-4K questions, and 8K and
  16K contexts and LongBench are settings it never saw. $\Delta$: variant minus parent, computed from unrounded means, $\pm1$ SD from bootstrapping
  the questions. Bottom: WikiText-2 perplexity at 50\,\% sparsity divided by the dense model's, lower is better; the
  search used only LLaMA-7B. The ranking holds across all thirteen pruning algorithms we ran; the table shows a
  selection, and Appendix Table~\ref{tab:prune-models} gives every algorithm.}
  \label{tab:settings}\label{tab:prune-selection}
  \centering\small
  \setlength{\tabcolsep}{3pt}
  \resizebox{\columnwidth}{!}{%
  \begin{tabular}{@{}lccc@{\hspace{10pt}}c@{\hspace{14pt}}c@{}}
    \toprule
     & \multicolumn{4}{c@{\hspace{14pt}}}{RULER} & LongBench \\
    \cmidrule(lr){2-5}\cmidrule(l){6-6}
     & \multicolumn{3}{c@{\hspace{10pt}}}{Llama-3.1-8B} & Qwen3-8B & Llama-3.1-8B \\
    \cmidrule(lr){2-4}\cmidrule(lr){5-5}\cmidrule(l){6-6}
    KV-cache eviction & 4K & 8K & 16K & 4K & 2.6K--30K \\
    \midrule
    RestoreKV & 73.9 & 74.8 & 73.3 & 72.1 & 37.8 \\
    \textbf{RestoreKV GM}\oursmark{} & \textbf{77.0} & \textbf{77.9} & \textbf{76.5} & \textbf{74.5} & \textbf{39.2} \\
    $\Delta$ & \begin{tabular}[c]{@{}c@{}}\textcolor{green!45!black}{+3.1}\\[-1pt]{\scriptsize\color{black!55}$\pm$0.16}\end{tabular} & \begin{tabular}[c]{@{}c@{}}\textcolor{green!45!black}{+3.1}\\[-1pt]{\scriptsize\color{black!55}$\pm$0.17}\end{tabular} & \begin{tabular}[c]{@{}c@{}}\textcolor{green!45!black}{+3.3}\\[-1pt]{\scriptsize\color{black!55}$\pm$0.17}\end{tabular} & \begin{tabular}[c]{@{}c@{}}\textcolor{green!45!black}{+2.4}\\[-1pt]{\scriptsize\color{black!55}$\pm$0.17}\end{tabular} & \begin{tabular}[c]{@{}c@{}}\textcolor{green!45!black}{+1.5}\\[-1pt]{\scriptsize\color{black!55}$\pm$0.29}\end{tabular} \\
    \midrule
     & \multicolumn{3}{c@{\hspace{10pt}}}{LLaMA} & LLaMA-2 & \\
    \cmidrule(lr){2-4}\cmidrule(lr){5-5}
    Weight pruning & 7B & 13B & 30B & 7B & \\
    \midrule
    \textbf{Pruner-Zero LG}\oursmark{} & \textbf{1.195} & \textbf{1.144} & \textbf{1.201} & \textbf{1.219} & \\
    Pruner-Zero & 1.213 & 1.165 & 1.224 & 1.226 & \\
    RIA & 1.254 & 1.194 & 1.238 & 1.244 & \\
    SparseGPT & 1.272 & 1.219 & 1.297 & 1.283 & \\
    Wanda & 1.279 & 1.209 & 1.280 & 1.265 & \\
    \bottomrule
  \end{tabular}
  }
\end{table}

The two best discovered variants generalize beyond the search setting.\label{sec:res-robust}\label{sec:exp-robust}\label{sec:exp-prune}
In KV-cache eviction, the search scored RestoreKV GM on 200 RULER-4K questions. On all 6,500 questions it improves its
parent by 3.1 points on Llama-3.1-8B-Instruct and by 2.4 points on Qwen3-8B. The gain holds at longer contexts, by 3.1
and 3.3 points at 8K and 16K tokens, and on another dataset, by 1.5 points on LongBench \citep{bai2024longbench}
(Table~\ref{tab:settings}, top). In LLM weight pruning, the search scored Pruner-Zero LG on LLaMA-7B at 50\,\%
sparsity. It has the lowest perplexity of all thirteen pruning algorithms on LLaMA-7B, 13B and 30B, and on average adds
10\,\% less perplexity above the dense model than its parent, the best published pruner. It is also best on a
different model, LLaMA-2-7B, by 3\,\% (Table~\ref{tab:settings}, bottom), and at every sparsity from 10\,\% to
50\,\% on LLaMA-7B, tied with its parent at 10\,\% (Figure~\ref{fig:grid}E).

The best variants remain practical, at costs close to their parents'.\phantomsection\label{sec:res-cost}
For each parent--variant pair we test the ratio of the variant's cost to its parent's for a difference. The time to first token of RestoreKV GM\oursmark{} is 1.03 times
RestoreKV's, pooled over contexts from 4K to 64K tokens, and not significantly different from it ($p=0.69$;
Figure~\ref{fig:grid}C; Appendix Table~\ref{tab:ttft-context}). At 4K tokens on both models the difference is also
not significant ($p=0.39$), and its peak GPU memory is identical to its parent's. SVIP LK\oursmark{} stays within 4\%
of SVIP's time per output token at every batch size up to 16 (Figure~\ref{fig:grid}I); paired over questions and
pooled over these batch sizes, it is 2.3\% faster (90\% CI 2.0--2.6\%). At batch size 32, however, it is 11.7\%
slower (90\% CI 10.6--12.7\%). Pruner-Zero LG\oursmark{} prunes LLaMA-7B in 48\,s against 53\,s for its parent
(Figure~\ref{fig:grid}F).

\subsection{Q2: How does the harness compare with other LLM harnesses?}
\label{sec:res-q2}

\begin{figure}[t]
  \centering
  \includegraphics{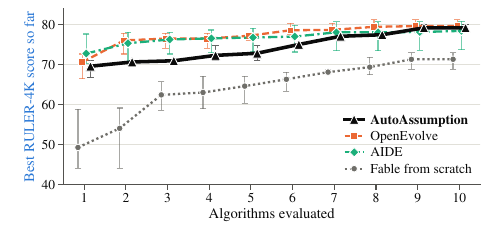}
  \caption{\textbf{AutoAssumption finishes level with the code-level LLM harnesses and improves more steadily, with a
  smaller spread across repeats.} OpenEvolve and AIDE jump early and then gain little; AutoAssumption, changing one
  assumption per step, rises at almost every step and ends at 79.3, against 79.7 and 78.5. All three start from
  RestoreKV; Fable from scratch has no parent. Scores average 200 RULER-4K questions, Llama-3.1-8B-Instruct and
  Qwen3-8B, and $2\times$--$64\times$ compression. Lines: mean of three repeats; error bars: minimum to maximum.}
  \label{fig:harness}
\end{figure}

Searching over assumptions improves more consistently than searching over code.\label{sec:res-search}\label{sec:exp-harness}
We compare with two code-level harnesses: OpenEvolve \citep{sharma2025openevolve}, an open-source implementation of
the prominent AlphaEvolve \citep{novikov2025alphaevolve}, and AIDE \citep{jiang2025aide}, a widely used tree search over
programs. All three start from RestoreKV with the same model, evaluator and budget. The code-level harnesses gain most
in their first steps: after two algorithms, OpenEvolve and AIDE reach 76.1 and 75.4 RULER points, against 70.7 for
ours. We hypothesize that each of their edits changes several design choices at once, which allows a large early
jump but gives little guidance for the next change; over the remaining eight algorithms they add only 3.6 and 3.1
points. Our harness changes one assumption per step and keeps rising, by 8.6 points over the same steps. After ten
algorithms it reaches 79.3 points, against 79.7 for OpenEvolve and 78.5 for AIDE, while Fable from scratch converges
lowest, at 71.4 (Figure~\ref{fig:harness}, Table~\ref{tab:harness-passes}). Searching over assumptions thus matches
searching over programs, with every change attributable to one assumption and a smaller spread across repeats (1.3
points after ten algorithms, against 2.5 and 7.1).

\section{Related Work}
\label{sec:related}
\textbf{Axiomatic design.} The Axiomatic Atlas \citep{guo2026axiomaticatlas} audits architectures against one fixed
axiom set; we recover each algorithm's own assumptions. The Atlas states the requirements on a neural architecture as
composable axioms over connectivity, operator contracts, depth stability and merges, and an audit returns a
certificate or names the violated axiom, which implies a targeted repair. Our assumptions come from a Lean-checked
proof of each algorithm, and a changed assumption dictates the re-proof and the code that the benchmark then scores.

\textbf{Automated design search.} Code-level harnesses evolve programs against a fixed evaluator
\citep{romera2024funsearch,novikov2025alphaevolve,sharma2025openevolve,lange2026shinkaevolve,jiang2025aide}, also for
ML research \citep{lu2024discopop,hu2025adas,lu2024aiscientist}, systems research
\citep{cheng2025barbarians,cheng2025letthem} and KV-cache eviction
\citep{poudel2026cachecraft,yu2025evolkv,park2026taskpress}. Their gains are difficult to attribute, and simple
baselines often match them \citep{gideoni2026simplebaselines,oved2026illusion}. Our harness instead changes one stated
assumption per step.

\textbf{Assumptions and formal verification.} Tracked assumptions go back to truth maintenance \citep{dekleer1986atms}
and structured proofs \citep{lamport2012proof} and underlie formal blueprints
\citep{massot2023blueprint,bertot2010dpdgraph}. AssumptionMiner makes the assumptions behind LLM-generated code
explicit \citep{wu2026assumptionminer}; SampCert, AutoOPT, LegoNE, LLMLift and Algorithmist derive algorithms together
with proofs \citep{demedeiros2025sampcert,kim2026autoopt,li2025legone,bhatia2024llmlift,kulkarni2026algorithmist}; and
ProofLoom formalizes published algorithms in Lean \citep{wang2026proofloom}. These works handle one algorithm at a
time; we pool assumptions across a field and, as in proof repair \citep{ringer2019qed}, re-prove only what a change
affects.

\section{Limitations}
\label{sec:limitations}
\label{sec:discussion}

Three limitations remain: human review, fields that resist formalization and non-canonical assumptions. First, the
formalization needs a human reader. Lean checks each proof against the problem statement, not the statement against
the field, so a person must confirm that the contract says what the field means. Second, a proof covers only what mathematics can represent. In ML systems, a
proof can show that an eviction algorithm uses less memory or that a decoder samples the target's distribution
exactly. Training dynamics, or the quality of an LLM's answers, depend on statistics of the data that we do not fully
understand, so we cannot build a proof around them; only the benchmark measures them. Third, the assumptions are
not canonical. The LLM chooses how to split an algorithm into assumptions, and some of them, such as how the GPU is
structured, admit several levels of detail or stay implicit. A second formalization could state a
different set that still proves the contract and suggests different variants. We have not measured how stable the
pool is across repeated formalizations.

\section{Conclusion}
\label{sec:conclusion}

ML-systems algorithms have advanced by breaking assumptions, yet these usually stay implicit. Writing algorithms as
proofs makes them explicit and turns algorithm design into a search over a small, structured space that an LLM can
explore and a proof checker can audit. In two fields, KV-cache eviction and weight pruning, this search improved on every published algorithm we ran
and produced state-of-the-art algorithms. Our results support the view that assumptions, not
code, are the productive unit of algorithm design: they make each change interpretable, its effect attributable and
its correctness checkable. As future work, we plan to formalize more fields and improve their algorithms. We then plan to look for
principal assumptions shared across fields and to combine the fields. Together, these steps would formalize the discipline as a whole.

\bibliography{references}
\bibliographystyle{mlsys2025}

\section*{AI Usage Disclosure}
Under the authors' direction, Claude Code (Anthropic) was used for experiment code, scheduling and evaluation
scripts, figure and table scripts, and manuscript assistance. OpenAI Codex assisted this revision by auditing
consistency, checking existing numerical artifacts, generating a matched comparison figure and editing the
manuscript. The authors are responsible for reviewing the final text, code, citations and claims. LLM-generated
algorithms, proof documents and code studied as experimental outputs are identified as such in the paper.

\appendix
\onecolumn
\makeatletter\global\@colroom\textheight\makeatother%
\etocdepthtag.toc{appendix}
\etocsettagdepth{body}{none}
\etocsettagdepth{appendix}{subsection}
\etocsettocstyle{\section*{Contents of the appendix}}{}
\tableofcontents
\renewcommand{\listfigurename}{List of figures}
\renewcommand{\listtablename}{List of tables}
\listoffigures
\listoftables
\raggedbottom%
\let\appendixsection\section
\renewcommand{\section}{\clearpage\appendixsection}
\section{Reproducibility}
\label{app:repro}
This appendix gives what is needed to rerun our experiments. Appendix~\ref{app:repro-external} lists the benchmarks,
models and machines of the three fields, and Appendix~\ref{app:repro-ours} gives the steps by which our harness
produces a design proof and a variant.

\subsection{Reproducing external results}
\label{app:repro-external}
\label{app:repro-hardware}
Each field ran on its own machine (Table~\ref{tab:repro-hardware}), one job per whole GPU through a shared queue.
Every score records its machine, software versions, run date and the SHA-256 of its code, and each algorithm is
released with its proof, code and scores. Checkpoints and datasets come from the Hugging Face Hub (Spec-Bench from
GitHub) at the revisions recorded with the results; Appendix~\ref{app:harnesses} pins OpenEvolve and AIDE.

\begingroup
\renewcommand{\caption}[2][]{\origcaption{\textbf{GPU, NVIDIA driver, CPU, Python, PyTorch, libraries and precision of the
machine behind each field's experiments (KV-cache eviction, weight pruning, speculative decoding).} One GPU per job.
Driver column: highest CUDA version the driver supports; brackets: CUDA runtime PyTorch was built
against.}\origlabel{tab:repro-hardware}\expandafter\let\expandafter\label\csname @gobble\endcsname}%
\renewcommand{\label}[1]{}%
\begin{table}[!ht]
\caption{Hardware and software of the three fields of the experiments.}
\label{tab:repro-hardware}
\vskip 0.1in
\begin{center}
\begin{footnotesize}
\setlength{\tabcolsep}{3pt}
\renewcommand{\arraystretch}{1.15}
\begin{tabular}{@{}>{\raggedright\arraybackslash}p{1.9cm}>{\raggedright\arraybackslash}p{2.1cm}>{\raggedright\arraybackslash}p{1.9cm}>{\raggedright\arraybackslash}p{2.6cm}>{\raggedright\arraybackslash}p{0.95cm}>{\raggedright\arraybackslash}p{1.9cm}>{\raggedright\arraybackslash}p{2.8cm}>{\raggedright\arraybackslash}p{1.3cm}@{}}
\toprule
Experiments & GPU (one per job) & NVIDIA driver (CUDA) & CPU & Python & PyTorch (CUDA runtime) & Libraries & Precision \\
\midrule
KV-cache eviction
  & NVIDIA H100 80GB HBM3
  & 550.144.03 (CUDA 12.4)
  & 2$\times$ AMD EPYC 9554, 128 cores
  & 3.10.12
  & 2.14.1+cu126 (12.6)
  & KVPress 0.5.5, Transformers 5.2.0
  & BF16 \\
\addlinespace
Weight pruning
  & NVIDIA H100 80GB HBM3
  & 550.144.03 (CUDA 12.4)
  & 2$\times$ AMD EPYC 9554, 128 cores
  & 3.10.12
  & 2.14.1+cu126 (12.6)
  & Transformers 5.2.0, Datasets 5.0.1, lm-eval 0.4.13
  & FP16 \\
\addlinespace
Speculative decoding
  & NVIDIA H200, 141\,GB
  & 595.58.03 (CUDA 13.2)
  & 2$\times$ Intel Xeon Platinum 8592+, 256 threads
  & 3.12.13
  & 2.5.1+cu124 (12.4)
  & Spec-Bench (commit fd2c1cd7), Transformers 4.37.1
  & FP16 \\
\bottomrule
\end{tabular}
\end{footnotesize}
\end{center}
\end{table}
\endgroup

\paragraph{KV-cache eviction.}
\label{app:repro-benchmarks}
\label{app:repro-sources}
Every algorithm runs as a press of KVPress \citep{nvidia2026kvpress} 0.5.5 on one H100, evicting after prefill and
before the question at compression ratios from $2\times$ to $64\times$. The models are Llama-3.1-8B-Instruct
(\texttt{NousResearch/Meta-Llama-3.1-8B-Instruct}, with Meta's Llama-3.1 chat template) and Qwen3-8B
(\texttt{Qwen/Qwen3-8B}), in BF16. RULER \citep{hsieh2024ruler} (\texttt{simonjegou/ruler}) has 13 synthetic tasks at
4,096 tokens, 500 questions each, scored by the mean per-task accuracy. The searches used a nested
200-question subset (seed 42), so the 6,500-question scores of Table~\ref{tab:together} are not held out.

The other KV-cache settings ran on two other machines. RULER at 8K and 16K tokens and the 16 English tasks of LongBench
\citep{bai2024longbench} (\texttt{Xnhyacinth/LongBench}, 3,750 questions) in Table~\ref{tab:settings} (top) ran on a second
H100 machine (NVIDIA driver 580.178.04, CUDA 13.0, Intel Xeon Platinum 8481C, PyTorch 2.14.1+cu130, otherwise as the first row of
Table~\ref{tab:repro-hardware}). The 300-question scores of the prior algorithms in Figures~\ref{fig:sota-assumptions}
and~\ref{fig:space} ran on the H200 machine with the KV-cache software (PyTorch 2.14.1+cu130). Our re-runs, the parent rows of
Tables~\ref{tab:together} and~\ref{tab:settings}, agree with NVIDIA's KVPress leaderboard within 0.4 RULER points across all
41 points (Figure~\ref{fig:repro}a).

\paragraph{Weight pruning.}
\phantomsection\label{app:repro-prune}
Pruning follows the protocol of SparseGPT and Wanda on LLaMA-7B, 13B and 30B (\texttt{huggyllama/llama-7b},
\texttt{-13b}, \texttt{-30b}) and LLaMA-2-7B (\texttt{NousResearch/Llama-2-7b-hf}) in FP16. Every linear layer of the
decoder is pruned once, layer by layer, without retraining. Calibration draws 128 random 2,048-token windows (seed 0)
from the first C4 training shard (\texttt{allenai/c4}, \texttt{en/c4-train.00000-of-01024}); Pruner-Zero computes its
gradient statistics of the dense model on WikiText-2. The scores are WikiText-2 test perplexity
(\texttt{wikitext-2-raw-v1}, 2,048-token windows) and, at 50\,\%, the mean zero-shot accuracy of lm-eval 0.4.13 on
Wanda's seven tasks (BoolQ, RTE, HellaSwag, WinoGrande, ARC-easy, ARC-challenge, OpenBookQA).

The eleven published algorithms are single files ported from their official code, run in one harness on the KV-cache
H100 machine (78 runs on the second H100 machine). Where their papers report a cell, our perplexities agree within 0.1.
Each cell is one run; its error bar is a bootstrap SE over the 166 WikiText-2 windows.

\paragraph{Speculative decoding.}
\phantomsection\label{app:spec-timing}
Spec-Bench \citep{xia2024specbench} (\texttt{hemingkx/Spec-Bench}, commit \texttt{fd2c1cd}) has 480 questions in six
tasks. The target--drafter pairs are Vicuna-7B-v1.3 (\texttt{lmsys/vicuna-7b-v1.3}) with vicuna-68m
(\texttt{double7/vicuna-68m}) and Llama-3.1-8B-Instruct (\texttt{NousResearch/Meta-Llama-3.1-8B-Instruct}) with
Llama-3.2-1B-Instruct (\texttt{unsloth/Llama-3.2-1B-Instruct}); the two Llama models share Meta's Llama-3.1 chat template and tokenizer.
The native baselines use Spec-Bench's code with the Medusa (\texttt{FasterDecoding/medusa-vicuna-7b-v1.3}), EAGLE
(\texttt{yuhuili/EAGLE-Vicuna-7B-v1.3}) and Hydra (\texttt{ankner/hydra-vicuna-7b-v1.3}) heads; SVIP follows its
authors' code (\texttt{Geralt-Targaryen/SVIP}, commit \texttt{0cd912e}). Every algorithm runs in Spec-Bench's
unmodified evaluation loop, which does the prompting, warm-up and timing, on all 480 questions, with at most 1,024 new tokens, in FP16 under Transformers 4.37.1.

SVIP keeps its authors' constants (stop drafting when $\sqrt{cH}>h$ with $c=0.15$, $h=0.3$; at most 40 draft tokens),
and SVIP LK\oursmark{} adds only PLD's prompt-lookup parameters (suffix $n$-grams up to 3, at most 10 lookup tokens). At
$T>0$, drafter and target sample with the same temperature and top-$k=50$. Each \#MAT and speedup cell, at five
temperatures, is the mean of three paired runs with seed 0, so the runs of an algorithm emit the same tokens and differ
only in timing.

Timed runs are isolated and paired. Each is its own queue job on a whole H200, whose only other context is an idle
placement anchor that launches no kernels; no two of the 60 SVIP and SVIP LK\oursmark{} runs shared a GPU. Run $k$ of
SVIP and of SVIP LK\oursmark{}, for the same model pair and temperature, started within 15 minutes of each other (median
3) on two GPUs of one machine, and each speedup divides by the autoregressive run $k$ of the same target and
temperature.

The systems profiles behind panels G and I of Figure~\ref{fig:grid}
ran on the H200 machine after a driver update (615.71.10, CUDA 13.4). Each job times the algorithm and an
autoregressive pass of the same target on the same GPU and inputs, synchronising CUDA at phase boundaries after three
untimed warm-ups; timings exclude loading, tokenization and warm-up. Our Spec-Bench re-runs agree with its leaderboard within 0.2 accepted tokens per target pass
(Figure~\ref{fig:repro}b).

\begin{center}
  \begin{minipage}{\textwidth}
  \centering
  \includegraphics[width=\textwidth]{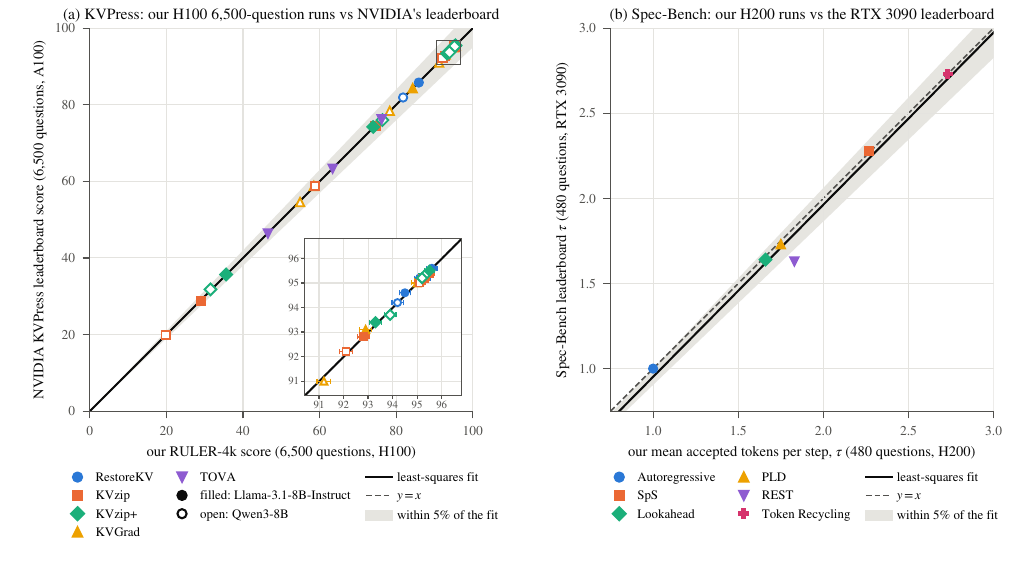}
  \plaincaption{figure}{\textbf{Our runs reproduce the published leaderboards.} Each point is one published algorithm, in (a)
  at one model and compression ratio ($2\times$--$32\times$). (a) Horizontal bars: $\pm1$ bootstrap standard deviation of
  our score over questions (the leaderboard publishes no per-question data); inset: the boxed region, enlarged.
  (b) Vicuna-7B-v1.3 target, greedy decoding ($T=0$); each published value is a mean of three runs.}
  \label{fig:repro}
  \label{app:repro-published}
  \end{minipage}
\end{center}

\subsection{Reproducing our harness}
\label{app:repro-ours}

The harness has three steps, each on its own page with a diagram: preparing the shared problem formulation
(Appendix~\ref{app:repro-selection}) and formalizing a published algorithm (Appendix~\ref{app:repro-stages}), which run
once per field as the offline stage of Section~\ref{sec:method-harness}, and making variants
(Appendix~\ref{app:repro-ablate}), the online stage. In each step an agent follows a prompt of
Appendix~\ref{app:prompts}: \hyperref[app:prompts-design]{Prompt~1} writes an algorithm's proof document from its code,
\hyperref[app:prompts-lean]{Prompt~2} transcribes it into Lean, and \hyperref[app:prompts-ablation]{Prompt~3} derives a
variant and its proof. The harness also needs the benchmark, a Lean~4 project with Mathlib, an agent that reads and
writes files and runs scripts, and GPUs.

\clearpage
\subsubsection{Preparing the shared problem formulation}
\label{app:repro-selection}

Start from the field's benchmark (Figure~\ref{fig:selection}a). Take the algorithms it implements and the papers
that cite them, and let a person keep those that are relevant and can run on the benchmark. Rewrite each as a
self-contained \texttt{code.py} that reproduces the benchmark's reported scores or, for a ported algorithm, some of the
results of its own paper. Ask an LLM harness for one problem that every algorithm could be proved to solve
(Figure~\ref{fig:selection}b), and formalize a random sample of algorithms on it (Appendix~\ref{app:repro-stages}). Keep
the problem and formalize every algorithm on it if a Lean file builds for each of them; otherwise, ask for a new formulation.

\begin{center}
  \begin{minipage}{\textwidth}
  \centering
  \includegraphics{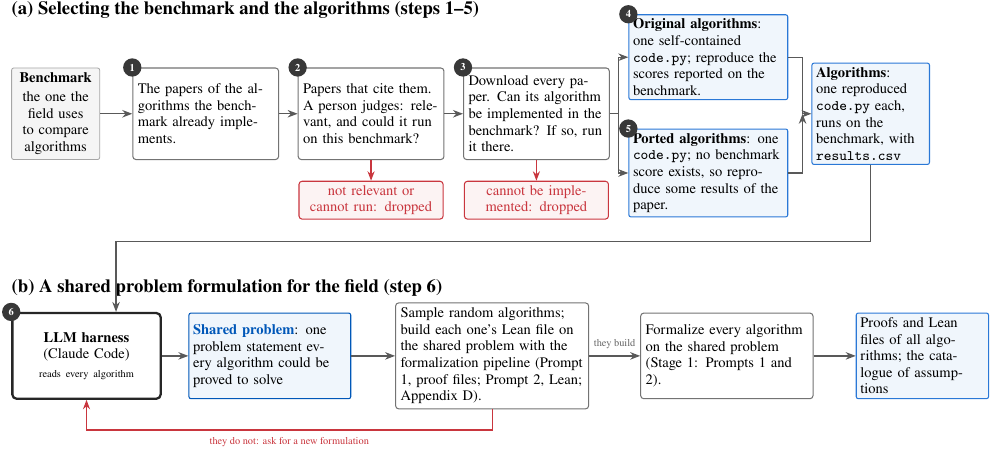}
  \plaincaption{figure}{\textbf{Stage~1 selects a field's reproducible algorithms and finds one problem that all of them can be proved to solve.}
  Black discs number the steps, and boxes with a blue border are outputs. Red marks the algorithms dropped at a step and
  the request for a new formulation when the Lean file of a sampled algorithm does not build.}
  \label{fig:selection}
  \end{minipage}
\end{center}

\clearpage
\subsubsection{Formalizing a published algorithm}
\label{app:repro-stages}

Two agents formalize each algorithm in turn (Figure~\ref{fig:formalization}). The first, with
\hyperref[app:prompts-design]{Prompt~1}, writes the proof document from the algorithm's code, not its paper: every
definition transcribes a line of the code, up to a main theorem that the code meets the shared problem's contract. The
second, with \hyperref[app:prompts-lean]{Prompt~2}, reads \texttt{proof.tex} and \texttt{code.py} and writes
\texttt{axioms.lean}, with the assumptions and the main theorem's statement, and \texttt{proof.lean}, which adds the
proofs. Last, we group the assumptions of all algorithms by the design question they answer into one shared pool.

\begin{center}
  \begin{minipage}{\textwidth}
  \centering
  \includegraphics[width=\textwidth]{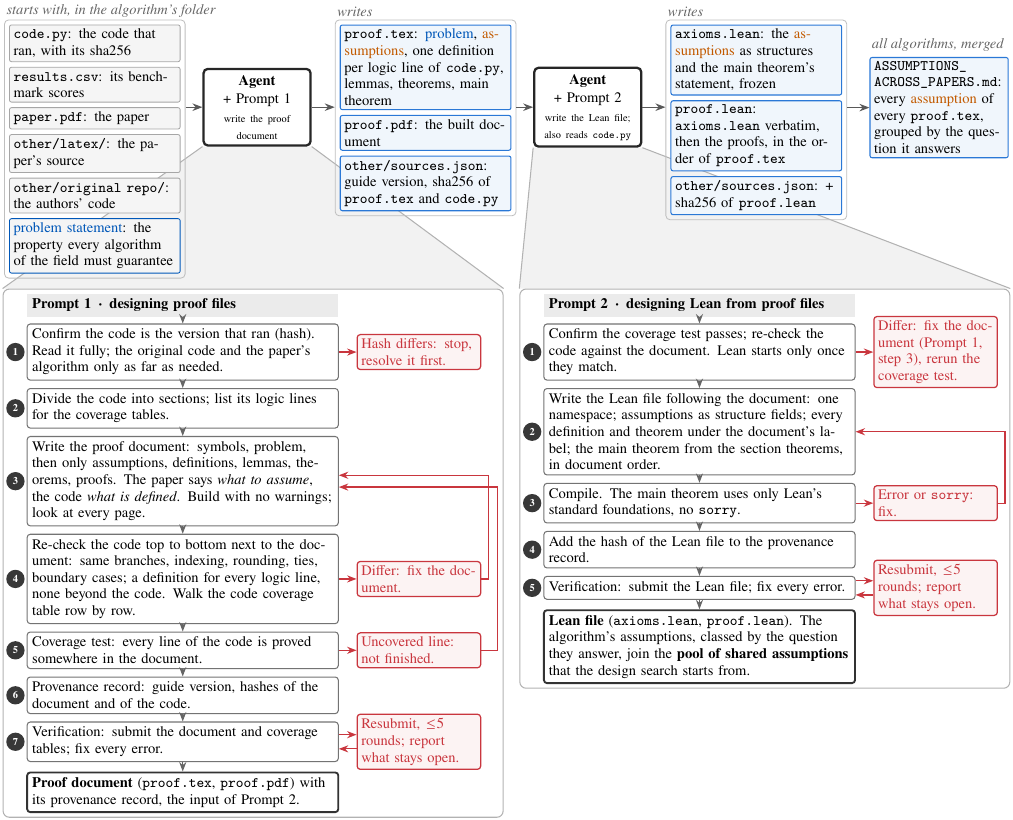}
  \plaincaption{figure}{\textbf{Formalizing a published algorithm with two agents.} Top: the files each agent reads and writes for one
  algorithm, and the pool merged over all algorithms. Bottom: each agent box expanded into the workflow of its prompt
  (Appendices~\ref{app:prompts-design} and~\ref{app:prompts-lean}), with steps numbered as in the prompt. Grey: files a
  step starts from; blue: files it writes; red: checks, with arrows to the step a failure returns to.}
  \label{fig:formalization}
  \end{minipage}
\end{center}

\clearpage
\subsubsection{Making variants}
\label{app:repro-ablate}

One agent makes each variant in a copy of the parent's folder (Figure~\ref{fig:design-search}). Following
\hyperref[app:prompts-ablation]{Prompt~3}, it reads the parent's \texttt{code.py}, \texttt{proof.tex} and scores and the
shared pool, chooses one move on one assumption, often borrowed from the pool, and writes the new assumption and its
hypothesis before the smallest code change. Before any GPU time, the variant must be no known algorithm, reproduce the
parent at one parameter value and differ from it on some input. Its proof edits the parent's only within the impact set
of the changed assumption (changes in red, borrowed text in yellow). We benchmark it in the parent's setting, compare
it with the literature and verify it; its scores guide the choice of the next parent and assumption.

\begin{center}
  \begin{minipage}{\textwidth}
  \centering
  \includegraphics[width=\textwidth]{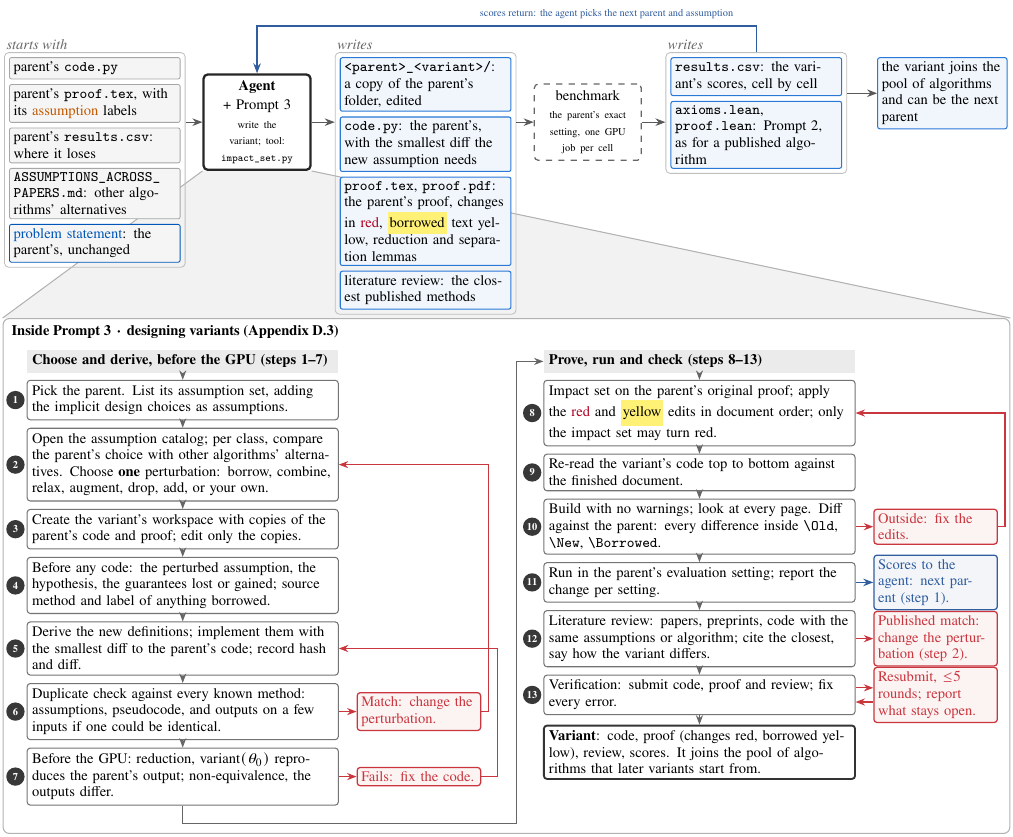}
  \plaincaption{figure}{\textbf{Making one variant.} Top: the files the agent with Prompt~3 (Appendix~\ref{app:prompts-ablation}) reads
  and writes. Bottom: the agent box opened into the workflow of Prompt~3, numbered as in the prompt. Grey: files a
  step starts from; blue: files it writes and the scores returning to the agent; red: checks and the step a failure
  returns to, as in Figure~\ref{fig:formalization}.}
  \label{fig:design-search}
  \end{minipage}
\end{center}

\section{Results in numbers}
\label{app:numbers}
This appendix gives the numbers behind the figures: one subsection per field (KV-cache eviction, speculative
decoding and LLM weight pruning), one for the harness comparison and one for the best variant of every parent;
uncertainties as in Appendix~\ref{app:uncertainty}.

\subsection{KV-cache eviction}
\label{app:numbers-kv}
\subsubsection{Main benchmark}
\label{app:numbers-kv-main}
\begin{figure}[H]
  \centering
  \includegraphics[width=0.5\textwidth]{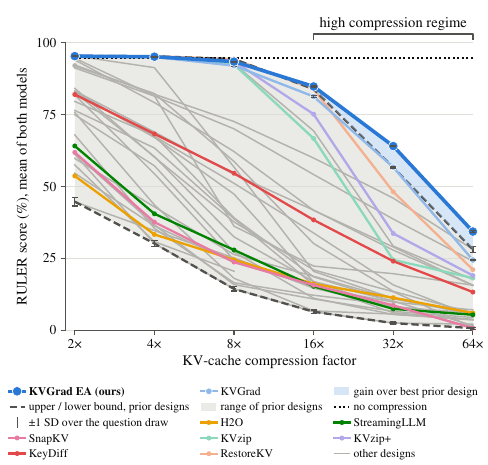}
  \caption{\textbf{KVGrad EA\oursmark{} lies above every prior design at $16\times$--$64\times$ compression, where it gains 6.9
  points on average over its parent KVGrad (3.7 points over $2\times$--$64\times$).} RULER-4K, mean of
  Llama-3.1-8B-Instruct and Qwen3-8B. The 33 query-agnostic prior designs (grey unless the legend names them) are each
  drawn at the largest number of questions both models finished (300 or 6,500); KVGrad EA\oursmark{} and KVGrad use
  6,500. Table~\ref{tab:space-scores} lists the named prior algorithms per model.}
  \label{fig:space}
\end{figure}

We ran 34 published KV-cache eviction algorithms on all twelve RULER-4K cells; RestoreKV GM\oursmark{} (75.7) has a
higher mean than every one of them, the best being KVGrad (74.1). Table~\ref{tab:space-scores} lists the published algorithms Figure~\ref{fig:space} names, with RestoreKV GM\oursmark{},
which has the best mean among them.
It holds the best score in four of the six $16\times$--$64\times$ cells; KVGrad keeps $64\times$ on Llama (26.0
against 25.2) and $32\times$ on Qwen (54.9 against 54.0). At $2\times$--$8\times$ the published algorithms are as
good. Three algorithms drawn only in grey would change a mark: CaM (TOVA) (38.4) and NestedKV (33.8) at $64\times$ on
Llama, and DefensiveKV \citep{feng2026defensivekv} (96.0) at $2\times$ on Llama, all at 300 questions.

\begingroup
\let\floattable\table\let\endfloattable\endtable
\renewenvironment{table}[1][]{\floattable[H]}{\endfloattable}%
\renewenvironment{table*}[1][]{\begin{table}}{\end{table}}%
\renewcommand{\caption}[2][]{\origcaption{\textbf{RULER-4K scores of the published algorithms named in Figure~\ref{fig:space} and of
RestoreKV GM\oursmark{}, per model and compression ratio.} Each algorithm at the largest number of questions both models
finished (Questions; 300 is nested in the 6,500); Mean covers the 12 cells. $\pm$: one bootstrap SD over the question
draw. Bold: best in the column; underlined: second best, at the printed precision.}\origlabel{tab:space-scores}\expandafter\let\expandafter\label\csname @gobble\endcsname}%
\renewcommand{\label}[1]{}%
\begin{table*}[t]
\caption{\textbf{Prior designs of Figure~\ref{fig:space} and RestoreKV GM, per model and compression factor.} RULER-4k scores; Mean covers the 12 cells. Each design is shown at the largest number of questions at which both models finished (Questions; 200 and 300 are nested in the 6,500). $\pm$: one bootstrap SD over the question draw. Bold: best in the column; underlined: second best.}
\label{tab:space-scores}
\centering\small
\setlength{\tabcolsep}{2pt}
\renewcommand{\arraystretch}{1.2}
\resizebox{\textwidth}{!}{%
\begin{tabular}{@{}l@{\hspace{6pt}}r@{\hspace{10pt}}*{6}{c}@{\hspace{10pt}}*{6}{c}@{\hspace{10pt}}c@{}}
\toprule
 & & \multicolumn{6}{c}{Llama-3.1-8B-Instruct} & \multicolumn{6}{c}{Qwen3-8B} & Mean \\
\cmidrule(lr){3-8}\cmidrule(lr){9-14}
 & Questions & $2\times$ & $4\times$ & $8\times$ & $16\times$ & $32\times$ & $64\times$ & $2\times$ & $4\times$ & $8\times$ & $16\times$ & $32\times$ & $64\times$ & \\
\midrule
H2O & {\color{black!55}300} & 54.6~{\scriptsize\color{black!45}$\pm$2.1} & 35.6~{\scriptsize\color{black!45}$\pm$1.6} & 25.5~{\scriptsize\color{black!45}$\pm$1.4} & 14.7~{\scriptsize\color{black!45}$\pm$1.3} & 10.4~{\scriptsize\color{black!45}$\pm$1.4} & 4.2~{\scriptsize\color{black!45}$\pm$1.0} & 52.7~{\scriptsize\color{black!45}$\pm$2.0} & 31.0~{\scriptsize\color{black!45}$\pm$1.7} & 23.5~{\scriptsize\color{black!45}$\pm$1.4} & 17.5~{\scriptsize\color{black!45}$\pm$1.3} & 12.2~{\scriptsize\color{black!45}$\pm$1.2} & 7.7~{\scriptsize\color{black!45}$\pm$1.1} & 24.13~{\scriptsize\color{black!45}$\pm$0.43} \\
SnapKV & {\color{black!55}300} & 70.0~{\scriptsize\color{black!45}$\pm$1.9} & 42.3~{\scriptsize\color{black!45}$\pm$2.0} & 25.7~{\scriptsize\color{black!45}$\pm$1.8} & 15.9~{\scriptsize\color{black!45}$\pm$1.6} & 8.9~{\scriptsize\color{black!45}$\pm$1.3} & 0.0~{\scriptsize\color{black!45}$\pm$0.0} & 53.7~{\scriptsize\color{black!45}$\pm$2.0} & 32.9~{\scriptsize\color{black!45}$\pm$1.9} & 21.7~{\scriptsize\color{black!45}$\pm$1.6} & 15.4~{\scriptsize\color{black!45}$\pm$1.5} & 8.3~{\scriptsize\color{black!45}$\pm$1.2} & 1.4~{\scriptsize\color{black!45}$\pm$0.7} & 24.68~{\scriptsize\color{black!45}$\pm$0.45} \\
StreamingLLM & {\color{black!55}300} & 63.4~{\scriptsize\color{black!45}$\pm$2.4} & 39.4~{\scriptsize\color{black!45}$\pm$2.2} & 28.0~{\scriptsize\color{black!45}$\pm$1.9} & 16.4~{\scriptsize\color{black!45}$\pm$1.6} & 8.4~{\scriptsize\color{black!45}$\pm$1.4} & 5.5~{\scriptsize\color{black!45}$\pm$1.2} & 64.8~{\scriptsize\color{black!45}$\pm$2.4} & 41.6~{\scriptsize\color{black!45}$\pm$2.2} & 27.8~{\scriptsize\color{black!45}$\pm$2.0} & 14.1~{\scriptsize\color{black!45}$\pm$1.6} & 6.4~{\scriptsize\color{black!45}$\pm$1.3} & 5.2~{\scriptsize\color{black!45}$\pm$1.1} & 26.75~{\scriptsize\color{black!45}$\pm$0.52} \\
KeyDiff & {\color{black!55}300} & 86.7~{\scriptsize\color{black!45}$\pm$1.4} & 73.5~{\scriptsize\color{black!45}$\pm$1.4} & 62.8~{\scriptsize\color{black!45}$\pm$1.1} & 52.4~{\scriptsize\color{black!45}$\pm$1.2} & 33.3~{\scriptsize\color{black!45}$\pm$1.4} & 17.8~{\scriptsize\color{black!45}$\pm$1.5} & 77.3~{\scriptsize\color{black!45}$\pm$1.6} & 63.1~{\scriptsize\color{black!45}$\pm$1.4} & 46.4~{\scriptsize\color{black!45}$\pm$1.8} & 24.3~{\scriptsize\color{black!45}$\pm$1.5} & 14.7~{\scriptsize\color{black!45}$\pm$1.1} & 8.7~{\scriptsize\color{black!45}$\pm$0.4} & 46.75~{\scriptsize\color{black!45}$\pm$0.40} \\
KVzip & {\color{black!55}6,500} & \underline{95.5}~{\scriptsize\color{black!45}$\pm$0.2} & \underline{95.3}~{\scriptsize\color{black!45}$\pm$0.2} & 92.8~{\scriptsize\color{black!45}$\pm$0.2} & 74.6~{\scriptsize\color{black!45}$\pm$0.4} & 29.1~{\scriptsize\color{black!45}$\pm$0.4} & 17.7~{\scriptsize\color{black!45}$\pm$0.2} & \underline{95.2}~{\scriptsize\color{black!45}$\pm$0.2} & 95.1~{\scriptsize\color{black!45}$\pm$0.2} & 92.1~{\scriptsize\color{black!45}$\pm$0.3} & 58.8~{\scriptsize\color{black!45}$\pm$0.5} & 19.8~{\scriptsize\color{black!45}$\pm$0.2} & 18.6~{\scriptsize\color{black!45}$\pm$0.2} & 65.38~{\scriptsize\color{black!45}$\pm$0.08} \\
KVzip+ & {\color{black!55}6,500} & \underline{95.5}~{\scriptsize\color{black!45}$\pm$0.2} & 95.2~{\scriptsize\color{black!45}$\pm$0.2} & 93.3~{\scriptsize\color{black!45}$\pm$0.2} & 74.0~{\scriptsize\color{black!45}$\pm$0.3} & 35.6~{\scriptsize\color{black!45}$\pm$0.4} & 18.2~{\scriptsize\color{black!45}$\pm$0.2} & \textbf{95.3}~{\scriptsize\color{black!45}$\pm$0.2} & \underline{95.2}~{\scriptsize\color{black!45}$\pm$0.2} & 93.9~{\scriptsize\color{black!45}$\pm$0.2} & 76.4~{\scriptsize\color{black!45}$\pm$0.4} & 31.5~{\scriptsize\color{black!45}$\pm$0.4} & 19.8~{\scriptsize\color{black!45}$\pm$0.2} & 68.66~{\scriptsize\color{black!45}$\pm$0.08} \\
RestoreKV & {\color{black!55}6,500} & \textbf{95.6}~{\scriptsize\color{black!45}$\pm$0.2} & \underline{95.3}~{\scriptsize\color{black!45}$\pm$0.2} & \textbf{94.5}~{\scriptsize\color{black!45}$\pm$0.2} & \underline{85.9}~{\scriptsize\color{black!45}$\pm$0.3} & 52.8~{\scriptsize\color{black!45}$\pm$0.5} & 19.4~{\scriptsize\color{black!45}$\pm$0.2} & 95.1~{\scriptsize\color{black!45}$\pm$0.2} & \textbf{95.3}~{\scriptsize\color{black!45}$\pm$0.2} & \textbf{94.2}~{\scriptsize\color{black!45}$\pm$0.2} & \underline{81.8}~{\scriptsize\color{black!45}$\pm$0.3} & 43.5~{\scriptsize\color{black!45}$\pm$0.4} & 22.7~{\scriptsize\color{black!45}$\pm$0.3} & 73.01~{\scriptsize\color{black!45}$\pm$0.08} \\
KVGrad & {\color{black!55}6,500} & \underline{95.5}~{\scriptsize\color{black!45}$\pm$0.2} & 95.1~{\scriptsize\color{black!45}$\pm$0.2} & 92.9~{\scriptsize\color{black!45}$\pm$0.2} & 84.3~{\scriptsize\color{black!45}$\pm$0.3} & \underline{58.5}~{\scriptsize\color{black!45}$\pm$0.4} & \textbf{26.0}~{\scriptsize\color{black!45}$\pm$0.3} & \textbf{95.3}~{\scriptsize\color{black!45}$\pm$0.2} & 95.0~{\scriptsize\color{black!45}$\pm$0.2} & 91.2~{\scriptsize\color{black!45}$\pm$0.3} & 78.3~{\scriptsize\color{black!45}$\pm$0.4} & \textbf{54.9}~{\scriptsize\color{black!45}$\pm$0.4} & \underline{22.8}~{\scriptsize\color{black!45}$\pm$0.3} & \underline{74.15}~{\scriptsize\color{black!45}$\pm$0.09} \\
\midrule
RestoreKV GM (ours) & {\color{black!55}6,500} & \underline{95.5}~{\scriptsize\color{black!45}$\pm$0.2} & \textbf{95.4}~{\scriptsize\color{black!45}$\pm$0.2} & \underline{94.4}~{\scriptsize\color{black!45}$\pm$0.2} & \textbf{86.6}~{\scriptsize\color{black!45}$\pm$0.2} & \textbf{64.8}~{\scriptsize\color{black!45}$\pm$0.4} & \underline{25.2}~{\scriptsize\color{black!45}$\pm$0.3} & \underline{95.2}~{\scriptsize\color{black!45}$\pm$0.2} & \textbf{95.3}~{\scriptsize\color{black!45}$\pm$0.2} & \underline{94.1}~{\scriptsize\color{black!45}$\pm$0.2} & \textbf{84.7}~{\scriptsize\color{black!45}$\pm$0.3} & \underline{54.0}~{\scriptsize\color{black!45}$\pm$0.4} & \textbf{23.7}~{\scriptsize\color{black!45}$\pm$0.3} & \textbf{75.74}~{\scriptsize\color{black!45}$\pm$0.08} \\
\bottomrule
\end{tabular}
}
\end{table*}
\endgroup

\subsubsection{Compression ratios and unseen settings}
\label{app:numbers-kv-sweep}
Table~\ref{tab:together} gives the scores behind Figure~\ref{fig:grid}B (Llama) and the same cells for Qwen,
with KVzip and KVzip+ for reference. RestoreKV GM\oursmark{} ties RestoreKV at $2\times$--$8\times$ and gains up to
12.0 points at $32\times$; its mean is 2.73 points higher. Errors: resampling the questions
(Appendix~\ref{app:unc-questions}).

\begingroup
\let\floattable\table\let\endfloattable\endtable
\renewenvironment{table}[1][]{\floattable[H]}{\endfloattable}%
\renewenvironment{table*}[1][]{\begin{table}}{\end{table}}%
\renewcommand{\caption}[2][]{\origcaption{\textbf{Full RULER-4K scores (6,500 questions) of KVzip, KVzip+, RestoreKV and
RestoreKV GM\oursmark{}, per model and compression ratio.} Small figures: RestoreKV GM minus RestoreKV, green where a
two-sided $z$-test gives $p<0.05$, grey otherwise (Appendix~\ref{app:repro-uncertainty}).
Table~\ref{tab:space-scores} gives the error bars.}\origlabel{tab:together}\expandafter\let\expandafter\label\csname @gobble\endcsname}%
\renewcommand{\label}[1]{}%
\begin{table*}[t]
\caption{\textbf{KV-cache eviction: each tested base improves under its best ablation.} Full RULER-4k scores on Llama-3.1-8B-Instruct and Qwen3-8B at $2\times$--$64\times$. Means cover 12 cells; signed differences are ablation minus base. Colors use uncorrected pairwise tests at $p<0.05$.}
\label{tab:kv-ablation}
\centering\small
\setlength{\tabcolsep}{2pt}
\renewcommand{\arraystretch}{1.25}
\resizebox{\textwidth}{!}{%
\begin{tabular}{@{}c@{\hspace{4pt}}l@{\hspace{12pt}}*{6}{c}@{\hspace{12pt}}*{6}{c}@{\hspace{12pt}}c@{}}
\toprule
\multirow{6}{*}{\rotatebox{90}{KV cache eviction}} & & \multicolumn{6}{c}{Llama-3.1-8B-Instruct} & \multicolumn{6}{c}{Qwen3-8B} & Mean \\
\cmidrule(lr){3-8}\cmidrule(lr){9-14}
 & & $2\times$ & $4\times$ & $8\times$ & $16\times$ & $32\times$ & $64\times$ & $2\times$ & $4\times$ & $8\times$ & $16\times$ & $32\times$ & $64\times$ & \\
\cmidrule(lr){2-15}
 & KVzip & 95.5 & 95.3 & 92.8 & 74.6 & 29.1 & 17.7 & 95.2 & 95.1 & 92.1 & 58.8 & 19.8 & 18.6 & 65.38 \\
 & KVzip+ & 95.5 & 95.2 & 93.3 & 74.0 & 35.6 & 18.2 & 95.3 & 95.2 & 93.9 & 76.4 & 31.5 & 19.8 & 68.66 \\
 & RestoreKV & 95.6 & 95.3 & 94.5 & 85.9 & 52.8 & 19.4 & 95.1 & 95.3 & 94.2 & 81.8 & 43.5 & 22.7 & 73.01 \\
\cmidrule(lr){2-15}
 & \textbf{RestoreKV GM (ours)} & 95.5~{\scriptsize \textcolor{black!45}{-0.1}} & 95.4~{\scriptsize \textcolor{black!45}{+0.1}} & 94.4~{\scriptsize \textcolor{black!45}{-0.1}} & 86.6~{\scriptsize \textcolor{black!45}{+0.7}} & 64.8~{\scriptsize \textcolor{green!45!black}{+12.0}} & 25.2~{\scriptsize \textcolor{green!45!black}{+5.8}} & 95.2~{\scriptsize \textcolor{black!45}{+0.1}} & 95.3~{\scriptsize \textcolor{black!45}{0.0}} & 94.1~{\scriptsize \textcolor{black!45}{-0.1}} & 84.7~{\scriptsize \textcolor{green!45!black}{+2.9}} & 54.0~{\scriptsize \textcolor{green!45!black}{+10.5}} & 23.7~{\scriptsize \textcolor{green!45!black}{+1.0}} & \textbf{75.74}~{\scriptsize \textcolor{green!45!black}{+2.73}} \\
\bottomrule
\end{tabular}
}
\end{table*}
\endgroup

In every RULER setting (Table~\ref{tab:settings-factors}) the gain is mostly at $32\times$ and $64\times$ ($+5.8$ to $+12.0$ points, except $+1.0$ for Qwen at
$64\times$), with $2\times$--$8\times$ within $0.3$ points; on LongBench it is spread more evenly and is largest at
$16\times$ ($+3.3$). Errors: resampling the questions (Appendix~\ref{app:unc-questions}).

\begingroup
\let\floattable\table\let\endfloattable\endtable
\renewenvironment{table}[1][]{\floattable[H]}{\endfloattable}%
\renewenvironment{table*}[1][]{\begin{table}}{\end{table}}%
\renewcommand{\caption}[2][]{\origcaption{\textbf{Scores of RestoreKV and RestoreKV GM\oursmark{} at 8K and 16K contexts, on Qwen3-8B and on LongBench, per compression ratio.} Mean is the value in the top of Table~\ref{tab:settings}. $\pm$: one bootstrap SD over the question
draw. Bold: the higher of the two.}\origlabel{tab:settings-factors}\expandafter\let\expandafter\label\csname @gobble\endcsname}%
\renewcommand{\label}[1]{}%
\begin{table*}[t]
\caption{\textbf{Table~\ref{tab:settings} per compression factor.} RestoreKV and RestoreKV group mean in each setting; Mean is the value of Table~\ref{tab:settings}. $\pm$: one bootstrap SD over the question draw. $^{\S}$: from the pilot (second question count). Bold: the higher of the two.}
\label{tab:settings-factors}
\centering\small
\setlength{\tabcolsep}{3.5pt}
\renewcommand{\arraystretch}{1.2}
\begin{tabular}{@{}lrl@{\hspace{10pt}}*{6}{c}@{\hspace{10pt}}c@{}}
\toprule
Setting & Questions & Design & $2\times$ & $4\times$ & $8\times$ & $16\times$ & $32\times$ & $64\times$ & Mean \\
\midrule
\multirow{2}{*}{\begin{tabular}[c]{@{}l@{}}RULER 4K\\Llama\end{tabular}} & \multirow{2}{*}{{\color{black!55}6,500}} & RestoreKV & \textbf{95.6}~{\scriptsize\color{black!45}$\pm$0.2} & 95.3~{\scriptsize\color{black!45}$\pm$0.2} & \textbf{94.5}~{\scriptsize\color{black!45}$\pm$0.2} & 85.9~{\scriptsize\color{black!45}$\pm$0.3} & 52.8~{\scriptsize\color{black!45}$\pm$0.5} & 19.4~{\scriptsize\color{black!45}$\pm$0.2} & 73.9~{\scriptsize\color{black!45}$\pm$0.11} \\
 &  & RestoreKV GM (ours) & 95.5~{\scriptsize\color{black!45}$\pm$0.2} & \textbf{95.4}~{\scriptsize\color{black!45}$\pm$0.2} & 94.4~{\scriptsize\color{black!45}$\pm$0.2} & \textbf{86.6}~{\scriptsize\color{black!45}$\pm$0.2} & \textbf{64.8}~{\scriptsize\color{black!45}$\pm$0.4} & \textbf{25.2}~{\scriptsize\color{black!45}$\pm$0.3} & \textbf{77.0}~{\scriptsize\color{black!45}$\pm$0.12} \\
\midrule
\multirow{2}{*}{\begin{tabular}[c]{@{}l@{}}RULER 8K\\Llama\end{tabular}} & \multirow{2}{*}{{\color{black!55}6,500}} & RestoreKV & \textbf{94.6}~{\scriptsize\color{black!45}$\pm$0.2} & \textbf{93.9}~{\scriptsize\color{black!45}$\pm$0.2} & \textbf{92.9}~{\scriptsize\color{black!45}$\pm$0.2} & 84.1~{\scriptsize\color{black!45}$\pm$0.3} & 61.1~{\scriptsize\color{black!45}$\pm$0.4} & 22.4~{\scriptsize\color{black!45}$\pm$0.3} & 74.8~{\scriptsize\color{black!45}$\pm$0.12} \\
 &  & RestoreKV GM (ours) & 94.5~{\scriptsize\color{black!45}$\pm$0.2} & \textbf{93.9}~{\scriptsize\color{black!45}$\pm$0.2} & 92.8~{\scriptsize\color{black!45}$\pm$0.2} & \textbf{84.6}~{\scriptsize\color{black!45}$\pm$0.3} & \textbf{68.1}~{\scriptsize\color{black!45}$\pm$0.4} & \textbf{33.5}~{\scriptsize\color{black!45}$\pm$0.4} & \textbf{77.9}~{\scriptsize\color{black!45}$\pm$0.12} \\
\midrule
\multirow{2}{*}{\begin{tabular}[c]{@{}l@{}}RULER 16K\\Llama\end{tabular}} & \multirow{2}{*}{{\color{black!55}6,500}} & RestoreKV & \textbf{93.2}~{\scriptsize\color{black!45}$\pm$0.2} & \textbf{92.9}~{\scriptsize\color{black!45}$\pm$0.2} & \textbf{90.0}~{\scriptsize\color{black!45}$\pm$0.3} & 82.1~{\scriptsize\color{black!45}$\pm$0.3} & 59.2~{\scriptsize\color{black!45}$\pm$0.4} & 22.1~{\scriptsize\color{black!45}$\pm$0.3} & 73.3~{\scriptsize\color{black!45}$\pm$0.12} \\
 &  & RestoreKV GM (ours) & \textbf{93.2}~{\scriptsize\color{black!45}$\pm$0.2} & \textbf{92.9}~{\scriptsize\color{black!45}$\pm$0.2} & 89.7~{\scriptsize\color{black!45}$\pm$0.3} & \textbf{83.5}~{\scriptsize\color{black!45}$\pm$0.3} & \textbf{66.4}~{\scriptsize\color{black!45}$\pm$0.4} & \textbf{33.5}~{\scriptsize\color{black!45}$\pm$0.4} & \textbf{76.5}~{\scriptsize\color{black!45}$\pm$0.13} \\
\midrule
\multirow{2}{*}{\begin{tabular}[c]{@{}l@{}}RULER 4K\\Qwen\end{tabular}} & \multirow{2}{*}{{\color{black!55}6,500}} & RestoreKV & 95.1~{\scriptsize\color{black!45}$\pm$0.2} & \textbf{95.3}~{\scriptsize\color{black!45}$\pm$0.2} & \textbf{94.2}~{\scriptsize\color{black!45}$\pm$0.2} & 81.8~{\scriptsize\color{black!45}$\pm$0.3} & 43.5~{\scriptsize\color{black!45}$\pm$0.4} & 22.7~{\scriptsize\color{black!45}$\pm$0.3} & 72.1~{\scriptsize\color{black!45}$\pm$0.12} \\
 &  & RestoreKV GM (ours) & \textbf{95.2}~{\scriptsize\color{black!45}$\pm$0.2} & \textbf{95.3}~{\scriptsize\color{black!45}$\pm$0.2} & 94.1~{\scriptsize\color{black!45}$\pm$0.2} & \textbf{84.7}~{\scriptsize\color{black!45}$\pm$0.3} & \textbf{54.0}~{\scriptsize\color{black!45}$\pm$0.4} & \textbf{23.7}~{\scriptsize\color{black!45}$\pm$0.3} & \textbf{74.5}~{\scriptsize\color{black!45}$\pm$0.12} \\
\midrule
\multirow{2}{*}{\begin{tabular}[c]{@{}l@{}}LongBench\\Llama\end{tabular}} & \multirow{2}{*}{{\color{black!55}3,750}} & RestoreKV & 47.3~{\scriptsize\color{black!45}$\pm$0.5} & 47.7~{\scriptsize\color{black!45}$\pm$0.5} & 45.8~{\scriptsize\color{black!45}$\pm$0.6} & 37.3~{\scriptsize\color{black!45}$\pm$0.6} & 26.8~{\scriptsize\color{black!45}$\pm$0.4} & 21.8~{\scriptsize\color{black!45}$\pm$0.3} & 37.8~{\scriptsize\color{black!45}$\pm$0.20} \\
 &  & RestoreKV GM (ours) & \textbf{47.9}~{\scriptsize\color{black!45}$\pm$0.5} & \textbf{47.9}~{\scriptsize\color{black!45}$\pm$0.5} & \textbf{46.8}~{\scriptsize\color{black!45}$\pm$0.5} & \textbf{40.6}~{\scriptsize\color{black!45}$\pm$0.6} & \textbf{29.0}~{\scriptsize\color{black!45}$\pm$0.5} & \textbf{23.3}~{\scriptsize\color{black!45}$\pm$0.4} & \textbf{39.2}~{\scriptsize\color{black!45}$\pm$0.20} \\
\bottomrule
\end{tabular}
\end{table*}
\endgroup

\subsubsection{Systems cost}
\label{app:numbers-kv-cost}
Table~\ref{tab:ttft-context} gives the time to first token of no eviction, four published algorithms and
RestoreKV GM\oursmark{} across context lengths; Figure~\ref{fig:grid}C plots these times, and
Section~\ref{sec:res-cost} tests it. RestoreKV GM stays within the run-to-run spread of RestoreKV at every context.
KVGrad is the slowest by far: its compression takes 3 to 14\,min at 32K and 64K.

\begingroup
\let\floattable\table\let\endfloattable\endtable
\renewenvironment{table}[1][]{\floattable[H]}{\endfloattable}%
\renewenvironment{table*}[1][]{\begin{table}}{\end{table}}%
\renewcommand{\caption}[2][]{\origcaption{\textbf{Time to first token of no eviction, KVGrad, RestoreKV, KVzip, KVzip+ and
RestoreKV GM\oursmark{} at 4K to 64K tokens.} Seconds, mean $\pm$ SD over seven timing passes of two fixed RULER prompts,
Llama-3.1-8B-Instruct on one H100, no eviction at full cache and every other method at $16\times$ compression; the
first row is the KV-cache size reduction of the compressing methods. Ours in bold.}\origlabel{tab:ttft-context}\expandafter\let\expandafter\label\csname @gobble\endcsname}%
\renewcommand{\label}[1]{}%
\begin{table*}[t]
\caption{\textbf{Time to first token across context lengths.} Seconds, mean $\pm$ SD over seven timing passes.}
\label{tab:ttft-context}
\centering\small
\setlength{\tabcolsep}{5pt}
\renewcommand{\arraystretch}{1.15}
\begin{tabular}{@{}l*{5}{c}@{}}
\toprule
Context & 4K & 8K & 16K & 32K & 64K \\
KV-cache size reduction (GB) & 0.46 & 0.97 & 1.95 & 3.99 & 8.03 \\
\midrule
No eviction (no compression) & 0.23~{\scriptsize\color{black!45}$\pm$0.11} & 0.40~{\scriptsize\color{black!45}$\pm$0.13} & 0.75~{\scriptsize\color{black!45}$\pm$0.04} & 3.48~{\scriptsize\color{black!45}$\pm$2.06} & 5.94~{\scriptsize\color{black!45}$\pm$0.90} \\
\midrule
KVGrad & 3.93~{\scriptsize\color{black!45}$\pm$1.29} & 11.7~{\scriptsize\color{black!45}$\pm$4.36} & 35.1~{\scriptsize\color{black!45}$\pm$3.15} & 175.4~{\scriptsize\color{black!45}$\pm$0.89} & 814.3~{\scriptsize\color{black!45}$\pm$1.97} \\
RestoreKV & 1.81~{\scriptsize\color{black!45}$\pm$0.69} & 3.51~{\scriptsize\color{black!45}$\pm$1.62} & 5.16~{\scriptsize\color{black!45}$\pm$2.04} & 12.2~{\scriptsize\color{black!45}$\pm$1.27} & 33.0~{\scriptsize\color{black!45}$\pm$0.38} \\
KVzip & 1.87~{\scriptsize\color{black!45}$\pm$0.79} & 3.36~{\scriptsize\color{black!45}$\pm$1.20} & 5.13~{\scriptsize\color{black!45}$\pm$1.36} & 11.4~{\scriptsize\color{black!45}$\pm$1.10} & 32.7~{\scriptsize\color{black!45}$\pm$0.31} \\
KVzip+ & 2.54~{\scriptsize\color{black!45}$\pm$1.33} & 3.73~{\scriptsize\color{black!45}$\pm$1.27} & 6.49~{\scriptsize\color{black!45}$\pm$1.88} & 13.3~{\scriptsize\color{black!45}$\pm$0.74} & 36.6~{\scriptsize\color{black!45}$\pm$0.51} \\
\midrule
\textbf{RestoreKV GM (ours)} & \textbf{2.52}~{\scriptsize\color{black!45}$\pm$1.50} & \textbf{3.52}~{\scriptsize\color{black!45}$\pm$1.33} & \textbf{4.94}~{\scriptsize\color{black!45}$\pm$0.80} & \textbf{10.9}~{\scriptsize\color{black!45}$\pm$0.12} & \textbf{33.0}~{\scriptsize\color{black!45}$\pm$0.51} \\
\bottomrule
\end{tabular}
\end{table*}
\endgroup

\subsection{Speculative decoding}
\label{app:numbers-spec}
\label{app:spec-methods}
The bars of Figure~\ref{fig:grid}G are the published decoders Hydra \citep{ankner2024hydra}, EAGLE-2
\citep{li2024eagle2}, Medusa \citep{cai2024medusa}, MCSD \citep{yang2024mcsd}, acceptance feedback \citep{gante2023assisted}, SVIP
\citep{zhang2025svip}, confidence stopping \citep{mamou2024disco,mamou2024dynamic}, AdaEDL \citep{agrawal2024adaedl},
speculative sampling with fixed $K=5$ \citep{leviathan2023fast,chen2023accelerating} and PLD \citep{saxena2023pld}.
SVIP LK\oursmark{} adds a prompt-lookup fallback to SVIP (Algorithm~\ref{alg:svip}). Figure~\ref{fig:grid}G uses the
systems-profile runs (Appendix~\ref{app:spec-timing}).

\subsection{LLM weight pruning}
\label{app:numbers-prune}
Pruner-Zero LG\oursmark{} is the best pruner in every perplexity column, and its lead grows with sparsity.
Tables~\ref{tab:prune-models} and~\ref{tab:prune-ratios} give the cells behind panels D and E of
Figure~\ref{fig:grid} and the bottom of Table~\ref{tab:settings}. The margin over the second
best grows from below 0.01 at 10\,\% to 0.10 at 40\,\% and 0.11 at 50\,\%. On the zero-shot tasks (Table~\ref{tab:prune-zeroshot}) no
pruner is best on every task; Pruner-Zero LG\oursmark{} gains on its parent on three of the seven tasks and on the mean
($+0.3$). Each cell is one run with a fixed calibration seed (Section~\ref{sec:exp-setup}); the LLaMA-30B cells of
ALPS, GBLM-Pruner, Pruner-Zero and Pruner-Zero LG\oursmark{} ran on the first H100 machine and the other
LLaMA-30B cells on a second one (Appendix~\ref{app:repro-prune}).

\subsubsection{Model sizes}
\begingroup
\let\floattable\table\let\endfloattable\endtable
\renewenvironment{table}[1][]{\floattable[H]}{\endfloattable}%
\renewenvironment{table*}[1][]{\begin{table}}{\end{table}}%
\renewcommand{\caption}[2][]{\origcaption{\textbf{WikiText-2 test perplexity of eleven published pruners and Pruner-Zero
LG\oursmark{} on LLaMA 7B, 13B and 30B and LLaMA-2 7B at 50\,\% unstructured sparsity.} The values behind
the bottom of Table~\ref{tab:settings} and Figure~\ref{fig:grid}D; lower is better. Rows: the two
baselines, the eleven published algorithms and \textbf{ours}, Pruner-Zero LG\oursmark{}, in bold. \textbf{Bold} = best,
\underline{underlined} = second best in the column, Dense excluded. One seed-0 run per cell, 128 C4 calibration
windows of 2,048 tokens, fp16, H100.}\origlabel{tab:prune-models}\expandafter\let\expandafter\label\csname @gobble\endcsname}%
\renewcommand{\label}[1]{}%
\begin{table}[t]
\caption{\textbf{Weight pruning across model sizes.} WikiText-2 test perplexity (lower is better) at 50\,\% unstructured sparsity. Rows: the two baselines, the published base methods and \textbf{ours} (bold): the ablation of Pruner-Zero that our harness derived from its proof. One seed-0 run per cell (128 C4 calibration windows of 2048 tokens, fp16, H100). \textbf{Bold} = best, \underline{underlined} = second best in the column (Dense excluded); -- = run not finished.}
\label{tab:prune-models}
\vskip 0.1in
\begin{center}\begin{small}
\setlength{\tabcolsep}{3pt}
\renewcommand{\arraystretch}{1.15}
\makeatletter\@ifundefined{prunebox}{\newsavebox{\prunebox}}{}\makeatother
\sbox{\prunebox}{\begin{tabular}{@{}lrrrr@{}}
\toprule
Method & LLaMA-7B & LLaMA-13B & LLaMA-30B & LLaMA-2-7B \\
\midrule
Dense & 5.68 & 5.09 & 4.10 & 5.47 \\
Magnitude & 17.26 & 20.14 & 7.55 & 16.03 \\
\midrule
SparseGPT & 7.22 & 6.21 & 5.32 & 7.02 \\
E-Sparse & 7.26 & 6.15 & 5.24 & 6.91 \\
GBLM-Pruner & 7.15 & 6.11 & 5.20 & 6.86 \\
Wanda & 7.26 & 6.15 & 5.25 & 6.92 \\
RI & 7.49 & 6.57 & 5.27 & 7.39 \\
RIA & 7.12 & 6.08 & 5.08 & 6.81 \\
OWL + Wanda & 7.22 & 6.08 & 5.25 & 6.86 \\
DSnoT + Wanda & 7.15 & 6.09 & 5.21 & 6.85 \\
Pruner-Zero & \underline{6.89} & \underline{5.93} & \underline{5.02} & \underline{6.71} \\
ALPS & 7.14 & 6.14 & 5.30 & 6.87 \\
Z-Pruner & 7.11 & 6.04 & 5.09 & 6.77 \\
\midrule
\textbf{Pruner-Zero LG} & \textbf{6.78} & \textbf{5.83} & \textbf{4.93} & \textbf{6.67} \\
\bottomrule
\end{tabular}}
\ifdim\wd\prunebox>\columnwidth\resizebox{\columnwidth}{!}{\usebox{\prunebox}}\else\usebox{\prunebox}\fi
\end{small}\end{center}
\vskip -0.1in
\end{table}
\endgroup

\subsubsection{Sparsity ratios}
\begingroup
\let\floattable\table\let\endfloattable\endtable
\renewenvironment{table}[1][]{\floattable[H]}{\endfloattable}%
\renewenvironment{table*}[1][]{\begin{table}}{\end{table}}%
\renewcommand{\caption}[2][]{\origcaption{\textbf{WikiText-2 test perplexity of eleven published pruners and Pruner-Zero
LG\oursmark{} on LLaMA-7B at 10, 20, 30 and 40\,\% unstructured sparsity.} The values of Figure~\ref{fig:grid}E;
lower is better; the 50\,\% column is in Table~\ref{tab:prune-models}. Same rows, runs and marking as
Table~\ref{tab:prune-models}.}\origlabel{tab:prune-ratios}\expandafter\let\expandafter\label\csname @gobble\endcsname}%
\renewcommand{\label}[1]{}%
\begin{table}[t]
\caption{\textbf{Weight pruning across sparsity ratios.} WikiText-2 test perplexity (lower is better) of LLaMA-7B at 10--40\,\% unstructured sparsity (50\,\% is the LLaMA-7B column of Table~\ref{tab:prune-models}). Rows: the two baselines, the published base methods and \textbf{ours} (bold): the ablation of Pruner-Zero that our harness derived from its proof. One seed-0 run per cell (128 C4 calibration windows of 2048 tokens, fp16, H100). \textbf{Bold} = best, \underline{underlined} = second best in the column (Dense excluded); -- = run not finished.}
\label{tab:prune-ratios}
\vskip 0.1in
\begin{center}\begin{small}
\setlength{\tabcolsep}{3pt}
\renewcommand{\arraystretch}{1.15}
\makeatletter\@ifundefined{prunebox}{\newsavebox{\prunebox}}{}\makeatother
\sbox{\prunebox}{\begin{tabular}{@{}lrrrr@{}}
\toprule
Method & 10\,\% & 20\,\% & 30\,\% & 40\,\% \\
\midrule
Dense & 5.68 & 5.68 & 5.68 & 5.68 \\
Magnitude & 5.81 & 6.02 & 6.67 & 8.59 \\
\midrule
SparseGPT & 5.70 & 5.80 & 5.97 & 6.32 \\
E-Sparse & 5.70 & 5.82 & 6.00 & 6.39 \\
GBLM-Pruner & 5.69 & 5.76 & 5.90 & 6.27 \\
Wanda & 5.70 & 5.82 & 6.00 & 6.39 \\
RI & 5.69 & 5.74 & 5.89 & 6.28 \\
RIA & 5.69 & 5.75 & 5.86 & 6.21 \\
OWL + Wanda & 5.71 & 5.81 & 6.02 & 6.39 \\
DSnoT + Wanda & 5.69 & 5.76 & 5.89 & 6.27 \\
Pruner-Zero & \underline{5.68} & \underline{5.73} & \underline{5.85} & \underline{6.18} \\
ALPS & 5.71 & 5.84 & 6.01 & 6.37 \\
Z-Pruner & 5.69 & 5.75 & 5.87 & 6.22 \\
\midrule
\textbf{Pruner-Zero LG} & \textbf{5.68} & \textbf{5.72} & \textbf{5.82} & \textbf{6.08} \\
\bottomrule
\end{tabular}}
\ifdim\wd\prunebox>\columnwidth\resizebox{\columnwidth}{!}{\usebox{\prunebox}}\else\usebox{\prunebox}\fi
\end{small}\end{center}
\vskip -0.1in
\end{table}
\endgroup

\subsubsection{Zero-shot accuracy}
\begingroup
\let\floattable\table\let\endfloattable\endtable
\renewenvironment{table}[1][]{\floattable[H]}{\endfloattable}%
\renewenvironment{table*}[1][]{\begin{table}}{\end{table}}%
\renewcommand{\caption}[2][]{\origcaption{\textbf{Zero-shot accuracy on seven tasks of eleven published pruners and
Pruner-Zero LG\oursmark{} on LLaMA-7B at 50\,\% unstructured sparsity.} lm-eval accuracy (\%, higher is better) on
BoolQ, RTE, HellaSwag, WinoGrande, ARC-easy, ARC-challenge and OpenBookQA, and their mean; same rows, runs and
marking as Table~\ref{tab:prune-models}.}\origlabel{tab:prune-zeroshot}\expandafter\let\expandafter\label\csname @gobble\endcsname}%
\renewcommand{\label}[1]{}%
\begin{table}[t]
\caption{\textbf{Weight pruning, zero-shot accuracy per task.} lm-eval accuracy (\%, higher is better) of LLaMA-7B at 50\,\% unstructured sparsity on BoolQ, RTE, HellaSwag, WinoGrande, ARC-easy, ARC-challenge, OpenBookQA and their mean. Rows: the two baselines, the published base methods and \textbf{ours} (bold): the ablation of Pruner-Zero that our harness derived from its proof. One seed-0 run per cell (128 C4 calibration windows of 2048 tokens, fp16, H100). \textbf{Bold} = best, \underline{underlined} = second best in the column (Dense excluded); -- = run not finished.}
\label{tab:prune-zeroshot}
\vskip 0.1in
\begin{center}\begin{scriptsize}
\setlength{\tabcolsep}{2pt}
\renewcommand{\arraystretch}{1.15}
\makeatletter\@ifundefined{prunebox}{\newsavebox{\prunebox}}{}\makeatother
\sbox{\prunebox}{\begin{tabular}{@{}lrrrrrrrr@{}}
\toprule
Method & BoolQ & RTE & HellaSwag & WinoGr. & ARC-e & ARC-c & OBQA & Mean \\
\midrule
Dense & 76.7 & 67.1 & 57.1 & 70.2 & 75.5 & 42.5 & 33.8 & 60.4 \\
Magnitude & 56.4 & 58.8 & 45.0 & 58.6 & 60.5 & 32.2 & 25.6 & 48.2 \\
\midrule
SparseGPT & 73.5 & 59.2 & 51.7 & 67.6 & 71.2 & 37.7 & \textbf{31.0} & 56.0 \\
E-Sparse & 71.5 & 57.0 & 52.0 & 67.6 & 70.5 & 38.0 & 30.6 & 55.3 \\
GBLM-Pruner & \textbf{74.4} & 59.6 & 52.2 & 67.2 & 71.4 & 38.1 & \textbf{31.0} & \underline{56.3} \\
Wanda & 71.7 & 57.0 & 52.0 & 67.6 & 70.8 & 37.6 & 30.0 & 55.2 \\
RI & 68.3 & 56.3 & \textbf{52.5} & 66.6 & 68.7 & 36.6 & 28.4 & 53.9 \\
RIA & 73.1 & 60.6 & 51.9 & 68.4 & 71.1 & \underline{38.2} & 29.2 & 56.1 \\
OWL + Wanda & 72.6 & 57.8 & 52.0 & \underline{69.5} & 71.2 & 36.9 & 29.2 & 55.6 \\
DSnoT + Wanda & \underline{73.6} & 59.2 & 52.1 & 66.8 & \textbf{72.1} & 37.9 & \underline{30.8} & 56.1 \\
Pruner-Zero & 71.8 & 60.6 & 51.0 & 67.1 & 70.7 & 36.3 & 27.6 & 55.0 \\
ALPS & 72.2 & 49.1 & \underline{52.4} & \textbf{69.5} & 70.4 & 38.0 & \textbf{31.0} & 54.7 \\
Z-Pruner & 72.2 & \textbf{63.5} & 52.0 & 68.3 & \underline{71.4} & \textbf{38.9} & 28.4 & \textbf{56.4} \\
\midrule
\textbf{Pruner-Zero LG} & 71.4 & \underline{61.7} & 50.7 & 66.9 & 70.8 & 35.8 & 29.8 & 55.3 \\
\bottomrule
\end{tabular}}
\ifdim\wd\prunebox>\columnwidth\resizebox{\columnwidth}{!}{\usebox{\prunebox}}\else\usebox{\prunebox}\fi
\end{scriptsize}\end{center}
\vskip -0.1in
\end{table}
\endgroup

\subsection{Harness comparison}
\label{app:numbers-harness}
OpenEvolve has the highest mean from the fifth algorithm on (Table~\ref{tab:harness-passes}). Ours ends 0.4 points
below it, inside its range, with the narrowest range between repeats. Errors: repeating the search
(Appendix~\ref{app:unc-searches}).

\begingroup
\let\floattable\table\let\endfloattable\endtable
\renewenvironment{table}[1][]{\floattable[H]}{\endfloattable}%
\renewenvironment{table*}[1][]{\begin{table}}{\end{table}}%
\renewcommand{\caption}[2][]{\origcaption{\textbf{Running best RULER-4k score after 1 to 10 algorithms of Fable from
scratch, AIDE, OpenEvolve and our harness (the numbers of Figure~\ref{fig:harness}).} Score: the mean of the 12
cells, Llama-3.1-8B-Instruct and Qwen3-8B at $2\times$--$64\times$ (200 questions); each cell is the mean over three repeats per search, with the minimum and maximum across the repeats
beneath it in small grey print. OpenEvolve, AIDE and ours start from RestoreKV, whose own score is excluded. Per
column, bold: the highest mean; underlined: the second highest, ranked before rounding. $^{\star}$: found by our
harness.}\origlabel{tab:harness-passes}\expandafter\let\expandafter\label\csname @gobble\endcsname}%
\renewcommand{\label}[1]{}%
\begin{table*}[t]
\caption{\textbf{Best scores after 1 to 10 designs (the numbers of Figure~\ref{fig:harness}).} Running best of the 12-cell mean RULER-4k score (200 questions; Llama-3.1-8B-Instruct and Qwen3-8B at $2\times$--$64\times$) after each number of designs: mean over three repeats per search, with the minimum and maximum across the repeats beneath it in small grey print. All searches use Fable 5.1 and a ten-design budget; OpenEvolve, AIDE and ours start from RestoreKV, whose own score is excluded. Per column the highest mean, ranked before rounding, is \textbf{bold}, the second highest \underline{underlined}.}
\label{tab:harness-passes}
\vskip 0.15in
\centering\small
\setlength{\tabcolsep}{3pt}
\renewcommand{\arraystretch}{1.25}
\begin{tabular}{@{}l@{\hspace{12pt}}*{10}{c}@{}}
\toprule
 & \multicolumn{10}{c}{Designs evaluated} \\
\cmidrule(l){2-11}
 & 1 & 2 & 3 & 4 & 5 & 6 & 7 & 8 & 9 & 10 \\
\midrule
Fable, from scratch & 49.3 & 54.1 & 62.5 & 63.1 & 64.7 & 66.4 & 68.2 & 69.4 & 71.4 & 71.4 \\
 & {\scriptsize\textcolor{black!45}{44.0--58.8}} & {\scriptsize\textcolor{black!45}{44.0--59.2}} & {\scriptsize\textcolor{black!45}{58.6--65.8}} & {\scriptsize\textcolor{black!45}{59.0--67.2}} & {\scriptsize\textcolor{black!45}{60.3--67.2}} & {\scriptsize\textcolor{black!45}{63.4--68.1}} & {\scriptsize\textcolor{black!45}{67.6--68.8}} & {\scriptsize\textcolor{black!45}{67.6--71.9}} & {\scriptsize\textcolor{black!45}{68.8--73.0}} & {\scriptsize\textcolor{black!45}{68.8--73.0}} \\[2pt]
AIDE & \textbf{72.8} & \underline{75.4} & \underline{76.3} & \textbf{76.6} & \underline{76.9} & \underline{77.1} & \underline{78.2} & \underline{78.2} & 78.2 & 78.5 \\
 & {\scriptsize\textcolor{black!45}{70.4--77.7}} & {\scriptsize\textcolor{black!45}{70.4--78.2}} & {\scriptsize\textcolor{black!45}{72.7--78.2}} & {\scriptsize\textcolor{black!45}{73.0--78.8}} & {\scriptsize\textcolor{black!45}{73.2--79.2}} & {\scriptsize\textcolor{black!45}{73.2--79.9}} & {\scriptsize\textcolor{black!45}{73.6--80.8}} & {\scriptsize\textcolor{black!45}{73.6--80.8}} & {\scriptsize\textcolor{black!45}{73.6--80.8}} & {\scriptsize\textcolor{black!45}{73.8--80.9}} \\[2pt]
OpenEvolve & \underline{70.7} & \textbf{76.1} & \textbf{76.6} & \underline{76.6} & \textbf{77.3} & \textbf{78.7} & \textbf{78.8} & \textbf{79.5} & \textbf{79.7} & \textbf{79.7} \\
 & {\scriptsize\textcolor{black!45}{66.5--72.7}} & {\scriptsize\textcolor{black!45}{72.7--77.9}} & {\scriptsize\textcolor{black!45}{74.1--77.9}} & {\scriptsize\textcolor{black!45}{74.1--77.9}} & {\scriptsize\textcolor{black!45}{76.2--77.9}} & {\scriptsize\textcolor{black!45}{77.7--80.4}} & {\scriptsize\textcolor{black!45}{77.9--80.4}} & {\scriptsize\textcolor{black!45}{78.5--81.3}} & {\scriptsize\textcolor{black!45}{78.8--81.3}} & {\scriptsize\textcolor{black!45}{78.8--81.3}} \\[2pt]
\midrule
Ours$^{\star}$ & 69.6 & 70.7 & 71.0 & 72.4 & 72.9 & 75.1 & 77.2 & 77.5 & \underline{79.3} & \underline{79.3} \\
 & {\scriptsize\textcolor{black!45}{66.8--71.1}} & {\scriptsize\textcolor{black!45}{70.1--71.1}} & {\scriptsize\textcolor{black!45}{71.0--71.1}} & {\scriptsize\textcolor{black!45}{71.2--74.8}} & {\scriptsize\textcolor{black!45}{71.2--74.8}} & {\scriptsize\textcolor{black!45}{75.0--75.2}} & {\scriptsize\textcolor{black!45}{77.1--77.2}} & {\scriptsize\textcolor{black!45}{77.2--77.8}} & {\scriptsize\textcolor{black!45}{78.8--80.1}} & {\scriptsize\textcolor{black!45}{78.8--80.1}} \\[2pt]
\bottomrule
\end{tabular}
\end{table*}
\endgroup

\subsection{Best variant of every parent}
\label{app:numbers-best-variant}
Figure~\ref{fig:best-variant-gain} shows the seven parents with the largest gain per field;
Table~\ref{tab:best-variant-all} lists all \the\numexpr\BTPkvFam+\BTPspecFam+\BTPpruneFam\relax{} parents. Every
parent has a variant that scores higher on its field's search benchmark, although the smallest gains lie within the
benchmark noise.

\begin{table}[H]
  \caption{\textbf{The best variant of each of the \the\numexpr\BTPkvFam+\BTPspecFam+\BTPpruneFam\relax{} parent
  algorithms on its field's search benchmark.} Variants: how many the harness derived from the parent; Better: how many
  score above it. Gain: best variant minus parent, in RULER-4K points (200 questions, mean of both models and six
  compression ratios), WikiText-2 perplexity of LLaMA-7B at 50\,\% sparsity (sign flipped, so positive is better) and
  speedup over autoregressive decoding (Vicuna-7B, mean of five temperatures); Gain (\%): relative to the parent.}
  \label{tab:best-variant-all}
  \centering\small
  \begin{tabular}{@{}llrrrr@{}}
\toprule
Parent & Best variant & Variants & Better & Gain & Gain (\%) \\
\midrule
\multicolumn{6}{@{}l}{\textit{KV-cache eviction} (gain in RULER points)} \\
PyramidKV & PyramidKV lastquery & 6 & 4 & +12.43 & +54.8 \\
KNorm & KNorm value norm & 6 & 5 & +8.41 & +44.0 \\
TOVA & TOVA pooling & 6 & 2 & +5.16 & +15.5 \\
SnapKV & SnapKV head mean & 6 & 6 & +2.99 & +12.3 \\
KVzip+ & KVzip+ residual norm & 40 & 22 & +4.45 & +6.5 \\
KVGrad & KVGrad exact attention & 40 & 13 & +3.57 & +4.8 \\
RestoreKV & RestoreKV GM (ours) & 36 & 16 & +3.52 & +4.8 \\
EchoPress & EchoPress group mean & 6 & 3 & +2.91 & +4.4 \\
KVzip & KVzip group mean & 6 & 3 & +2.27 & +3.5 \\
Expected Attention & Expected Attention group max & 6 & 2 & +0.46 & +1.0 \\
CaM+TOVA & CaM+TOVA head max & 6 & 1 & +0.61 & +1.0 \\
\midrule
\multicolumn{6}{@{}l}{\textit{Weight pruning} (gain in perplexity, lower perplexity counted as positive)} \\
RI & RI act inside & 10 & 5 & +0.293 & +3.9 \\
Pruner-Zero & Pruner-Zero LG (ours) & 10 & 2 & +0.103 & +1.5 \\
RIA & RIA owl alloc & 10 & 5 & +0.093 & +1.3 \\
Wanda & Wanda colnorm & 10 & 3 & +0.072 & +1.0 \\
ALPS & ALPS diag shrink & 10 & 5 & +0.066 & +0.9 \\
DSnoT (Wanda) & DSnoT (Wanda) down proj & 10 & 2 & +0.039 & +0.6 \\
Z-Pruner & Z-Pruner alpha half & 10 & 5 & +0.019 & +0.3 \\
\midrule
\multicolumn{6}{@{}l}{\textit{Speculative decoding} (gain in speedup)} \\
SVIP & SVIP LK (ours) & 7 & 3 & +0.216 & +14.5 \\
Tree Monte Carlo & Tree Monte Carlo mixture target & 7 & 3 & +0.162 & +12.2 \\
Acceptance-feedback $\gamma$ & Acceptance-feedback $\gamma$ cumconf cut & 7 & 3 & +0.082 & +5.5 \\
Confidence stop & Confidence stop discounted chain & 7 & 6 & +0.078 & +5.1 \\
AdaEDL & AdaEDL maxconf & 7 & 3 & +0.074 & +5.0 \\
SpS & SpS prompt lookup & 7 & 1 & +0.044 & +3.2 \\
SpS fixed k & SpS fixed k lookup tail & 7 & 1 & +0.046 & +3.1 \\
Block verification & Block verification margin stop & 7 & 1 & +0.014 & +1.0 \\
\bottomrule
\end{tabular}

\end{table}

\section{Algorithm sketches}
\label{app:algorithms}
Pseudocode of the three variants the paper discusses in detail, one per field: RestoreKV GM\oursmark{} (KV-cache
eviction), Pruner-Zero LG\oursmark{} (weight pruning) and SVIP LK\oursmark{} (speculative decoding).

\subsection{KV-cache eviction}
\label{app:algorithms-kv}
\subsubsection{RestoreKV GM\oursmark{}}
\begingroup
\renewenvironment{algorithm*}[1][]{\begin{algorithm}[H]}{\end{algorithm}}%
\renewcommand{\caption}[2][]{\origcaption{\textbf{RestoreKV GM\oursmark{}.}}\origlabel{alg:restorekv}\expandafter\let\expandafter\label\csname @gobble\endcsname}%
\renewcommand{\label}[1]{}%
\begin{algorithm*}[t]
\caption{\textbf{RestoreKV group mean.} Pseudocode of \emph{code.py} in \emph{kv cache eviction\slash algorithms\slash best ablations\slash RestoreKV\_group\_mean\slash }.}
\label{alg:restorekv}
\small
\renewcommand{\algorithmiccomment}[1]{\hfill{\normalfont\color{black!60}$\triangleright$~#1}}
\begin{algorithmic}[1]
\REQUIRE model with $L$ layers, $H$ KV heads per layer, $G$ query heads per KV head, head dimension $d$; context
  tokens $x_1, \dots, x_n$, of which the first $p$ are the chat-template prefix; eviction fraction $r \in [0, 1)$;
  $n_s = 4$ sink positions; replay chunk size $c = 2048$; $n_E = 8$ trained restore embeddings $E$ and LoRA restore
  adapter $\Delta$
\REQUIRE group aggregation $\in \{\mathrm{mean}, \mathrm{max}\}$, set to $\mathrm{mean}$ ($\mathrm{max}$ recovers
  RestoreKV)
\ENSURE for $r > 0$, the KV cache with $n_E$ restore pairs added per layer and KV head and the
  $\lfloor \min(1,r + n_E/n)\,LHn \rfloor$ lowest-scored context pairs masked out
\STATE \textbf{Prefill:} run the model on $x_1, \dots, x_n$ and keep keys $k^{l}_{h,i}$ and values $v^{l}_{h,i}$ for
  every layer $l$, KV head $h$ and position $i \le n$
\STATE \textbf{if} $r = 0$, \textbf{return} the cache unchanged \COMMENT{no replay, no restore pass}
\STATE $S^{l}_{h,i} \leftarrow 0$ for all $l, h, i$ \COMMENT{prefix positions are never scored; those past the sinks keep $0$}
\STATE split $x_{p+1}, \dots, x_n$ into consecutive chunks $C_1, \dots, C_m$ of $c$ tokens each (the last may be shorter)
\FOR{$j = 1, \dots, m$}
  \STATE $P \leftarrow$ the tokens of ``\texttt{\textbackslash n\textbackslash n}Repeat the previous context exactly.''
    if $j = 1$; else the tokens of ``\texttt{\textbackslash n\textbackslash n}Repeat the part of the previous context
    exactly, starting with'' followed by the last 8 tokens of $C_{j-1}$
  \STATE replay $R \leftarrow P$, the chat-template text that follows a user message (end of the user turn, assistant
    header), $C_j$; its $q$ tokens take positions $n+1, \dots, n+q$
  \STATE run the model on $R$ on top of the cache; after the attention sublayer of every layer $l$ has run (its output unchanged):
  \STATE \quad $Q_{hg,t} \leftarrow$ query of replay token $t$ for query head $(h, g)$, rotary-encoded at position $n + t$
  \STATE \quad $\mathcal{K} \leftarrow$ the first $\min(n_s, |\text{positions before } C_j|)$ cached positions, the
    positions of $C_j$ and the $q$ replay positions, whose keys the replay tokens have just written
  \STATE \quad $a_{hg}(t, i) \leftarrow \mathrm{softmax}_{i \in \mathcal{K}}\bigl(Q_{hg,t} \cdot k^{l}_{h,i} /
    \sqrt{d}\bigr)$, replay positions masked causally
  \STATE \quad keep $a_{hg}(t, i)$ only for the positions $i$ of $C_j$
  \STATE \quad $m_{hg}(i) \leftarrow \max_{t}\ a_{hg}(t, i)$ for $i$ in $C_j$ \ (per query head)
  \STATE \quad $S^{l}_{h,i} \leftarrow \tfrac{1}{G} \sum_{g=1}^{G} m_{hg}(i)$ for $i$ in $C_j$ \ (mean over the $G$ query heads of the group)
  \STATE \quad drop the replay tokens' keys and values from the cache (it holds the $n$ context pairs again)
\ENDFOR
\STATE \textbf{Restore pass:} switch on $\Delta$, run the model on $E$ at positions $n+1, \dots, n+n_E$ on top of the
  full cache (no scoring), append the $n_E$ new pairs per layer and KV head to the cache, switch $\Delta$ off
\STATE $r' \leftarrow \min(1, r + n_E/n)$ \COMMENT{the restore pairs use part of the budget}
\STATE $S^{l}_{h,i} \leftarrow \max S + 1$ for $i \le n_s$ \COMMENT{rank sinks above other context pairs}
\STATE $N \leftarrow \lfloor r' \cdot LHn \rfloor$; $\mathcal{P} \leftarrow$ the $N$ triples $(l, h, i)$ with the
  lowest $S^{l}_{h,i}$, ranked jointly over all layers, heads and positions
\FOR{every layer $l$}
  \STATE $N_l \leftarrow$ the number of triples of $\mathcal{P}$ in layer $l$
  \STATE mask the $N_l$ pairs $(h, i)$ of layer $l$ with the lowest $S^{l}_{h,i}$: later attention ignores them
\ENDFOR
\STATE \textbf{return} the cache
\end{algorithmic}
\end{algorithm*}
\endgroup

\clearpage

\subsection{LLM weight pruning}
\label{app:algorithms-prune}
\subsubsection{Pruner-Zero LG\oursmark{}}
\begingroup
\renewenvironment{algorithm*}[1][]{\begin{algorithm}[H]}{\end{algorithm}}%
\renewcommand{\caption}[2][]{\origcaption{\textbf{Pruner-Zero LG\oursmark{}.}}\origlabel{alg:prunerzero}\expandafter\let\expandafter\label\csname @gobble\endcsname}%
\renewcommand{\label}[1]{}%
\begin{algorithm*}[t]
\caption{\textbf{Pruner-Zero LG.} Pseudocode of \emph{code.py} in \emph{llm weight prunning\slash algorithms\slash best
ablations\slash prunerzero\_layer\_group\slash }.}
\label{alg:prunerzero}
\small
\renewcommand{\algorithmiccomment}[1]{\hfill{\normalfont\color{black!60}$\triangleright$~#1}}
\begin{algorithmic}[1]
\REQUIRE dense model in fp16 with $L$ decoder layers; the weight matrices $W^{l}_{\lambda}$ of every linear $\lambda$ of
  layer $l$ (query, key, value, output, gate, up and down projections), each with $R$ output rows and $C$ input columns;
  sparsity $s \in [0, 1)$ (unstructured) or an $N{:}M$ pattern; $n_g = 128$ gradient windows and $n_c = 128$
  calibration windows of $T = 2048$ tokens, seed $0$; gradient scale $\alpha = 100$
\REQUIRE comparison group $=$ matrix: a weight matrix loses its $\lfloor s R C \rfloor$ lowest-scored weights as a whole
  (\texttt{layer} in the code; each output row losing its $\lfloor s C \rfloor$ lowest, \texttt{row}, recovers Pruner-Zero)
\ENSURE the model with, in every decoder linear, $\lfloor s R C \rfloor$ weights set to $0$ (or $N$ of every $M$
  consecutive inputs of every row); embeddings, norms and the output head unchanged
\STATE \textbf{Gradient pass:} draw $n_g$ random windows of $T$ tokens from the WikiText-2 training text; for every window
  $w$, take the gradient $g^{l}_{\lambda,w}$ of the dense model's language-model loss on $w$ with respect to every
  $W^{l}_{\lambda}$
\STATE $G^{l}_{\lambda} \leftarrow \sqrt{\sum_{w} (\alpha\, g^{l}_{\lambda,w})^2}$ elementwise, rounded to fp16
  \COMMENT{one $R \times C$ gradient norm per matrix, computed once per model}
\STATE \textbf{Calibration inputs:} draw $n_c$ random windows of $T$ tokens from the C4 training text and record the
  inputs $X_1$ of decoder layer $1$ for all of them
\FOR{$l = 1, \dots, L$}
  \FOR{every weight matrix $W = W^{l}_{\lambda}$ of layer $l$, in module order}
    \STATE $A_{ij} \leftarrow |W_{ij}|^2$ \ ($|W|$ in fp16, squared in fp32) \COMMENT{the magnitude term of the evolved metric}
    \STATE $G \leftarrow G^{l}_{\lambda}$ in fp32; \ $\tilde G_{ij} \leftarrow (G_{ij} - \min G) / (\max G - \min G)$,
      the minimum and maximum over the whole matrix
    \STATE $S_{ij} \leftarrow A_{ij}\, \tilde G_{ij}$ \COMMENT{Pruner-Zero's evolved score $|(|W| \times |W|)| \times
      \mathrm{mms}(G)$; no activations}
    \IF{unstructured}
      \STATE $k \leftarrow \lfloor s R C \rfloor$; \ $\mathcal{P} \leftarrow$ the $k$ positions $(i, j)$ with the lowest
        $S_{ij}$, ranked jointly over the whole matrix, ties to the earlier position in row-major order
    \ELSE
      \STATE $\mathcal{P} \leftarrow$ in every row $i$ and every block of $M$ consecutive columns, the $N$ positions
        with the lowest $S_{ij}$
    \ENDIF
    \STATE $W_{ij} \leftarrow 0$ for $(i, j) \in \mathcal{P}$ \COMMENT{the remaining weights are not updated}
  \ENDFOR
  \STATE $X_{l+1} \leftarrow$ the outputs of the pruned layer $l$ on $X_l$ \COMMENT{layers are pruned in order; the
    score reads no activations, so the pass only feeds the next layer}
\ENDFOR
\STATE \textbf{return} the pruned model
\end{algorithmic}
\end{algorithm*}
\endgroup

\clearpage

\subsection{Speculative decoding}
\label{app:algorithms-spec}
\subsubsection{SVIP LK\oursmark{}}
\begingroup
\renewenvironment{algorithm*}[1][]{\begin{algorithm}[H]}{\end{algorithm}}%
\renewcommand{\caption}[2][]{\origcaption{\textbf{SVIP LK\oursmark{}.}}\origlabel{alg:svip}\expandafter\let\expandafter\label\csname @gobble\endcsname}%
\renewcommand{\label}[1]{}%
\begin{algorithm*}[t]
\caption{\textbf{SVIP lookup fallback.} Pseudocode of \emph{code.py} in \emph{speculative decoding\slash algorithms\slash best ablations\slash svip\_lookup\_fallback\slash }.}
\label{alg:svip}
\small
\renewcommand{\algorithmiccomment}[1]{\hfill{\normalfont\color{black!60}$\triangleright$~#1}}
\begin{algorithmic}[1]
\REQUIRE target model $\mathcal M_\theta$ (logits $z(u)$ for the token after a sequence $u$) and drafter $\mathcal M_q$
  (logits $z_q(u)$), each with a KV cache; prompt $x_1, \dots, x_n$; cap $N = 1024$ on new tokens; EOS token $e$; draft
  cap $D_{\max} = 40$ per step; entropy threshold $h = 0.3$ and constant $c_H = 0.15$ of the stop test; setting
  $\sigma$: greedy ($T = 0$, the Spec-Bench protocol) or sample with temperature $\tau$ and top-$k$ $\kappa = 50$
  ($w_{\tau,\kappa}(v)$: softmax of $v / \tau$ restricted to its $\kappa$ largest entries)
\REQUIRE lookup fallback: at most $\Lambda = 10$ lookup tokens per step (lookup\_tokens; $\Lambda = 0$ recovers SVIP), largest $n$-gram size $\nu = 3$ of the lookup (lookup\_ngram)
\ENSURE the sequence $s$ (prompt followed by the emitted tokens), the number of emitted tokens $|s| - n$, the number
  of steps $J$ (target forward passes after the prefill) and the tokens emitted per step $\ell_1, \dots, \ell_J$; at
  $T = 0$ the emitted tokens are the target's greedy continuation of the prompt
\STATE \textbf{Prefill:} $s \leftarrow x_{1:n}$; $\mathcal C_\theta, \mathcal C_q \leftarrow$ caches of $x_{1:n-1}$ by
  one target and one drafter pass; $c \leftarrow 0$; $j \leftarrow 0$ \COMMENT{$c$: upstream counter}
\REPEAT
  \STATE $j \leftarrow j + 1$; \ $m \leftarrow \min(D_{\max},\, N - c - 1)$ \COMMENT{draft budget of the step}
  \STATE \textbf{Draft:} $d \leftarrow ()$; $i \leftarrow 0$
  \WHILE{$i < m$}
    \STATE $i \leftarrow i + 1$; \ $v \leftarrow z_q(s \cdot d_{1:i-1})$ by one drafter pass over the tokens of
      $s \cdot d_{1:i-1}$ not yet in $\mathcal C_q$; \ $\mathcal C_q \leftarrow$ cache of $s \cdot d_{1:i-1}$
    \STATE $H \leftarrow -\sum_x \pi(x) \ln \pi(x)$ for $\pi = \mathrm{softmax}(v)$ (nats; raw logits, no $\tau$ or
      top-$k$) \COMMENT{the drafter's own uncertainty about $d_i$}
    \STATE $d_i \leftarrow \arg\max v$ \ (sample: $q_i \leftarrow w_{\tau,\kappa}(v)$, draw $d_i \sim q_i$); \
      $d \leftarrow d \cdot d_i$
    \STATE \textbf{if} $i = m$ \textbf{or} $d_i = e$\textbf{: break} \COMMENT{count cap, EOS}
    \IF{$\sqrt{c_H H} > h$}
      \STATE $b \leftarrow \min(\Lambda,\, m - i)$ \ (the room left under the budget; $b \ge 1$ unless
        $\Lambda = 0$)
      \STATE \textbf{if} $b \ge 1$: \textbf{lookup} in $u \leftarrow s \cdot d$, for $\nu' = \nu, \nu - 1,
        \dots, 1$ in turn: $a^* \leftarrow$ the earliest $a$ such that $u_{a:a+\nu'-1}$ equals the suffix
        $u_{|u|-\nu'+1:|u|}$ (an earlier copy of the last $\nu'$ tokens), with $a \le |u| - 2\nu'$ and
        $a + \nu' + \Lambda - 1 \le |u|$
      \STATE \quad $w \leftarrow u_{a^*+\nu' : a^*+\nu'+\Lambda-1}$ at the first $\nu'$ with such an $a^*$
        (the $\Lambda$ tokens that followed the copy); \ $w \leftarrow ()$ if no size has one
      \STATE \quad $w \leftarrow$ the first $b$ tokens of $w$, cut after their first $e$; \ $d \leftarrow d
        \cdot w$ \ (no drafter pass for these tokens; sample: $q_{i'} \leftarrow \delta_{d_{i'}}$, the point mass, for
        $i < i' \le |d|$)
      \STATE \textbf{break}
    \ENDIF
  \ENDWHILE
  \STATE $k \leftarrow |d|$ \ ($i$ drafter tokens followed by $k - i$ lookup tokens)
  \STATE \textbf{Verify:} $(z_0, \dots, z_k) \leftarrow$ the target logits at positions $|s|, \dots, |s| + k$ by one
    target pass over $\mathrm{last}(s) \cdot d$ on top of $\mathcal C_\theta$; \ $\mathcal C_\theta \leftarrow$ cache
    of $s \cdot d$
  \STATE \textbf{Accept:} $t_a \leftarrow \arg\max z_a$ for $a = 0, \dots, k$; \ $r \leftarrow \max\{a : d_b =
    t_{b-1} \text{ for all } b \le a\}$; \ $y \leftarrow (t_0, \dots, t_r)$
  \STATE \quad (sample, $k \ge 1$: $p_a \leftarrow w_{\tau,\kappa}(z_a)$; $r \leftarrow$ the number of leading coins $\xi_b <
    p_{b-1}(d_b) / q_b(d_b)$, $\xi_b$ uniform on $[0, 1)$; $y \leftarrow d_{1:r} \cdot c^*$ with $c^* \sim
    \mathrm{norm}\bigl((p_r - q_{r+1})_+\bigr)$ if $r < k$, else $c^* \sim p_k$)
  \STATE \textbf{if} $\mathrm{last}(s \cdot d) = e$ \textbf{and} $r = k$: $y \leftarrow y_{1:|y|-1}$ \COMMENT{no bonus
    token after an accepted EOS draft}
  \STATE $\ell_j \leftarrow |y|$; \ $s \leftarrow s \cdot y$ \COMMENT{emit the block of the step}
  \STATE $\mathcal C_\theta \leftarrow \mathcal C_\theta[1{:}|s|-1]$; \ $\mathcal C_q \leftarrow \mathcal
    C_q[1{:}\min(|\mathcal C_q|, |s|-1)]$ \COMMENT{both caches cropped to the new length $- 1$}
  \STATE $c \leftarrow c + r + 1$ \COMMENT{upstream counter: counts the bonus token even when it was dropped}
\UNTIL{$c \ge N$ \textbf{or} $\mathrm{last}(s) = e$}
\STATE \textbf{return} $s$, $|s| - n$, $J = j$, $(\ell_1, \dots, \ell_J)$
\end{algorithmic}
\end{algorithm*}
\endgroup

\makeatletter\setlength{\@fptop}{0pt}\setlength{\@fpbot}{0pt plus 1fil}\setlength{\@dblfptop}{0pt}%
\setlength{\@dblfpbot}{0pt plus 1fil}\makeatother
\clearpage

\section{Prompts for models}
\label{app:prompts}

The model works from three prompts, one per step, each printed on its own page. We give them here in a
setup-independent form: we leave out file layout, tooling and algorithm-specific examples, so the prompts apply to any
algorithm with a stated specification and can be adapted to another code base. The prompts are printed as plain text: their instructions on the typesetting of
the documents are left out.

Each prompt covers one step of the method. Appendix~\ref{app:prompts-design} writes the proof document of an algorithm, built mainly on its benchmarked code
(Section~\ref{sec:method-designproof}). Appendix~\ref{app:prompts-lean} writes the Lean proof from a finished proof
document. Appendix~\ref{app:prompts-ablation} derives a variant from a parent's proof by changing its
assumptions, with every change marked and every borrowed statement attributed.
Figures~\ref{fig:formalization} and~\ref{fig:design-search} (Appendix~\ref{app:repro-ours}) show where each prompt
is called and the workflow it sets out.

\clearpage
\subsection{Designing proof files}
\label{app:prompts-design}
\begin{promptlisting}{Prompt 1: writing the proof document of a design}
How to write a proof document: the code of one design as mathematics

A proof document takes one design (a method with its paper and the code that was benchmarked) and produces an exact mathematical description of that code, as it actually ran, stated only as assumptions, definitions, lemmas, theorems and proofs, ending in a theorem that running this code meets the design's specification: the one property the design is required to guarantee. That property is the only claim proved. Nothing about optimality, nothing about quality, nothing beyond it. The finished document is later formalized in Lean, following its proof structure.

The point of the whole exercise. The benchmarked code is the code for which we have performance numbers. The document is about that code: it shows that this code, when run, provably meets its specification, the way verified software is shown to meet a specification. Therefore the single most important requirement is that the document represents the code exactly, so that every definition is a faithful transcription of a line of the code that runs, and every design decision the code makes is explicit in the document. If the description is exact, a Lean proof about the description is a proof about the code. If it is not, the Lean proof certifies nothing. Everything else in this guide serves that requirement.

Inputs and outputs

Inputs, gathered for the design before a proof is started:

- the benchmarked code, with a hash identifying the exact version that ran;
- the paper;
- the authors' original paper source and original code, when available, with a record of where they came from (if either is missing, record that; do not pretend).

Outputs, stored with the design's other inputs:

- the proof document, structured as described under the heading on the proof document in order;
- a provenance record: the version of this guide the proof follows, the hashes of the proof document and of the code it was written against.

The proof is built mainly on the benchmarked code

The proof is built mainly on the benchmarked code. It is the code that produced the results; it is the ground source. Every assumption, definition, lemma and theorem of the document is about it. Before starting, confirm that the code at hand is the version that ran (its hash must match the version that produced the results). If it does not, stop and resolve that first; a proof about a different version is worthless.

The proof can also look at the paper and at the authors' reference code, to understand the benchmarked code:

- The authors' original code shows what a line of the benchmarked code is a port of and what an ambiguous line intends. Nothing from it enters the document unless the benchmarked code does the same thing, and then the document describes the benchmarked code.
- The paper and its source matter for one specific thing: assumptions. The paper motivates the method's steps and states the assumptions under which they are meant to work. Read it to understand the motivation of a part of the code and to find assumptions relevant to it (what is taken to be given, what is taken to be negligible, what a step presumes about its input). Such an assumption, when the code indeed relies on it, is written in the document as an assumption in the document's own words. Nothing is quoted from the paper, and no definition or theorem is written from the paper: the paper may tell you what to assume, the code tells you what is defined.

What is proved

For each design, one statement: running the benchmarked code on any input that satisfies the stated assumptions produces an output with the property the specification requires.

The specification is fixed by the problem the design solves, not by the design. State it precisely: what the input is, what the output is, the quantity or relation the claim is about, and the exact form of the claim (an equality, an inequality, a containment, an identity of distributions). Where the design's output determines an exact amount (how much is saved, how many items are kept), prove the exact amount, not only its direction.

The problem statement is written so that the specification is the natural thing to prove: the problem states what must be achieved, the code is a procedure, and the theorem says the procedure achieves it. If the claim depends on how the output is stored or used, state that as an explicit assumption; without it the claim may not hold, and the document must say so in that assumption.

The proof document, in order

Everything in the document is one of: assumption, definition, lemma, theorem, proof. There are no explanatory paragraphs and no prose that describes the design in words, because such prose is not a statement that can be checked. What belongs is a definition as a formula, an assumption stating what is taken as given, and a theorem stating what follows. Every statement must be one that Lean could verify: it has a precise form (a formula, a set, a function, an inequality, an equality, a count), it names the assumptions it uses, and if it is a lemma or theorem it has a proof. Where a reader would need a word of explanation, put it inside the statement being explained (a definition's label, a remark line inside a definition saying which part of the code it transcribes), not in a paragraph between statements.

Symbols

First, list every symbol that appears in the definitions and in the theorems: shapes and sizes, inputs, parameters, intermediate objects, the quantity the specification is about. For each symbol give its meaning and where it is defined.

Problem statement

State what the input is, what the output is, the quantity or relation the specification is about, and the claim to be proved. Immediately after the problem statement: the assumptions the problem statement itself needs so that its objects exist, each a formal statement with quantifiers written out, and one lemma that the problem is well posed. All other assumptions are introduced at the section of the code that first needs them.

Where the problem definition comes from. If several designs should share one problem definition, the user provides it; use it as given, for every one of those designs. Otherwise, come up with the simplest problem definition that can be verified: the smallest input, output and claim that the code's behaviour can be checked against.

Reading the code exactly

Always, before writing a line of the proof, read the code itself. The document transcribes it directly. Every definition must follow the code:

- in the order the code runs, with the same indexing (say which convention is used and convert consistently), the same rounding, the same tie behaviour and the same boundary cases;
- covering every branch of the code that can run in the benchmark, including flags fixed to a constant. A branch that cannot run is handled by an assumption stating the setting that excludes it, as described under the heading on sections of the code;
- without programming-language constructs. Every library call, array operation or sentinel value becomes mathematics: a set-builder, a sum with its indices written out, a maximum over an explicit set, a selection with its tie-break, a slice with its bounds.

Sections of the code: assumptions, definitions, theorems

Divide the code into logical sections following its own structure: the main body in execution order and each function at the point where it is called. A section is one idea; it can be one line or several. If a single line needs small subsections to be defined properly, make the subsections. A line coverage list at the start assigns every line of the code that matters for the output (function headers, loops, returns, constant flags) to exactly one section; no line is skipped. Every line of the code is proved: it is defined, its assumptions are stated, and a theorem says what it establishes.

Each section consists only of:

1. Assumptions. What the section needs to be true, continuing the numbering. Each assumption is a formal statement with quantifiers written out, in method-independent terms (no line numbers, no document-internal symbols, no forward references), so it can be compared with other methods' assumptions. When a section reuses an earlier assumption, restate it in one line with its label. A benchmark setting that fixes a flag or excludes a branch of the code is an assumption here. An assumption the paper makes about a step and that the code relies on, as described under the heading on building the proof on the benchmarked code, is an assumption here too, in the document's own words.
2. Definitions. One definition per object the section computes, as a formula, with every choice the code makes spelled out (order, tie-break, boundaries, rounding). Each definition carries, inside it, the part of the code it transcribes, so that the correspondence from code to mathematics is itself part of the record. Where the code computes a thing in a particular way, the definition is the code's way, and a lemma states what it equals.
3. Lemmas and theorems, with proofs. From the section's definitions and assumptions: well-definedness, counts, ranges, partition facts, and always a specification clause: what the section does to the property the specification is about (preserves it, establishes it, or changes it by how much).

Every logic line of the code is a definition

Each line of the code that does something for the design's purpose is stated explicitly as a definition (or as an assumption, when it is a setting). Administrative lines, those that initialise components, register callbacks, log, reset counters or move data between devices, are not stated; everything else is, down to initial values, shapes, boundary cases, reductions and the exact form of the output.

A code coverage list names every function, branch and non-trivial statement of the code with the definition, assumption or theorem that states it, and for the administrative lines the word administrative. Nothing is too small. Each entry names the part of the code, the document object, and one of: defined, assumed (the setting that excludes or fixes it), or administrative (cannot affect the output). An entry that could affect the output and has no definition is a gap in the proof and must be closed before the document is finished.

Run the coverage test. When the document is drafted, and again after any change to it, go through the code line by line and check that every line is proved somewhere: each line of the code maps to a definition, an assumption or a theorem of the document, or is marked administrative with a reason. Check it in both directions: every entry of the coverage lists points to an existing statement, and every logic line of the code appears in an entry. The test passes only when no line is left uncovered; a single uncovered line means the proof does not describe the code, and the document is not finished.

The purpose is that every design decision in the code is explicit in the document, that every concept introduced in the code is covered by the mathematical notation, and that the covered pieces connect: each definition feeds the next, and the main theorem is reached by chaining them.

The main theorem: the code meets its specification

Combine the section theorems into the claim of the problem statement. The proof of the main theorem is pointers to the section theorems only. State explicitly every condition the claim needs (for strictness, for the output to be well defined) and where each comes from (assumption or lemma).

Index of assumptions

At the end: number, title, where stated (problem statement or section name), short form. By section name, never by line number.

Rules on sources

- Build mainly on the benchmarked code. Every definition is a transcription of a statement of the code. Use the paper and the authors' original code to understand the code, not to write the document.
- No quotations from the paper. Nothing from the paper's text, formulas or numbers appears in the document. The paper is read for motivation and for assumptions: when it states or implies an assumption behind a step the code performs, that assumption is written as an assumption in the document's own words, at the section that needs it, and used like any other assumption. Every definition and theorem is derived from the code and the assumptions alone.

Workflow for one design

1. Confirm the code is the version that ran. Read it fully. Read the matching parts of the original code and the method section of the paper only as far as needed to understand it.
2. Divide the code into sections, as described under the heading on sections of the code, and list its logic lines for the coverage lists.
3. Write the proof document in the order given above: only assumptions, definitions, lemmas, theorems, proofs.
4. Re-check the code. Read it again from top to bottom, with the finished document next to it, and confirm that every definition, assumption and theorem describes the code exactly: same branches, same indexing, same rounding, same ties, same boundary cases, every logic line with a definition and nothing in a definition that the code does not do. Walk the code coverage list entry by entry against the code. Fix the document where they differ.
5. Run the coverage test: every line of the code must be proved somewhere in the document. The document is finished only when it passes with no uncovered line.
6. Write the provenance record, as described under the heading on inputs and outputs.
7. Verification. Once everything is done, submit the proof document and its coverage lists to verification to catch any errors. Fix every error it reports and submit again; repeat until no errors remain, at most 5 rounds. Report any error still open after the fifth round.

Checklist

- The code is the version that ran: its hash matches the version that produced the results.
- Every line of the code is covered by the proof: the coverage test passes with no uncovered line.
- Every definition transcribes the code exactly, not the paper: same branches, indexing, rounding, ties and boundary cases.
- The proof answers the problem: the main theorem proves the specification of the problem statement.
- Every condition the main theorem needs is a stated assumption or a proved lemma; nothing is assumed silently.
\end{promptlisting}

\clearpage

\subsection{Designing Lean from proof files}
\label{app:prompts-lean}
\begin{promptlisting}{Prompt 2: writing the Lean proof from a proof document}
How to write the Lean proof of a design from its proof document

The Lean proof formalizes a finished proof document. It shows that the design as implemented meets its specification. Its main theorem is a statement about the document's definitions, and the document's definitions are transcriptions of the benchmarked code, so the theorem is a statement about the code.

Inputs and outputs

Inputs:
- the finished proof document of the design, whose coverage test passes with no uncovered line;
- the benchmarked code the document was written against, with its hash.

Outputs, stored with the proof document:
- the Lean file;
- the provenance record of the proof document, extended with the hash of the Lean file.

Before writing any Lean

Re-check the code against the document. Read the code again from top to bottom, with the document next to it, and confirm that every definition, assumption and theorem describes the code exactly: same branches, same indexing, same rounding, same ties, same boundary cases, every logic line with a definition and nothing in a definition that the code does not do. Walk the list that assigns every line of the code to a statement, entry by entry, against the code. If they differ, fix the document first and run its coverage test again. Lean is only started once the document and the code match.

The Lean file follows the document

The Lean file follows the document's proof structure as closely as possible:
- one namespace per design, and a comment naming each document section, in document order;
- the problem's assumptions are the fields of a structure; later assumptions are fields of further structures introduced at the section that needs them, with the same labels in their docstrings;
- every definition of the document is a definition with the document's label in its docstring; every section theorem is a theorem with the same name and statement, including its specification clause; counts are their own lemmas;
- the main theorem is the statement of the document's main theorem, proved from the section theorems in the same order as the document;
- the main theorem depends only on Lean's standard foundations (propext, Quot.sound, Classical.choice); no sorry placeholder anywhere;
- indexing: say at the top which convention the Lean file uses and how document positions map to it.

The Lean file proves the same statement as the document, with the same intermediate lemmas in the same order.

Why the chain matters

A reader who trusts the Lean proof must be able to trust the code, because every object of the Lean file is a definition of the document, and every definition of the document is a transcription of a named part of the code. If that chain breaks anywhere (a definition that paraphrases the paper rather than the code, a logic line of the code with no definition), the Lean proof does not say anything about the code, and the document must not claim that it does.

Workflow

1. Confirm the document's coverage test passes and re-check the code against the document, as described under the heading on what to do before writing any Lean.
2. Write the Lean file following the document, as described under the heading on how the Lean file follows the document.
3. Compile it. Check what the main theorem depends on: only Lean's standard foundations, no sorry placeholder.
4. Add the hash of the Lean file to the provenance record.
5. Verification. Once everything is done, submit the Lean file to verification to catch any errors. Fix every error it reports and submit again; repeat until no errors remain, at most 5 rounds. Report any error still open after the fifth round.

Checklist

- The Lean file covers the code and the original proof document: every definition and theorem of the document is in it, under the same label.
- Lean compiles with no errors, internal errors included; no sorry placeholder.
\end{promptlisting}

\clearpage

\subsection{Designing variants}
\label{app:prompts-ablation}
\begin{promptlisting}{Prompt 3: deriving a variant by changing one assumption}
How to write a variant proof: perturb the assumptions of a design, derive a new design, prove its specification

A variant proof starts from a finished design proof (the parent) and produces the design proof of a new design (the child). The child's proof is the parent's proof rewritten:

- every unchanged statement is copied verbatim;
- every change is marked so that a reader can tell what was removed and what was added, with the new text written right after the old text it replaces;
- everything taken from another method is also marked as borrowed, with its source named.

The claim proved is the same: running the child's code meets the parent's specification. Always use the parent's problem definition: the child's problem statement is the parent's, copied verbatim.

Always start from the parent's code and proof document. Before anything else:

1. Create a new workspace for the child, named after the parent and the variant.
2. Copy the parent's code and proof document into it unchanged.
3. Make every change on these copies. The originals stay untouched: they give the parent's hash, the impact set and the final diff.

The principle: new designs come from perturbing assumptions

The parent's proof makes every assumption of the design explicit. That list is the parent's assumption set. Ideally, a new design is made by changing that set and deriving the algorithm it calls for:

1. Relax an assumption: weaken it, widen its bound, or drop it.
2. Augment the system: strengthen an assumption, narrow its bound, or add a new one.
3. Re-derive. Work out which definitions no longer follow from the assumptions, which new definitions the changed assumption requires or allows, and what the code must do differently. That code is the new design.

The direction is from assumptions to code, never from code to assumptions. A code tweak that changes no assumption is not a variant in this sense. If an idea starts from the code, first find the assumption it perturbs. If the parent's proof does not state that assumption (because the design choice is built into a definition), write it as an assumption first, then perturb it.

The best source of perturbations: what other methods assume

Methods for the same problem answer the same design questions differently. The answers are the assumptions of their proofs. They are already collected in the assumption catalog that comes with the design proofs: every assumption of every design proof, grouped by the question it answers (its class), with each method's pseudocode. Reuse this catalog; do not collect the assumptions again. Look through it for design features to reuse: an assumption, a definition, a rule, a pseudocode line.

State every borrowing. If the child uses an assumption or design feature of another method, name the source (method and label) in the perturbation summary and at the statement itself, and mark the borrowed text as borrowed in the child's proof.

To create a new design:

1. Pick a class in which the parent's choice differs from what other methods assume.
2. Replace the parent's assumption by another method's variant, or by a combination of the two (both, a mixture, an interpolation).
3. Derive the new design from the changed assumption set.

A technique that works in one method is then tried inside another design whose other assumptions stay fixed. This is how ideas move between methods in a controlled way: one assumption at a time.

Before borrowing, check the interface: the borrowed assumption must make sense with the parent's other assumptions (same objects, comparable units). If it does not, the derivation needs a compatibility step, marked as a change: a change of units, a renormalisation, or an extra assumption.

Kinds of perturbation

The moves below are examples, not a complete list. You can also use a perturbation that fits none of them, or a design feature that no other method has, if it is reasonable for the parent's design. It still has to change an assumption and be a good variant in the sense below. Such a feature is the child's own, so mark it as a change, not as borrowed.

- Borrow: replace the parent's variant of a class by another method's variant.
- Combine: keep the parent's variant and add another method's.
- Relax a bound: widen a range, or move to its endpoint.
- Augment a bound: narrow a range, or impose a floor.
- Drop: remove an assumption the code relies on.
- Add: a new requirement the design must satisfy.
- Split: a constant with two roles becomes two assumptions; perturb one.
- De-uniformise: one constant for all components becomes one per component.
- Change a quantifier: "for all" becomes "for a subset"; one aggregation becomes another.
- Interpolate: a parameter moving between two methods' variants of one class.

Controls (random choices, inverted decisions) are not new designs. They are reference points and make effect sizes interpretable.

What makes a good variant

1. One perturbation. One assumption is relaxed, augmented or borrowed, and every other edit follows from it. Two independent perturbations make two new designs. Compose them afterwards if both are interesting.
2. The change is real. The perturbed assumption changes what the code executes in the benchmark. The child is not equivalent to the parent: prove a separation lemma, a concrete input on which the two outputs differ. Before spending GPU time, measure how often the two outputs differ on a few inputs.
3. The parent is a special case. Prefer a family with a parameter that has a value which recovers the parent. Prove a reduction lemma: the child at that parameter value equals the parent. The lemma doubles as the check of the implementation, and sweeping the parameter gives a dose-response curve.
4. Small proof delta. A small perturbation touches one or two sections of the proof. If changes spread everywhere, the new design is far from the parent and no longer isolates one assumption.
5. Hypothesis from the proof, before the run. State what guarantee the perturbation removes or adds, and what that predicts.
6. Let the proof reveal confounds. Compute the impact set first. It shows when one assumption plays two roles, or when a perturbation silently triggers another theorem.
7. All else equal. Same hyperparameters, harness, samples, settings and models as the parent. Report the change against the parent per setting, not only the mean.
8. Not a duplicate. The same (parent, perturbation) pair has not been built already. The finished child is not one-to-one an existing method; otherwise it reuses that method's code and is not new, so it is not a variant.

Marking changes

- Mark every change so that a reader can tell what was removed, what was added and what was borrowed, and name the source of anything borrowed.
- Unchanged statements are verbatim: no rewording.
- Granularity: change the smallest unit (one bound, one symbol, one table cell, one pseudocode line, one code line). If more than half of a statement changes, mark the whole statement as removed and restate it.
- A removed statement stays in place, marked as removed.
- Assumption labels stay the parent's. A replaced assumption keeps the same number, with the old text marked as removed and the new text added. A changed bound keeps the same number, with only the bound edited. A removed assumption's number is retired, never reused. An added assumption gets a suffix label (for example a letter after the number) and is placed where first needed. Never renumber.
- An assumption, definition or pseudocode line taken from another method is new text and is also marked as borrowed, with its source (method and label) named first. A borrowed assumption keeps the child's label. Its title also names the class it comes from. Its symbols are rewritten in the parent's notation.

The document, in order

1. Title: from the parent to the child, naming the changed assumptions, the changed lines of the code, and the proof that the child meets the specification. Then a short note that tells the reader how removed, added and borrowed text is marked.
2. Perturbation summary (facts only): the parent proof and its hash; the perturbed assumption, old and new, with its class, and, if it is borrowed, the method and label it comes from; the code lines changed and the hash of the child's code; the impact set; the guarantees lost and gained.
3. Symbols and problem statement: the problem statement is the parent's, verbatim; edits in the symbols only where the perturbation reaches them.
4. Pseudocode, checked against the child's code: a changed line is marked as removed, then the replacement follows as an unnumbered added line; an added line is tagged with the previous number and a letter; a removed line is marked as removed and keeps its number. The parent's line numbers stay valid.
5. The code change: the touched lines of the code in file order, each marked as removed or added, ending with "All other lines are unchanged". Then the list that assigns every line of the code to a statement and the list that assigns every statement to the code it covers, with the changed entries marked.
6. Sections, in the parent's order. For each statement, the treatment depends on its status. Outside the impact set: copied verbatim. Hypothesis list changes: edit the list and re-check the proof. Conclusion changes: edit the conclusion and the changed proof steps, and re-prove in full. Guarantee lost: mark the clause as removed and add a weaker version with its extra hypothesis, if one holds. Guarantee gained: add a new lemma or theorem. Every section theorem keeps its specification clause. If the child computes new quantities, extend the assumptions that cover the parent's quantities to cover them.
7. Main theorem for the child, then a short section "the child versus the parent": the reduction lemma; the separation lemma; the proof delta list: every parent statement, by label and title, with its status.
8. Index of assumptions: changed, removed and added entries marked as such.

The impact set

Before editing, compute on the parent's clean proof document, for the perturbed assumption (and, if needed, a symbol it introduces):

1. the statements that cite the perturbed assumption (rewrite candidates);
2. the statements downstream of them (re-check what they use);
3. the statements that name it only inside a hypothesis range;
4. the table and index lines that mention it.

Only these statements can change. Everything else is copied verbatim. The list is a reading aid, not a proof: the final re-read against the child's code is still required.

Workflow

1. Pick the parent. List its assumption set, adding the implicit design choices as assumptions.
2. Open the assumption catalog. For each class, compare the parent's choice with the other methods' variants. Choose one perturbation: borrow, combine, relax, augment, drop or add, or one of your own.
3. Create the child's workspace and copy the parent's code and proof document into it. All later steps edit these copies.
4. Write the perturbed assumption, the hypothesis and the expected guarantees lost or gained, before any code. Name the source method and label of anything borrowed.
5. Derive the new definitions. Implement them in the copied code with the smallest diff to the parent's. Record the hash and the diff.
6. Duplicate check, once the design is finished. Compare the child's assumptions and pseudocode with every known method. Look hardest at the method it borrows from and at the parent's family. There must be no one-to-one match. If one method could be identical, compare its outputs with the child's on a few inputs. Identical outputs mean the child is that method, not a variant: change the perturbation.
7. Before the GPU: the reduction check (the child at the parameter value that recovers the parent reproduces the parent's output) and the non-equivalence check (the outputs differ).
8. Compute the impact set on the parent's original proof document. Apply the edits to the copied document in document order, marking each as removed, added or borrowed.
9. Re-read the child's code top to bottom against the finished document.
10. Diff the child's document against the parent's: every difference must be a marked change (removed, added or borrowed).
11. Run on the parent's evaluation setting. Report the change against the parent per setting.
12. Literature review, once the design is finished. Search published papers, preprints and open-source code for methods with the same assumptions or the same algorithm. List the closest ones with citations and say how the child differs from each. If a published method matches the child one-to-one, the design is not novel: change the perturbation.
13. Verification. Once everything is done, submit the child's code, proof document and literature review to verification to catch any errors. Fix every error it reports and submit again; repeat until no errors remain, at most 5 rounds. Report any error still open after the fifth round.

Checklist

- Child workspace created; started from copies of the parent's code and proof document.
- One assumption relaxed, augmented or borrowed; every other edit follows from it.
- Everything borrowed from another method: source method and label named, text marked as borrowed, interface checked.
- The finished child checked against every known method: no one-to-one match.
- The perturbation changes what the code executes; separation lemma proved.
- Reduction lemma, if the child is a family.
- Impact set computed; everything outside it verbatim, everything inside it re-checked.
- Removed text and added text marked as such; assumption labels stable.
- Specification clauses re-confirmed; main theorem proved for the child.
- Guarantees lost and gained listed; hypothesis written before the run.
\end{promptlisting}

\section{OpenEvolve and AIDE as run here}
\label{app:harnesses}
\label{app:repro-harnesses}
OpenEvolve \citep{sharma2025openevolve}, an open implementation of AlphaEvolve \citep{novikov2025alphaevolve}, evolves
a population of programs in a MAP-Elites archive; AIDE \citep{jiang2025aide} is a tree search over whole programs that
drafts, improves or debugs one node per step. Both run unmodified from pinned releases (OpenEvolve v0.4.0, commit
162c3fe; AIDE v0.2.2, commit ad73d22), start from the same parent code as our harness, and use the same
model (Claude Fable 5.1, high effort, through an OpenAI-compatible proxy), the same two-stage evaluator (CPU checks on
tiny random models, then the 12 protocol cells on the GPU queue) and the same budget: ten GPU-evaluated algorithms or 25
model calls.

\paragraph{Shared settings.}
Settings not listed below are the release defaults. Both tools call Claude Fable 5.1 (effort high) with no
temperature, a reply cap of 100,000 tokens and a 3,600\,s call timeout, and give the evaluator a 36,000\,s timeout,
since one evaluation waits for 12 queued GPU jobs. Each search stops at ten GPU-evaluated algorithms or 25 model calls,
and we repeat it with three seeds per parent. None of these changes tunes a tool for a better score; each adapts it to
the shared protocol.

\paragraph{OpenEvolve.}
OpenEvolve runs diff-based evolution with evolve blocks enforced (\texttt{enforce\_evolve\_blocks = true}), so edits
outside the evolve block are reverted, which closes a reward hack that CacheCraft reports
\citep{poudel2026cachecraft}. We shrink the population to fit ten algorithms: one island with a population of 20 and an
archive of 10, no migration (\texttt{migration\_rate = 0}), and five MAP-Elites feature bins, since the default five
islands of 1,000 programs would never fill in ten algorithms. The elite, exploration and exploitation ratios are 0.2,
0.2 and 0.7; the prompt shows the three top programs and no diverse programs.

OpenEvolve evaluates the algorithms one at a time, so each algorithm sees all earlier ones. With
\texttt{parallel\_evaluations = 1} and a single cascade threshold of 0.5, algorithm $k$ sees algorithms $1,\dots,k-1$
and their per-cell score tables, which needs an artifact limit of 40,960 bytes. The code-length limit is 80,000
characters, since the parents have 400--650 lines.

\paragraph{AIDE.}
AIDE starts from the parent, with Fable as the coding model for drafting, improving and debugging. We insert the
parent as the root node with its measured score (\texttt{num\_drafts = 1}) instead of five drafts from scratch, so AIDE
starts where the other searches start. Code formatting is off, so the parent's file is not rewritten, and every program
runs in our two-stage evaluator instead of AIDE's interpreter.

AIDE takes its metric only from our evaluator. There is no review or report model: the evaluator's score is the
metric, so no model call can report a metric that was not measured, and the metric is maximised
(\texttt{metric\_maximize = true}) rather than inferred. The step limit is 25 with the same stop at ten GPU-evaluated
algorithms; \texttt{debug\_prob} and \texttt{max\_debug\_depth} keep their defaults of 0.5 and 3.

\section{All algorithms}
\label{app:wall}
We formalized 89 algorithms, each with one proof graph: 44 for KV-cache eviction, 26 for speculative decoding and 19 for
LLM weight pruning. The count is of proof graphs. It includes published configurations that combine or modify
published methods (such as CaM on top of TOVA or LAMP on top of Wanda) and four standard baselines: autoregressive
decoding, speculative sampling in two draft-length settings and the dense model. It leaves out the two reference
points of KV-cache eviction, no compression and random eviction. The 44 KV-cache eviction algorithms are 30 KVPress algorithms, among them H2O \citep{zhang2023h2o},
SnapKV \citep{li2024snapkv},
PyramidKV \citep{cai2024pyramidkv}, KNorm \citep{devoto2024knorm}, CaM \citep{zhang2024cam}, AdaKV \citep{feng2024adakv},
Expected Attention \citep{devoto2025expected}, KeyDiff \citep{park2025keydiff}, CUR \citep{sengupta2025curdkv} and CAP
\citep{yang2026capkv}, and 14 ported algorithms, among them EchoPress \citep{lin2026echopress} and DefensiveKV
\citep{feng2026defensivekv}. The 26 speculative-decoding algorithms are 5 Spec-Bench algorithms (Lookahead
\citep{fu2024lookahead}, PLD \citep{saxena2023pld}, REST \citep{he2024rest}, SAM-Decoding \citep{hu2025samdecoding} and
token recycling \citep{luo2025tokenrecycling}), 18 ported implementations and 3 baselines.

The 19 LLM weight-pruning algorithms are the dense model and 18 published algorithms and published settings
(such as DSnoT without skipping, RIA with channel permutation and LAMP on top of Wanda or SparseGPT), among them
magnitude pruning \citep{han2015magnitude},
LAMP \citep{lee2021lamp}, SparseGPT \citep{frantar2023sparsegpt}, Wanda \citep{sun2024wanda}, OWL
\citep{yin2024owl}, DSnoT \citep{zhang2024dsnot}, E-Sparse \citep{li2023esparse}, GBLM-Pruner \citep{das2023gblm},
RIA \citep{zhang2024ria}, Pruner-Zero \citep{dong2024prunerzero}, ALPS \citep{meng2024alps} and Z-Pruner
\citep{bhuiyan2025zpruner}.
Figure~\ref{fig:fourgraphs} draws three algorithms of each field.

\section{Uncertainty estimation}
\label{app:uncertainty}
\label{app:repro-uncertainty}
The error bars of the paper come from three sources of variation, and each figure uses the one that varies between
its runs. A KV-cache eviction score is deterministic for a given set of questions (greedy decoding, fixed seed), so its
uncertainty comes from which questions were drawn. A speculative-decoding speedup is a ratio of wall-clock times, so its
uncertainty comes from repeating the same runs. A search result depends on the search's own random choices and on the
model's replies, so its uncertainty comes from repeating the search. Unless stated otherwise, $\pm$ and error bars are
one standard deviation (SD), and tests are two-sided, made cell by cell at $p<0.05$ and not corrected for the number of
cells.

\subsection{KV-cache eviction: resampling the questions}
\label{app:unc-questions}
The SD of a KV-cache eviction score comes from a bootstrap over its questions. A RULER score is the mean over the 13
tasks of the per-task accuracy on the drawn questions (LongBench: 16 tasks). We resample the questions of each task
with replacement, keeping each task's count, recompute the score, and repeat this 10,000 times; the SD of the 10,000
scores is the bootstrap SD reported next to every score. It says how much the score would move under another draw of
the same number of questions, and it shrinks with that number: 0.2 to 0.5 points per cell at 6,500 RULER-4K questions.

Means over cells and differences between two algorithms combine the per-cell SDs. A mean over cells (the 12
model--factor cells, the two models, or the six factors) treats the cells as independent runs, so the SD of a mean of
$k$ cells is $\sqrt{\sum_i \mathrm{sd}_i^2}/k$. A difference between two algorithms has SD
$\sqrt{\mathrm{sd}_a^2+\mathrm{sd}_b^2}$ and we test it with a $z$-test; this ignores that both algorithms answer the same
questions, which usually makes the test conservative. The mutation study pairs the two instead: it scores the mutant
and its parent on the same resampled questions, and the 2.5th and 97.5th percentiles of the 10,000 resampled
differences give a 95\% interval.

\subsection{Speculative decoding and systems profiles: repeating the runs}
\label{app:unc-runs}
Each Spec-Bench cell runs three times with seed 0, and its error bar is the SD over these runs. We submit run $k$ of
the parent and run $k$ of the variant back to back to the same machine, and each speedup divides by the autoregressive
run of the same model and temperature. A cell reports the mean of the three runs and, as its error bar, their sample
SD; we test a difference with a paired $t$-test over the three run pairs (two degrees of freedom). This SD measures
timing variation (clock speeds, scheduling, other jobs on the machine), not variation over questions: every run uses
the same 480 questions. Accepted tokens per target pass (\#MAT) are identical in every run at a fixed seed, so they
carry no error bar and their differences are exact.

The systems profiles measure variation over questions instead, with a paired bootstrap. The profiles behind
panels G and I of Figure~\ref{fig:grid} run three paired timing runs per cell on all 480
Spec-Bench questions; the error bars of panel G are the SD over the three runs, those of panel I the SD of a paired question bootstrap (10,000 resamples of the question ids,
the same resample applied to every run and to the paired autoregressive passes). Table~\ref{tab:ttft-context} and the
panel C of Figure~\ref{fig:grid} use seven passes over two fixed prompts; their error bars are the sample SD
over the passes.

\subsection{Design searches: repeating the search}
\label{app:unc-searches}
Each harness runs three times from the same parent. OpenEvolve and AIDE take a different seed in each repeat. Our
harness has no seed: the repeat number only labels the run, and the model's proposals are not seeded, so its repeats
largely replay the same sequence of moves. Figure~\ref{fig:harness} plots, after each algorithm, the mean over the three
repeats of the running-best score, with error bars spanning the minimum and maximum over the repeats; the text also
reports the sample SD over the repeats.

The min--max range and the SD are rough estimates, and both understate the variation of our harness. Both contain the
search's own variation and the question-draw uncertainty of each score (about 0.45 points for a 12-cell mean at 200
questions). With three repeats, each is itself a rough estimate, and for our harness, whose repeats are not
independent, each understates the variation one would see across independent searches.

\section{The worked example in full}
\label{app:sps}
\label{app:sps-assumptions}

Section~\ref{sec:method-sps} sketched why speculative sampling \citep{leviathan2023fast} is exact; this appendix walks
through the whole argument. A small drafter proposes tokens, the large target model checks them in one pass, and the
text that comes out must be distributed exactly as if the target had written it alone. We first fix the setting and
the problem, then follow one drafted token through the accept-or-resample step, show that the token that comes out
has the target's law, extend this to a block of drafts and to the whole loop, and end with how many tokens one target
pass buys.

Every statement carries the name it has in Figure~\ref{fig:proofgraph}, whose graph is repeated below;
that figure shows Assumption~\ref{as:gamma}, Definition~\ref{def:step-full}, Theorem~\ref{thm:exact-full} and the
problem $\mathcal P$ word for word. Every lemma, theorem and the
corollary below is proved in Lean~4 with Mathlib, statement for statement (no \texttt{sorry}, no added unproved statement).

\begin{figure}[H]
  \centering
  \includegraphics[width=0.3\textwidth]{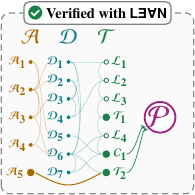}
  \renewcommand{\thefigure}{\ref*{fig:proofgraph}}\def\theHfigure{proofgraph.repeated}%
  \caption{\textbf{The proof graph of this appendix, repeated from Figure~\ref{fig:proofgraph}.}
  Lines are drawn after transitive reduction, so indirect citations are omitted; the arrows into $\mathcal P$ come
  from the claims that answer the problem. Enlarged discs mark Assumption~\ref{as:gamma}, which SVIP relaxes, and the
  exactness and token-count claims; the dark curve along the bottom is the only citation of
  Assumption~\ref{as:gamma}, by Theorem~\ref{thm:tokens-full}.}
  \label{fig:proofgraph-repeated}
  \addtocounter{figure}{-1}
\end{figure}

In the repeated graph (Figure~\ref{fig:proofgraph-repeated}), each disc is the statement of the same name below; a
line from one disc to another means the second cites the first.

\setcounter{assumption}{0}\def\theHassumption{app.\arabic{assumption}}
\setcounter{definition}{0}\def\theHdefinition{app.\arabic{definition}}
\setcounter{lemma}{0}\def\theHlemma{app.\arabic{lemma}}
\setcounter{theorem}{0}\def\theHtheorem{app.\arabic{theorem}}
\setcounter{corollary}{0}\def\theHcorollary{app.\arabic{corollary}}

\subsection{The setting and the problem}

Two language models share a finite vocabulary $\mathcal V$. We write $\mathcal V^*$ for the finite sequences over it,
$u\in\mathcal V^*$ for a prefix, and call a map $\mathcal V\to[0,1]$ that sums to $1$ a \emph{distribution} on
$\mathcal V$. The first assumption says what the large model is: something that, given a prefix, returns a
distribution over the next token. Where it matters, an assumption also says what it does \emph{not} require,
because that is what the agent is allowed to change.

\begin{assumption}[Target distribution over a finite vocabulary]\label{as:target-full}
For every prefix $u\in\mathcal V^*$ the target model $M_p$ defines a distribution $p(\cdot\mid u)$ on $\mathcal V$.
Temperature, top-$k$ and nucleus adjustments are folded into $p$ (standardized sampling); temperature $0$ makes
$p(\cdot\mid u)$ a point mass on an arg max (ties broken by a fixed rule). Nothing is assumed about how $p$ depends on $u$.
\end{assumption}

The second model is the drafter. It must use the same vocabulary, and nothing else is asked of it: the argument has
to work however poor the drafter is.

\begin{assumption}[Draft distribution over the same vocabulary]\label{as:draft}
For every prefix $u$ the draft model $M_q$ defines a distribution $q(\cdot\mid u)$ on the same $\mathcal V$ as
Assumption~\ref{as:target-full}. Nothing relates $q$ to $p$: $q$ need not be close to $p$, need not have $q(x\mid u)>0$ where $p(x\mid u)>0$, and
may be a point mass.
\end{assumption}

With the two models fixed, the problem can be stated.

\paragraph{The problem $\mathcal P$.}
Given a target model $M_p$ and a cheaper drafter $M_q$ over the same vocabulary, emit a continuation whose law is
exactly that of autoregressive sampling from $M_p$ (Corollary~\ref{cor:loop-full}), while one target call emits up to
$\gamma+1$ tokens, with the expected number given by Theorem~\ref{thm:tokens-full}.

\subsection{One position: accept or resample}
\label{app:sps-defs}
\label{app:sps-proofs}

Speculative sampling decides the fate of one drafted token at a time. Fix one position and write $p,q$ for the
target's and the drafter's distributions there. A token is accepted with certainty when the target likes it at least as
much as the drafter, and otherwise only in proportion:

\begin{definition}[Acceptance probability]\label{def:accept}
For $p$ and $q$ at one position (Assumptions~\ref{as:target-full} and \ref{as:draft}),
$a(x)=\min\!\big(1,\,p(x)/q(x)\big)$ when $q(x)>0$, and $a(x)=1$ when $q(x)=0$. (The value at $q(x)=0$ never
matters: such an $x$ is never drawn from $q$. A Lean file is free to use its own convention for division by zero;
Lemma~\ref{lem:pointwise} holds for every choice.)
\end{definition}

A rejected token is not replaced by a fresh draw from $p$, which would give the accepted tokens too much weight. The
replacement comes from the part of $p$ that acceptance leaves uncovered:

\begin{definition}[Residual distribution]\label{def:residual-full}
For $p$ and $q$ at one position (Assumptions~\ref{as:target-full} and \ref{as:draft}) let
$Z=\sum_{x}\max(0,\,p(x)-q(x))$. If $p\ne q$ then $Z>0$ (if $Z=0$ then $p\le q$ everywhere, and as both sum to $1$,
$p=q$), and $p'(x)=\max(0,\,p(x)-q(x))\big/Z$. For $p=q$ set $p'=p$; by
Lemma~\ref{lem:beta-one} the branch that draws from $p'$ then has probability $0$, so any distribution would do.
\end{definition}

The step also needs its randomness to behave: drafts drawn from $q$, an acceptance coin drawn independently of them,
and a replacement drawn from the residual. The assumption is stated for a step of $\gamma$ drafts, used later; a
single position is its first draft.

\begin{assumption}[Independent verifier randomness]\label{as:uniform-full}
Within a step at prefix $u$ the random draws are the draft tokens $x_1,\dots,x_\gamma$, the verifier's
$r_1,\dots,r_\gamma$ and one final token $x'$ (the residual token after a rejection, otherwise the token appended
after a full acceptance). (i) Given everything drawn before it, $x_i$ has law
$q(\cdot\mid u\,x_{1:i-1})$ of Assumption~\ref{as:draft}. (ii) Each $r_i$ is uniform on $[0,1]$ and independent of $(x_{1:\gamma},
r_{1:\gamma}\setminus r_i)$. (iii) The residual token, given everything drawn before it, has the residual
distribution of Definition~\ref{def:residual-full} for the position at which it is drawn. (iv) The draws of a step
are independent of the draws of earlier steps given the prefix at which the step starts.
\end{assumption}

Together these make one step:

\begin{definition}[One speculative-sampling step at one position]\label{def:step-full}
Draw $x\sim q$ (Assumption~\ref{as:draft}) and $r\sim U[0,1]$ (Assumption~\ref{as:uniform-full}). If $r\le a(x)$
(Definition~\ref{def:accept}) the step \emph{accepts} and emits $x$; otherwise it \emph{rejects} and emits $x'\sim p'$
(Definition~\ref{def:residual-full}).
\end{definition}

Two pointwise identities carry all the algebra that follows. The first says that a token is drafted and then accepted
with probability $\min(p,q)$; the second that the residual weight of a token is what $p$ has left beyond that
minimum.

\begin{lemma}[Pointwise identities]\label{lem:pointwise}
For every $x\in\mathcal V$, with $a$ of Definition~\ref{def:accept}: (a) $q(x)\,a(x)=\min(p(x),q(x))$; (b) $\max(0,\,p(x)-q(x))=p(x)-\min(p(x),q(x))$.
\end{lemma}
\begin{proof}
(a) If $q(x)=0$ both sides are $0$, since $p(x)\ge0$. If $q(x)>0$ and $p(x)\ge q(x)$ then $a(x)=1$ and both sides
are $q(x)$. If $q(x)>0$ and $p(x)<q(x)$ then $a(x)=p(x)/q(x)$ and both sides are $p(x)$.
(b) If $p(x)\ge q(x)$ both sides are $p(x)-q(x)$; otherwise both are $0$.
\end{proof}

How often does a step accept? \citet{leviathan2023fast} name this rate and relate it to a divergence between the two
models:

\begin{definition}[Acceptance rate and divergence, Defs.~3.1 and 3.2 of \citet{leviathan2023fast}]\label{def:beta-full}
$\beta=\Pr(\text{accept})$ in Definition~\ref{def:step-full}, and $D_{LK}(p,q)=\sum_{x}\lvert p(x)-M(x)\rvert$
with $M=\tfrac12(p+q)$.
\end{definition}

Summing the first identity over all tokens gives the rate in closed form, and the second identity shows that the
residual is normalized by exactly the probability of a rejection:

\begin{lemma}[Acceptance rate, normalizer, divergence; Lemma~3.3 and Theorem~3.5 of \citet{leviathan2023fast}]\label{lem:beta-full}
Under Assumption~\ref{as:uniform-full}, the acceptance rate $\beta$ and the divergence $D_{LK}$ of
Definition~\ref{def:beta-full} and the normalizer $Z$ of Definition~\ref{def:residual-full} satisfy
$\beta=\sum_x\min(p(x),q(x))$, $Z=1-\beta$, and $D_{LK}(p,q)=1-\beta$.
\end{lemma}
\begin{proof}
By Assumption~\ref{as:uniform-full}(ii), $r$ is independent of $x$ and uniform, so $\Pr(\text{accept}\mid x)=
\Pr(r\le a(x))=a(x)$ because $a(x)\in[0,1]$; hence $\beta=\sum_x q(x)\,a(x)=\sum_x\min(p(x),q(x))$ by
Lemma~\ref{lem:pointwise}(a). By Lemma~\ref{lem:pointwise}(b), $Z=\sum_x\big(p(x)-\min(p(x),q(x))\big)=1-\beta$.
Pointwise, $\lvert p-M\rvert=\tfrac12\lvert p-q\rvert=\tfrac12\big(\max(p,q)-\min(p,q)\big)$ and
$\max(p,q)+\min(p,q)=p+q$, which sums to $2$; so $D_{LK}=\tfrac12\big(2-2\sum_x\min(p,q)\big)=1-\beta$.
\end{proof}

Finally, a step can only be sure to accept when the two models agree, which is the case that
Definition~\ref{def:residual-full} set aside:

\begin{lemma}[When nothing is rejected]\label{lem:beta-one}
$0\le\beta\le1$, and $\beta=1$ if and only if $p=q$.
\end{lemma}
\begin{proof}
By Lemma~\ref{lem:beta-full}, $\beta=\sum_x\min(p(x),q(x))$; $0\le\min(p,q)\le p$ pointwise and
$\sum_x p(x)=1$, so $0\le\beta\le1$. If $\beta=1$ then
$\sum_x\big(p(x)-\min(p(x),q(x))\big)=0$ with non-negative terms, so $\min(p(x),q(x))=p(x)$, i.e.\ $p(x)\le q(x)$,
for every $x$; if $p(x_0)<q(x_0)$ for some $x_0$ then $1=\sum_x p(x)<\sum_x q(x)=1$, a contradiction, so $p=q$.
Conversely $p=q$ gives $\beta=\sum_x p(x)=1$.
\end{proof}

\subsection{The emitted token has the target's law}

Everything is now in place for the central claim: whatever the drafter, the token that one step emits is distributed
as $p$. The proof adds up the two ways a token can come out, accepted as a draft or drawn from the residual.

\begin{theorem}[Exactness of one step; \citet{leviathan2023fast}]\label{thm:exact-full}
Under Assumptions~\ref{as:target-full}--\ref{as:uniform-full}, the token emitted by Definition~\ref{def:step-full}
has law $p$: $\Pr(\text{emit }y)=p(y)$ for every $y\in\mathcal V$.
\end{theorem}
\begin{proof}
The step emits $y$ either by accepting a draft equal to $y$ or by rejecting and drawing $y$ from $p'$; the two
events are disjoint, so
\[
\Pr(\text{emit }y)=\Pr(\text{accept},\,x=y)+\Pr(\text{reject},\,\text{residual}=y).
\]
\emph{Accept path.} $\Pr(\text{accept},\,x=y)=q(y)\Pr(r\le a(y)\mid x=y)=q(y)\,a(y)=\min(p(y),q(y))$,
by Assumption~\ref{as:uniform-full}(ii) and Lemma~\ref{lem:pointwise}(a).
\emph{Reject path.} By Assumption~\ref{as:uniform-full}(iii) the residual token, given $(x,r)$, has law $p'$, so
$\Pr(\text{reject},\,\text{residual}=y)=\Pr(\text{reject})\,p'(y)=(1-\beta)\,p'(y)$.
\emph{Case $p\ne q$.} By Lemmas~\ref{lem:beta-full} and \ref{lem:beta-one}, $Z=1-\beta>0$, so
$(1-\beta)\,p'(y)=\max(0,p(y)-q(y))=p(y)-\min(p(y),q(y))$ by Lemma~\ref{lem:pointwise}(b), and the two
paths add to $p(y)$.
\emph{Case $p=q$.} By Lemma~\ref{lem:beta-one}, $\beta=1$, the reject path has probability $0$, and the accept
path gives $\min(p(y),p(y))=p(y)$.
\end{proof}

One position is therefore exact. A real step, however, drafts several tokens before the target checks any of them.

\subsection{From one token to a block, and to the whole loop}

In the code the drafter writes $\gamma$ tokens one after another, and the target then checks all of them in a single
forward pass. The next assumption says what that pass returns and how its verdicts are used.

\begin{assumption}[Token-wise verification]\label{as:tokenwise}
At prefix $u$ with drafts $x_{1:\gamma}$, the target forward pass returns the $\gamma+1$ distributions
$p_i=p(\cdot\mid u\,x_{1:i-1})$, $i=1,\dots,\gamma+1$, of Assumption~\ref{as:target-full}; drafted token $x_i$ is
verified against $p_i$ with draft distribution $q_i=q(\cdot\mid u\,x_{1:i-1})$ of Assumption~\ref{as:draft}; the
first rejection ends the step;
if all $\gamma$ drafts are accepted, one token is drawn from $p_{\gamma+1}$, given everything drawn before it, and
appended.
\end{assumption}

Exactness is a claim about whole sequences, so we name the law that the target alone gives a sequence,

\begin{definition}[Autoregressive reference]\label{def:ar}
For a prefix $u$ and $L\ge0$, $P^{\mathrm{AR}}_L(y_{1:L}\mid u)=\prod_{t=1}^{L}p(y_t\mid u\,y_{1:t-1})$ with $p$ of
Assumption~\ref{as:target-full}, the law of $L$ tokens sampled one at a time from $M_p$.
\end{definition}

and the step exactly as the code runs it:

\begin{definition}[A step of $\gamma$ tokens, as the code runs it]\label{def:block}
At prefix $u$: draw $x_1,\dots,x_\gamma$ in order, $x_i\sim q(\cdot\mid u\,x_{1:i-1})$ (Assumption~\ref{as:draft};
the drafter's autoregressive loop); obtain $p_1,\dots,p_{\gamma+1}$ from one target pass
(Assumption~\ref{as:tokenwise}); draw $r_1,\dots,r_\gamma$ (Assumption~\ref{as:uniform-full}); let $n=\min\{\,i:\ r_i>a_i(x_i)\,\}-1$, with $n=\gamma$ if there is no such $i$, where $a_i$
is Definition~\ref{def:accept} for $(p_i,q_i)$. The step emits the block $y=x_{1:n}\,x'$, where $x'\sim p'_{n+1}$
(Definition~\ref{def:residual-full} for $p_{n+1}$ against $q_{n+1}$) if $n<\gamma$ and $x'\sim p_{\gamma+1}$ if $n=\gamma$. The block has
$n+1\in\{1,\dots,\gamma+1\}$ tokens.
\end{definition}

Because all $\gamma$ drafts are drawn before any of them is verified, it is not obvious that the emitted tokens still
behave as fresh steps of Definition~\ref{def:step-full}. The next lemma shows that they do; it is the only place
where the order of the draws matters.

\begin{lemma}[Token-level kernel of a step]\label{lem:kernel}
Consider Definition~\ref{def:block} at prefix $u$ under Assumptions~\ref{as:target-full}--\ref{as:tokenwise}.
Let $1\le i\le\gamma$, let $y_{1:i-1}\in\mathcal V^{i-1}$, and let $E_i=\{n\ge i-1,\ x_{1:i-1}=y_{1:i-1}\}$ be the
event that the first $i-1$ drafts equal $y_{1:i-1}$ and were all accepted. If $\Pr(E_i)>0$ then, conditional on
$E_i$, the $i$-th emitted token $y_i$ has law $p(\cdot\mid u\,y_{1:i-1})$. For every $y_{1:\gamma}$ with
$\Pr(n=\gamma,\,x_{1:\gamma}=y_{1:\gamma})>0$, conditional on this event the $(\gamma+1)$-th emitted token has law
$p(\cdot\mid u\,y_{1:\gamma})$.
\end{lemma}
\begin{proof}
$E_i$ is determined by $(x_{1:i-1},r_{1:i-1})$. By Assumption~\ref{as:uniform-full}(i), conditional on
$x_{1:i-1}=y_{1:i-1}$ the draft $x_i$ has law $q(\cdot\mid u\,y_{1:i-1})$, and conditioning further on
$r_{1:i-1}$ does not change this, since by (ii) the $r$'s are independent of the drafts; by (ii) again, $r_i$ is
uniform and independent of $(x_i,x_{1:i-1},r_{1:i-1})$. So conditional on $E_i$ the pair $(x_i,r_i)$ has law
$q(\cdot\mid u\,y_{1:i-1})\otimes U[0,1]$. By Assumption~\ref{as:tokenwise}, $p_i=p(\cdot\mid u\,y_{1:i-1})$ and
$q_i=q(\cdot\mid u\,y_{1:i-1})$ on $E_i$. On $E_i$ the step is still running at position $i$, and the $i$-th
emitted token is $x_i$ if $r_i\le a_i(x_i)$ and the residual draw of position $i$ otherwise, which by (iii) has law
$p'_i$ given everything before it. This is exactly Definition~\ref{def:step-full} for $(p_i,q_i)$, so
Theorem~\ref{thm:exact-full} gives the law $p_i=p(\cdot\mid u\,y_{1:i-1})$. The last claim is
Assumption~\ref{as:tokenwise}: given everything drawn before it, the appended token has law
$p_{\gamma+1}=p(\cdot\mid u\,y_{1:\gamma})$.
\end{proof}

Repeating the block step gives the loop,

\begin{definition}[The speculative loop]\label{def:loop}
Starting from $u$, repeat Definition~\ref{def:block} at the current prefix and append the block, until at least
$L$ tokens have been emitted; $P^{\mathrm{SpS}}_L(\cdot\mid u)$ is the law of the first $L$ emitted tokens
$Y_{1:L}$.
\end{definition}

and the loop is exact: its first $L$ tokens have the law of $L$ tokens sampled from the target one at a time.

\begin{corollary}[Exactness of the loop]\label{cor:loop-full}
Under Assumptions~\ref{as:target-full}--\ref{as:tokenwise}, $P^{\mathrm{SpS}}_L(\cdot\mid u)=P^{\mathrm{AR}}_L(\cdot\mid u)$
for every prefix $u$ and every $L\ge0$, with $P^{\mathrm{SpS}}_L$ of Definition~\ref{def:loop} and $P^{\mathrm{AR}}_L$ of
Definition~\ref{def:ar}.
\end{corollary}
\begin{proof}
Let $Y_1,Y_2,\dots$ be the emitted tokens of Definition~\ref{def:loop} in order. We show that for every $1\le t\le L$
and every $y_{1:t-1}$ with $\Pr(Y_{1:t-1}=y_{1:t-1})>0$,
\begin{equation}\label{eq:markov}
\Pr(Y_t=y\mid Y_{1:t-1}=y_{1:t-1})=p(y\mid u\,y_{1:t-1})\qquad\text{for every }y\in\mathcal V;
\end{equation}
the chain rule then gives $\Pr(Y_{1:L}=y_{1:L})=\prod_{t=1}^L p(y_t\mid u\,y_{1:t-1})=P^{\mathrm{AR}}_L(y_{1:L}\mid u)$.

Token $t$ is emitted by some step; which step, and at which position $i\in\{1,\dots,\gamma+1\}$ within it, is
determined by the lengths of the blocks emitted before it. Let $H=(\ell_1,\dots,\ell_k)$ range over the lengths of
the blocks that end before token $t$ ($\ell_1+\dots+\ell_k\le t-1$, the next block containing token $t$); the events $\{Y_{1:t-1}=y_{1:t-1}\}\cap H$ partition
$\{Y_{1:t-1}=y_{1:t-1}\}$, so it suffices to prove \eqref{eq:markov} conditional on each such cell of positive
probability. Fix $H$; it places token $t$ at position $i=t-(\ell_1+\dots+\ell_k)$ of a step that started at prefix
$w=u\,y_{1:t-i}$, and on the cell the first $i-1$ tokens of this step are $y_{t-i+1:t-1}$ and were accepted.
By Assumption~\ref{as:uniform-full}(iv), the draws of this step are independent of the earlier steps given $w$,
so conditioning on the earlier steps (which fix $w$ and $H$ up to this step) changes nothing about this step beyond
fixing $w$. If $i\le\gamma$ the cell is the event $E_i$ of Lemma~\ref{lem:kernel} for the step at $w$ with
$y_{1:i-1}:=y_{t-i+1:t-1}$, so $Y_t$ has law $p(\cdot\mid w\,y_{t-i+1:t-1})=p(\cdot\mid u\,y_{1:t-1})$. If
$i=\gamma+1$ the cell says $n=\gamma$ with $x_{1:\gamma}=y_{t-\gamma:t-1}$, and the last claim of
Lemma~\ref{lem:kernel} gives the same law. Hence \eqref{eq:markov} holds on every cell, and therefore on their
union.
\end{proof}

The proof applies to the idealized loop defined above, with its stated conditional sampling and
independence assumptions. It does not certify arbitrary stopping rules or the measured SVIP cap branch.

\subsection{How many tokens one target pass buys}

Exactness says nothing about speed. The speedup comes from emitting more than one token per target pass, and to count
them \citet{leviathan2023fast} idealize the acceptances:

\begin{assumption}[Fixed draft length, independent acceptances]\label{as:gamma}
$\gamma\ge1$ is a constant, and in a step of Definition~\ref{def:block} the first $i$ drafts are all accepted with
probability $\alpha^i$ ($0\le i\le\gamma$) for a rate $\alpha\in[0,1]$, as when acceptances are independent with
common rate $\alpha$, e.g.\ when $\beta$ of Definition~\ref{def:beta-full} equals $\alpha$ at every prefix. This is an idealization used only by the cost claim
(Theorem~\ref{thm:tokens-full}); the exactness claims never cite it.
\end{assumption}

Under this idealization the count is a geometric sum.

\begin{theorem}[Expected tokens per target pass; Equation~(1) of \citet{leviathan2023fast}]\label{thm:tokens-full}
Under Assumption~\ref{as:gamma}, a step of Definition~\ref{def:block} emits
$\mathbb E[n+1]=\sum_{i=0}^{\gamma}\alpha^i$ tokens, which is $(1-\alpha^{\gamma+1})/(1-\alpha)$ for $\alpha<1$ and
$\gamma+1$ for $\alpha=1$.
\end{theorem}
\begin{proof}
$n+1=\sum_{i=0}^{\gamma}\mathbf 1[\text{the first $i$ drafts are accepted}]$, since the indicator is $1$ exactly for
$i\le n$. Under Assumption~\ref{as:gamma} the $i$-th indicator has expectation $\alpha^i$; sum the geometric
series.
\end{proof}

Without Assumption~\ref{as:gamma}, for a fixed maximum draft length the general tail-sum identity is
$\mathbb E[n+1]=\sum_{i=0}^{\gamma}\Pr(A_1\cap\cdots\cap A_i)$, with the empty intersection ($i=0$) having probability one.
Here $A_i$ denotes acceptance at position $i$. Prefix-dependent rates cannot in general be multiplied as
unconditional marginal probabilities; adaptive draft lengths additionally require accounting for whether a
position was proposed.

\section{Mutation study}
\label{app:method-ablations}

\subsection{Exactness proofs of published decoders}
\label{sec:res-proofs}
\label{sec:exp-exactness}
Formalizing a field also checks the algorithms already in it. We prove exactness for eight published speculative
decoders under stated cache, tie-breaking and exact-arithmetic assumptions; five of them claimed it without proof
\citep{gante2023assisted,mamou2024disco,zhang2025svip,agrawal2024adaedl}.

\subsection{Proof checks catch every injected bug, benchmarks only about half}
\label{app:mutation-study}
\label{sec:exp-mutations}
\label{app:mutants}
\noindent\begin{minipage}[t]{0.41\textwidth}
  \vspace{0pt}\centering
  \includegraphics[width=\linewidth]{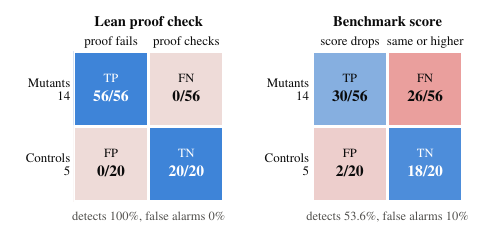}
  \plaincaption{figure}{\textbf{Lean flags every mutant; the benchmark flags about half.} An evaluation is one design at one
  setting, Llama-3.1-8B-Instruct or Qwen3-8B at $16\times$ or $64\times$ compression, so each design counts four times.
  Lean gives one verdict per design, shared by its four settings; the benchmark flags any score below the parent's,
  the rule most favourable to it. Blue cells are correct decisions and red cells wrong ones, shaded by their share of
  the row.}
  \label{fig:mutation-confusion}
\end{minipage}\hfill
\begin{minipage}[t]{0.56\textwidth}
  \vspace{0pt}
  We took five verified KV-cache eviction algorithms (KNorm, SnapKV, KVGrad, RestoreKV and KVzip) and wrote 14 code
  mutants. Each mutant injects one realistic bug, such as keeping one pair too many or too few, reversing a ranking,
  skipping a layer or forgetting to protect the sink tokens; comment-only edits and an equivalent rewrite serve as
  five controls. We checked each parent's fixed Lean proof against the mutated code and ran every mutant and control
  on 200 RULER-4K questions. Every mutant breaks its proof and every control passes, yet in \MCMBenchUp{} of the
  \MCMPos{} mutated evaluations the score matches or exceeds the parent's: some bugs are too small to move the score,
  and others raise it by silently exceeding the memory budget, such as keeping every pair.
\end{minipage}

\end{document}